\documentclass[10pt,twocolumn]{article}

\usepackage[letterpaper,margin=0.68in,columnsep=0.23in]{geometry}
\usepackage{amsmath,amssymb,amsfonts,bm,mathtools}
\usepackage[final]{graphicx}
\usepackage{booktabs,array,tabularx}
\usepackage{ragged2e}
\usepackage{microtype}
\usepackage{xcolor}
\usepackage{caption}
\usepackage{tikz}
\usetikzlibrary{arrows.meta,decorations.pathmorphing,positioning,calc}
\usepackage[most]{tcolorbox}
\usepackage{enumitem}
\usepackage{cuted}
\usepackage{placeins}
\usepackage[hidelinks]{hyperref}
\usepackage{stfloats} % allows two-column floats at the bottom of a page

\definecolor{boxblue}{RGB}{24,73,122}
\definecolor{boxlight}{RGB}{241,246,252}
\definecolor{accent}{RGB}{155,35,35}
\definecolor{grayline}{RGB}{80,80,80}
\definecolor{boxlightblue}{RGB}{238,244,252}

\usepackage{xcolor}

\newcommand{\Pred}{P_{\mathrm{red}}}

\newcommand{\WW}{\texttt{WeightWatcher}}
\newcommand{\Tr}{\operatorname{Tr}}

\newcommand{\diag}{\operatorname{diag}}

\newcommand{\Fzero}{\mathfrak F_0}
\newcommand{\Ferg}{\mathfrak F_{\rm ERG}}
\newcommand{\Opt}{\mathcal O_{\rm t}}
\newcommand{\dd}{\,d}
\newcommand{\W}{\mathbf W}
\newcommand{\X}{\mathbf X}
\newcommand{\U}{\mathbf U}
\newcommand{\V}{\mathbf V}
\newcommand{\Vt}{\mathbf V^{\top}}
\newcommand{\A}{\mathbf A}
\newcommand{\Xtilde}{\widetilde{\mathbf X}}
\newcommand{\Atilde}{\widetilde{\mathbf A}}

\newcommand{\XR}{\mathbf X_{\mathcal R}}
\newcommand{\Rell}{\mathcal R}

\newcommand{\Tell}{\mathcal T}
\newcommand{\TPL}{\mathcal T_{\rm PL}}
\newcommand{\RECS}{\mathcal R_{\rm ECS}}
\newcommand{\Splus}{\mathcal S^+}
\newcommand{\rhohat}{\widehat\rho}
\newcommand{\IZQ}{\beta\Gamma_{\overline Q^2}^{\rm IZ}}
\newcommand{\ECSstart}{\tilde{\lambda}^{+}_{\Rell}}
\newcommand{\PLstart}{\lambda^{+}_{\rm PL}}
\newcommand{\MPstart}{\lambda^{+}_{\rm MP}}

\newcommand{\Mred}{M_{\Rell}}

\newcommand{\Ered}{\widetilde{E}}

\tcbset{
  marginalbox/.style={
    enhanced,
    colback=boxlightblue,
    colframe=boxblue,
    colbacktitle=boxblue,
    coltitle=white,
    fonttitle=\bfseries\small,
    boxrule=0.75pt,
    arc=1.2mm,
    left=2.0mm,
    right=2.0mm,
    top=1.6mm,
    bottom=1.6mm,
    before skip=7pt,
    after skip=7pt,
    before upper=\RaggedRight
  }
}
\newcommand{\takeaway}[1]{%
  \par\vspace{4pt}\noindent\textcolor{boxblue}{#1}%
}

\newcommand{\eqtight}[1]{\[\vspace{-1pt}#1\vspace{-3pt}\]}

\setlist[itemize]{leftmargin=1.2em,itemsep=0.15em,topsep=0.25em}
\setlist[enumerate]{leftmargin=1.4em,itemsep=0.15em,topsep=0.25em}
\makeatletter
\renewenvironment{abstract}{%
  \par\vspace{-0.85em}\noindent{\bfseries\large Abstract}\par\vspace{0.35em}%
  \small\noindent
}{\par\vspace{0.8em}}
\makeatother

\newcommand{\safeincludegraphics}[2][]{%
  \IfFileExists{#2}{\includegraphics[#1]{#2}}{%
  \fbox{\begin{minipage}[c][0.16\textheight][c]{0.92\linewidth}\centering
  {\small Missing figure file}\par\vspace{2pt}{\small\texttt{\detokenize{#2}}}
  \end{minipage}}}}

\title{\bfseries A Spectral Renormalization-Group View of Learning}
\author{Charles H. Martin, PhD\\
\small Calculation Consulting, San Francisco, CA\\[-0.2em]
\small \texttt{charles@calculationconsulting.com}}
\date{November 5, 2026}

\begin{document}
\maketitle

\begin{abstract}
  We study how the layers in deep neural networks (NNs) converge, using Wilsonian Renormalization-Group (RG) ideas to organize the spectral picture.
  During training, these layers often develop heavy-tailed power-law (PL) spectra,
  and the strongest checkpoints often appear near the PL exponent $\alpha\approx 2$.
  This pattern raises a simple question: why should this value mark a good trained layer?
  The answer is both statistical and physical.
  Our statistical hypothesis is that $\alpha=2$ marks the boundary between stable if not optimal learning
  and overfitting via non-self-averaging.
  The physical answer is that the same value  makes the spectral energy gain scale evenly across
  logarithmic spectral scales, which is the proposed Wilsonian marginal condition.
  We then apply the ERG trace-log condition, the proposed SETOL theory, which we justify
  by taking  volume-preserving,  scale-invariant transformations
  to remove arbitrary layer scale along the proposed RG flow.
   The resulting spectral RG flow connects the HTSR and SETOL theories to optimizers
   such as Muon, and motivates a new optimizer extension, WW-PGD.
   Together, these results point toward a spectral Renormalization-Group view of learning,
   in which NNs may approach shared spectral structure in the settings studied here.
\end{abstract}

\section{Introduction}
\label{sxn:intro}
\enlargethispage{\baselineskip}

\emph{Why does Deep Learning Work?} Artificial neural networks now power large language models,
AlphaGo and AlphaFold-class systems~\cite{SilverAlphaGo2016,JumperAlphaFold2021},
and other foundation models~\cite{BommasaniFoundationModels2021,Qwen25TechnicalReport2025,OpenAIGPTOSS2025},
yet we still lack a generally accepted theory that both explains their convergence and tells practitioners how to improve training. Modern networks optimize enormous nonconvex objectives. They memorize parts of their data. They generalize in ways that remain surprising. A theory of learning should explain this coexistence, not merely describe the loss curve.

We don't have a theory of learning; we simply train until the loss stops.
But two models can reach similar losses and still build different internal structures.
If you just consider the loss, overfitting looks like pre-grokking memorization.
Optimizers like Muon whiten or orthogonalize updates without understanding why it helps~\cite{JordanMuon2024}.
These examples point to the same gap: training is not just minimizing the loss; it is motion through internal layer states.

Layer convergence is the missing object. A network trains all layers together, but each dense layer changes its own weight matrix and concentrates information in its own way. Some concentration is necessary because learning compresses data into useful representations.
Too much concentration is dangerous because a few dominant tail components can dominate the representation.
This is part of the story. But this does not explain how finite trained layers reach correlated spectral endpoints.

HTSR and SETOL provide the starting points. HTSR applies \emph{Random Matrix Theory} (RMT) to trained neural-network weight matrices.
It posits $5+1$ phases of training by analyzing a layer's \emph{Empirical Spectral Density} (ESD) and identifying signatures of heavy-tailed (HT)
universality classes beyond classical \emph{Marchenko--Pastur} RMT.
Central to HTSR is the observation that the best layers develop HT spectra fit by a \emph{power law}, with exponent $\alpha \simeq 2$.
This motivates treating $\alpha \simeq 2$ not merely as a fitted value, but as a
spectral boundary between near optimal self-averaging behavior and overfitting due to non-self-averaging.

SETOL was introduced to explain why this spectral criterion correlates with generalization. Under physically motivated approximations, SETOL expresses a trained layer's contribution to generalization solely as a functional of its empirical covariance spectrum, derived from a layer free energy written as an HCIZ integral. Thus the ESD is the relevant object: the same spectral structure that determines the HTSR exponent $\alpha$ also determines the SETOL energy functional. SETOL further suggests that the optimal regime at $\alpha = 2$ corresponds to approach toward a Wilsonian, scale-invariant, volume-preserving \emph{Exact Renormalization Group} (ERG) fixed point. In this picture, the generalizing components concentrate into a low-rank \emph{Effective Correlation Space} (ECS), whose normalized spectrum satisfies the ERG trace-log gauge-fixing constraint.

Here we thread those two theories into a proposed Wilsonian RG theory of layer convergence. The NN remains a coupled system, but near convergence each layer can be viewed as approaching its own Wilsonian-style spectral fixed point. The layers must remain compatible, but they need not share one global fixed point. The explicit claims are that SETOL derives the trace-log condition, the RG map is operational, and the $\alpha\simeq2$ claim is empirical and supported by self-averaging and scale-counting arguments.
Together, these arguments interpret the HTSR observations and give optimizer design a concrete target: move each layer toward stable correlated structure
while preventing overfitting.

\section{HTSR: phases of learning}
\label{sxn:htsr}

\begin{tcolorbox}[marginalbox,title=\textbf{HTSR heavy-tailed phases}]
  RMT-based phenomenology of layer training.
\eqtight{\rho_{\rm tail}(\lambda)\sim\lambda^{-\alpha}.}
\begin{itemize}
\item \textbf{$\alpha>2$:} stable, but the tail can be too weak.
\item \textbf{$\alpha\simeq2$:} practical optimum for layer convergence.
\item \textbf{$\alpha<2$:} overcorrelated / overfit
\end{itemize}
\takeaway{HTSR is the empirical rule: well-trained layers often sit near the $\alpha\simeq2$ boundary.}
\end{tcolorbox}

\emph{Heavy-Tailed Self-Regularization} (HTSR) is a phenomenological theory that describes the phases of training
of a NN layer using a unique application of the Marchenko--Pastur (MP)
Random Matrix Theory (RMT) in the heavy-tailed (HT) setting, classifying individual
NN layers into HT/PL universality classes~\cite{MartinMahoneyJMLR2021}.

The starting point is a single layer.
Write the weight matrix as $\W\in\mathbb R^{N\times M}$, with $N\ge M$, and define
\begin{equation}
\X=\frac{1}{N}\W^T\W.
\label{eq:Xdef}
\end{equation}
The operator $\X$ is the layer's weight-correlation operator. Its eigenvectors give input directions in the layer. Its eigenvalues give the spectral quality gain assigned to those eigen-directions. Importantly, HTSR recognizes the elements $\W_{i,j}$ should not be strongly heavy-tailed, whereas
the eigenvalues $\lambda$ of $\X$ are as this is a signature of a strongly correlated system near criticality.

HTSR works by fitting the layer ESD to a finite-window power law, where
the exponent $\alpha$ places the layer in an RMT/HT universality class.
Empirically, strong trained layers often approach the optimal state
$\alpha \simeq 2$.
Section~\ref{sxn:selfavg} explains why 
this boundary is optimal for learning using statistical/statistical mechanics
arguments.

The empirical premise comes from observations on hundreds of well-trained models and thousands of layers, using the open-source \WW{} tool~\cite{WeightWatcherTool}. \WW{} implements a practical version of the HTSR (and SETOL) diagnostics, and makes these diagnostics reproducible and empirically testable
and/or falsifiable.

\begin{figure}[t]
\centering
\safeincludegraphics[width=\columnwidth]{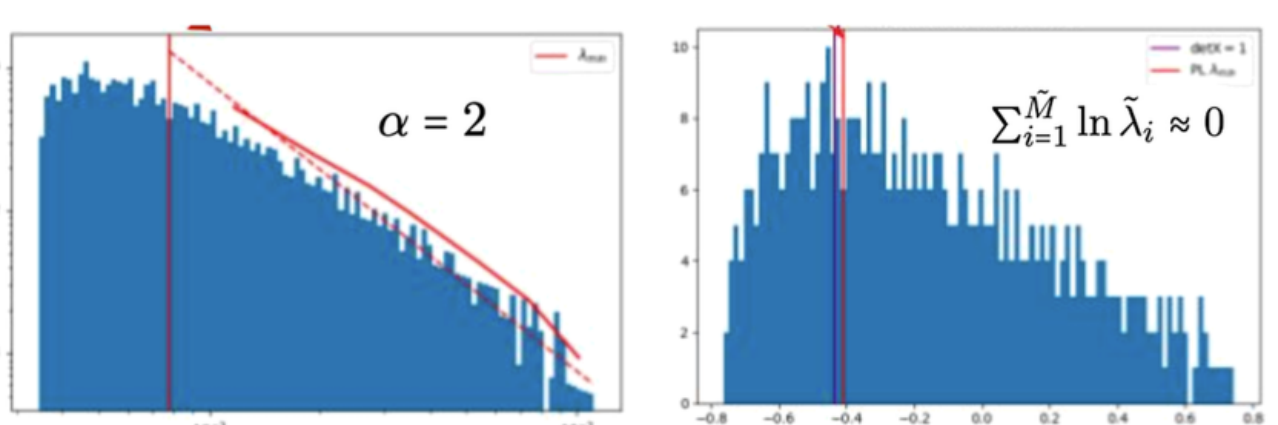}
\caption{Empirical Spectral Density (ESD) of a well-trained dense layer. Panel (a) shows the ESD of $\X=N^{-1}\W^T\W$ in log-log coordinates. The tail is fit by a finite-window MLE/KS power law with $\alpha\simeq2$. Panel (b) shows the trace-log-normalized ECS, where $\Tr\log\Xtilde_{\Rell}\simeq0$
  ($\Rell$ denotes the ECS space).  The red and purple vertical lines denote the PL tail, $\tilde{\lambda}^{+}_{\rm PL}$, and the ECS,
  $\tilde{\lambda}^{+}_{\Rell}$, resp.
  Plots have been generated using the open-source~\WW~tool.\cite{WeightWatcherTool}}
\label{fig:spectral}
\vspace{-0.5em}
\end{figure}

In many well-trained models, the layers have an ESD with a finite-window tail, starting at $\PLstart$, that can be
well fit to a power law (PL):
\begin{equation}
\rho(\lambda)\sim\lambda^{-\alpha},\quad\lambda\ge\PLstart
\label{eq:rho_tail}
\end{equation}
HTSR uses this tail, together with related power-law metrics, as a data-free diagnostic of learned correlation structure \cite{MartinMahoneyICML2019,MartinMahoneyJMLR2021,MartinPengMahoney2021}. In practice, the power-law exponent is estimated with the standard Clauset--Shalizi--Newman maximum-likelihood fit
which finds the best $\PLstart$, 
as implemented in the open-source WeightWatcher tool \cite{WeightWatcherTool,Clauset2009}.
The observed optimum is not simply the smallest exponent. Strong checkpoints often sit near
\begin{equation}
\alpha\simeq2.
\label{eq:alphaobs}
\end{equation}
Figure~\ref{fig:spectral} shows a representative fitted spectrum.

\begin{figure}[b]
\centering
\includegraphics[width=0.48\columnwidth]{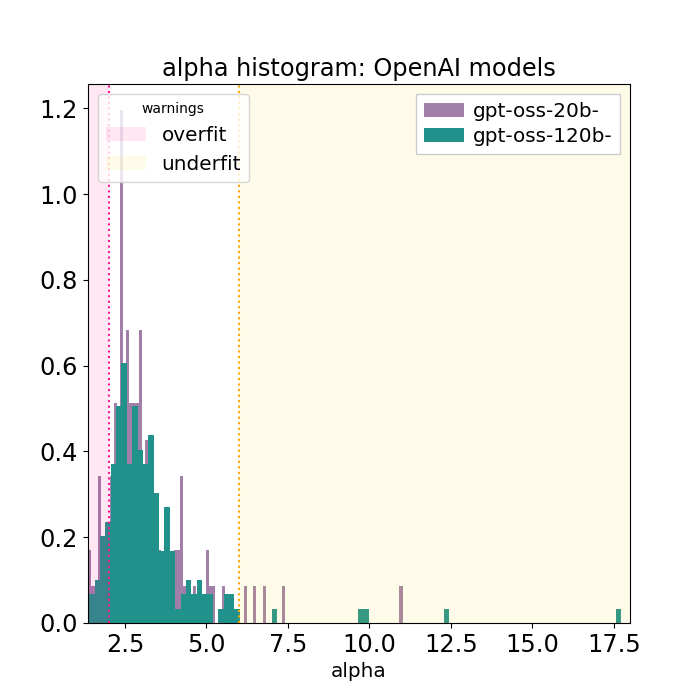}\hfill
\includegraphics[width=0.48\columnwidth]{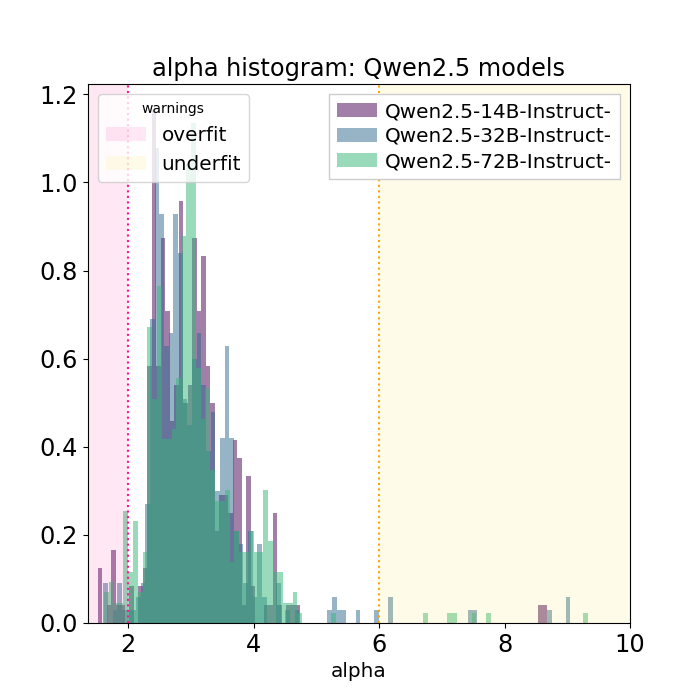}
\caption{Layerwise HTSR $\alpha$ histograms from linear layers, computed with the open-source \WW{} tool. Left: GPT-OSS 20B/120B models. Right: Qwen2.5 instruct models. The warning bands are heuristic: small $\alpha$ marks overfit or trap risk, while large $\alpha$ marks weakly correlated or underfit tails. Optimizer notebook details are listed in Appendix~\ref{app:optimizer_experiments}.}
\label{fig:alpha_histograms}
\vspace{-0.6em}
\end{figure}

The diagnostic is practical. The open-source \WW{} tool estimates the layerwise HTSR PL exponents $\alpha$ from trained weights alone. One can obtain PL fits for both linear and convolutional layers (see \cite{MartinMahoney2020SDM}).
Prior work showed that these data-free spectral metrics can predict model-quality trends without access to training or testing data \cite{MartinPengMahoney2021}. Figure~\ref{fig:alpha_histograms} shows recent example scans for the
OpenAI GPT-OSS 20B and 120B models~\cite{OpenAIGPTOSS2025},
and the instruct component $(\W:=\W_{\rm instruct}-\W{\rm base})$~of the Qwen2.5 instruct models.
Each histogram summarizes fitted layerwise $\alpha$ values across all layer weight matrices.

HTSR is phenomenological. It extends the familiar Marchenko-Pastur (MP)  Random Matrix Theory (RMT) in a unique way to the layer weight matrices to classify them into several distinct heavy-tailed (HT) universality classes.
HT RMT was first noted in the theoretical physics literature in the 1990s that when a random matrix is heavy-tailed power-law (HT-PL) elementwise, the ESDs no longer follow the MP law and themselves become HT-PL and fall into different universality classes~\cite{CizeauBouchaud1994,PottersBouchaud2020}. HTSR generalized this, and noted that when ESDs themselves are HT-PL, they fall into different universality classes even when the underlying matrix is not HT-PL elementwise.  Moreover, HTSR notes that  $\alpha\simeq 2.0$ is special, and corresponds to a candidate optimal state of the layer. Roughly, then, for our purposes here, we can reduce the $5+1$ HTSR classes into three phases: $\alpha>2.0$,  $\alpha\simeq 2.0$, and  $\alpha< 2.0$.

Still, HTSR is phenomenological. The central question here seeks a more precise but also simple answer: why does $\alpha\simeq 2.0$ correspond to an optimal trained layer with a stable HT-PL empirical spectral density?

Here, we develop an RG argument for  $\alpha\simeq 2.0$ that is a posteriori consistent with HTSR (and SETOL), but goes further by integrating HTSR and SETOL; there are two main arguments. First, $\alpha=2$ is the boundary where generalization is  supported by the
maximum number of eigencomponents
before overfitting.
Second, $\alpha=2$ is what is called marginal in Wilsonian Renormalization Group.
But first, to formulate the RG theory, we need to understand its motivation in the SETOL approach.

\section{SETOL: the Wilsonian ERG }
\label{sxn:setol}
\enlargethispage{1.5\baselineskip}

%\begin{tcolorbox}[marginalbox,title=\textbf{2. Phase-map summary}]
%The best NN layers align the HTSR $\alpha=2$ with the SETOL trace-log law and ECS.
%\begin{itemize}
%\item HTSR identifies the power-law tail with $\alpha=2$.
%\item SETOL identifies the ERG trace-log condition
%\item Strong checkpoints make the two windows align.
%\item Overfitting appears as traps or $\alpha<2$.
%\end{itemize}
%\takeaway{This alignment is a proposed RG signature.}
%\end{tcolorbox}

SETOL is a \emph{Semi-Empirical Theory of Learning}.
It connects the layer empirical spectral density (ESD) to multilayer neural-network generalization accuracy using statistical mechanics,
providing a rigorous foundation\footnote{At a physics level of rigor.} for HTSR.
It is semi-empirical because it takes the layer ESD as empirical input and derives a layer-quality functional
whose diagnostics can be associated with the HTSR PL metric $\alpha$.
In doing so, it assumes that a NN, near the optimal state of learning, satisfies a volume-preserving, scale-invariant transformation
(the trace-log condition) that resembles taking a single step of the Wilsonian Exact Renormalization Group (ERG)~\cite{WilsonKogut1974,Polchinski1984}.

SETOL starts by approximating the average generalization accuracy (one minus the error) of a model,
which it calls the average model \emph{Quality} $\overline Q_{\rm NN}$. 
\begin{equation}
\overline Q_{\rm NN}\simeq 1-\overline E_{\rm gen}^{\rm NN}.
\label{eq:setol_quality}
\end{equation}
(where the average is over all data samples).
For a multilayer model, SETOL then approximates total model average quality by a product of layer contributions,
\begin{equation}
\overline Q_{\rm NN}\simeq\prod_{\rm layers}\overline Q_\ell .
\label{eq:setol_layer_product}
\end{equation}
Thus a layer quality $\overline Q:=\overline Q_\ell$ estimates how much each layer supports the model's generalization accuracy.

Layer quality has a simple meaning here. It is one layer's contribution to generalization accuracy. A useful layer lowers error. A bad layer adds unstable gain or loses useful capacity.
This is the practical meaning of ``layer quality'' throughout the paper.

SETOL derives the layer quality using the layer-quality (squared) generating function,
\begin{equation}
\begin{aligned}
\overline Q^{\,2}
&=\frac{1}{\beta}\frac{\partial}{\partial n}
  \left[\lim_{N\gg1}\IZQ\right] \\
&\simeq_{\rm high\text{-}T}
  \frac{1}{n}\frac{\partial}{\partial\beta}
  \left[\lim_{N\gg1}\IZQ\right] .
\end{aligned}
\label{eq:setol_q_derivative}
\end{equation}
Here $n$ is the sample count, and $\beta$ is the inverse-temperature parameter.
This is evaluated in the thermodynamic large-$n$ (and large-$N$) limit.
This generating function $\IZQ$ is essentially an effective layer free energy.

SETOL models each trained layer using a matrix generalization of the classical student--teacher perceptron~\cite{SeungSompolinskyTishby1992} to matrices.
The trained weight matrix plays the role of the fixed teacher, $\mathbf T=\mathbf W$, while $\mathbf S$ denotes a possible student matrix that reproduces the teacher-layer outputs.
SETOL then generalizes the ST perceptron free energy in terms of the student correlation operator $\widetilde{\mathbf A}$. The compatible matrices $\widetilde{\mathbf A}_N$ and $\widetilde{\mathbf A}_M$ denote the induced $N\times N$ and $M\times M$ forms of this operator, restricted to the Effective Correlation Space (ECS). In an annealed, high-temperature approximation, the resulting effective free energy,
or quality (squared) generating function $\IZQ$, takes the form of a random matrix (HCIZ) integral
\begin{equation}
\begin{aligned}
\IZQ
&=\frac{1}{N}\log\int d\mu(\widetilde{\mathbf A})
   \exp\{N\mathcal S(\widetilde{\mathbf A})\},\\
\mathcal S(\widetilde{\mathbf A})
&=n\beta\,\Tr\!\left[
   \frac{1}{N}\mathbf T^T\widetilde{\mathbf A}_{N}\mathbf T
   \right]
   +\frac{1}{2}\Tr\log(\widetilde{\mathbf A}_{M}) .
\end{aligned}
\label{eq:setol_hciz}
\end{equation}
The first term measures teacher--student alignment: it favors student correlations that match the trained teacher layer. The second term is the log-volume contribution that appears when the theory changes variables from weights to correlations. In the thermodynamic limit, and under the SETOL RG ansatz, this free energy depends only on the empirical spectral density of the teacher covariance.
Thus the layer's contribution to generalization accuracy reduces to a functional of the trained weight matrix's ESD.

\begin{tcolorbox}[marginalbox,title=\textbf{SETOL free energy and ERG trace-log}]
  Matrix-generalized statistical mechanics student-teacher model of NN generalization accuracy:
  \eqtight{
       1-\overline E_{\rm gen}^{\rm NN}\simeq
     \overline Q_{\rm NN}\simeq\prod_\ell \overline Q_\ell}
\begin{itemize}
\item \textbf{Layer Quality $\overline Q_\ell$:} one layer's contribution to total accuracy (1-error). Depends simply on $\X$.
\item \textbf{Effective Free Energy} $\IZQ$ estimates $\overline Q_\ell$ with a random-matrix HCIZ integral
  \item \textbf{Gauge fixing:} remove volume drift with
\eqtight{
  \Tr\log\Xtilde_{R}=0
  \qquad\Longleftrightarrow\qquad
  \det\Xtilde_{R}=1 .
}
\end{itemize}
\takeaway{SETOL says: relate the HTSR power-law $\alpha$ to the generalization accuracy by evaluating $\IZQ$}
\end{tcolorbox}

\begin{figure*}[t]
\centering
\safeincludegraphics[width=0.98\textwidth]{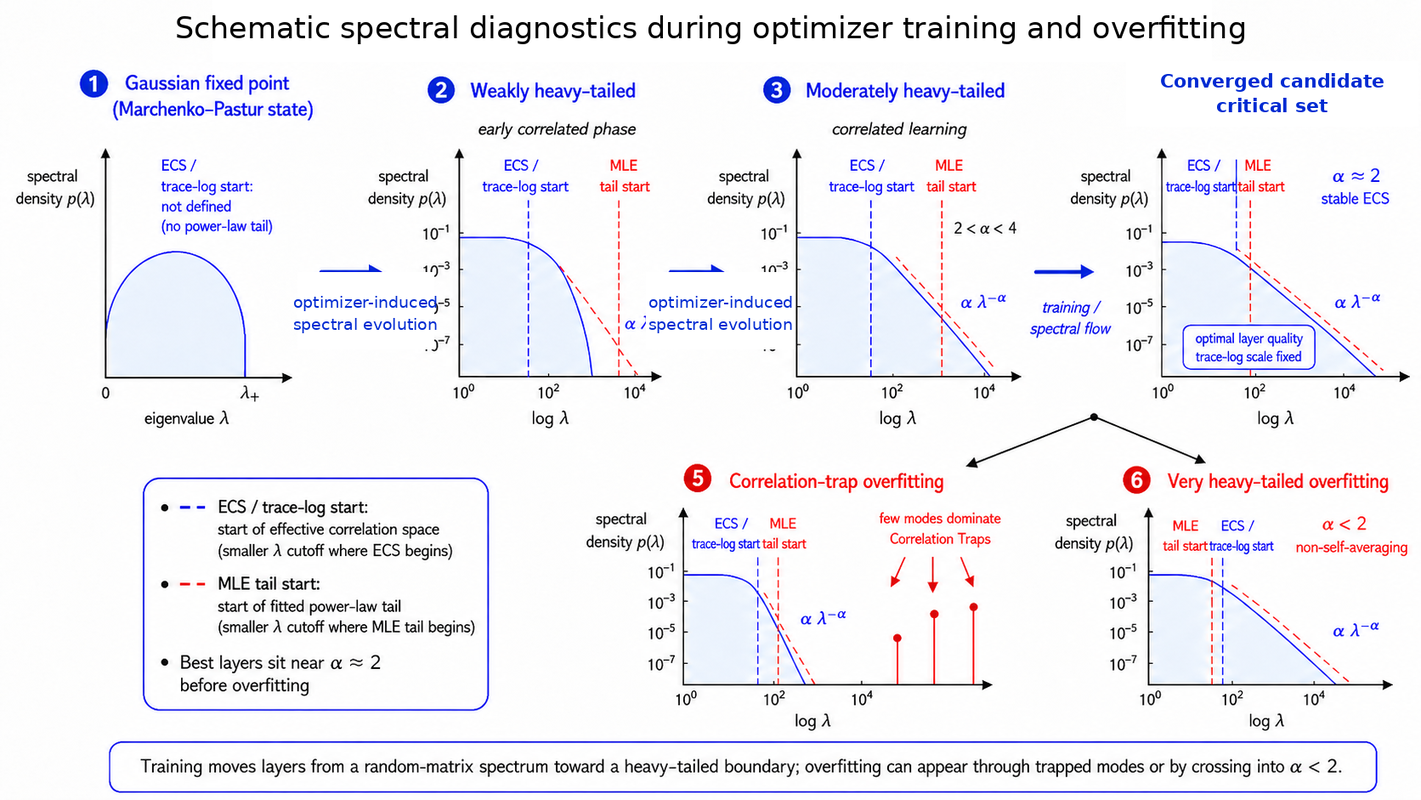}
\caption{
HTSR/SETOL phases during training. 
The HTSR power-law tail $\mathcal T_{\rm PL}$ and the SETOL Effective Correlation Space $\Rell_{\rm ECS}$ are selected independently.
During training, they overlap. At convergence, they align
$\Rell_{\rm ECS}^{\star}\simeq \mathcal T_{\rm PL}^{\star}$, at $\alpha\simeq2$.
}
\label{fig:spectral_flow_schematic}
\end{figure*}

The Wilsonian ERG step in SETOL is a change of variables and a change of scale. The bare layer variables are the student weights. The effective variables are the retained correlation modes in $\Atilde_{\Rell}$. The SETOL paper describes this step as an Effective-Hamiltonian construction: a bare layer Hamiltonian for $\overline Q^2$ is rewritten as a renormalized effective Hamiltonian on the ECS \cite{MartinHinrichsSETOL2025}.
\begin{equation}
 d\mu(\mathbf S)\,e^{n\beta N\Tr\mathcal H_{\overline Q^2}}
 \ \xrightarrow{\rm ERG}\
 d\mu(\widetilde{\mathbf A})\,
 e^{n\beta N\Tr\mathcal H_{\overline Q^2}^{\rm ECS}} .
 \label{eq:setol_RG_step}
\end{equation}
The arrow here is a scale-invariant, volume-preserving transformation, and is
effectively taking a single step of the Wilsonian Exact RG (ERG).
It is from this convenient construction that the ERG trace-log condition emerges.
Here, the optimal ECS flow should approach a stable scale-balanced state near $\alpha\simeq2$.

%For this work, which considers the learning dynamics under a more complete RG lens,
%the ERG trace-log condition removes volume drift while retaining scale.

At each checkpoint, we first normalize the full spectrum to unit mean,
\begin{equation}
\Tr{\X}=M
\end{equation}.
We call this \emph{First-Moment Rescaling}.

Then,  SETOL defines a volume-preserving, scale-invariant transformation 
on the student $\A_{\Rell}$ and teacher $\X_{\Rell}$ correlation matrices
\begin{equation}
\Tr\log\widetilde{\mathbf A}_{\Rell}=0 \leftrightarrow \Tr\log\Xtilde_{\Rell}\simeq0 .
\label{eq:setol_trlog_AX}
\end{equation}.
where  $\Xtilde_{\Rell}$ denotes the restriction of the normalized operator to the selected ECS.
This is also called \emph{Gauge Fixing}, and we may say we are now on the \emph{Gauge Slice}.
%We may drop $\Rell$ on $\Xtilde$ or $\tilde{\lambda}$ if the retained ECS is clear.

The layer quality (squared) is now expressed
simply in terms of the tail of the ``teacher'' ESD 
using the integrated R-Transform $R_{\Xtilde}(z)$~\cite{Zee1996}:
\begin{equation}
\overline{Q}^{2}
=
\sum_{\tilde{\lambda}_{i}\in\Rell}
\int_{\ECSstart}^{\tilde{\lambda}_{i}}
R_{\Xtilde}(z)\, dz
\label{eq:setol_Q}
\end{equation}
where $R_{\Xtilde}(z)$ is an analytic function chosen to model the PL tail
(and is analogous to a renormalized self-energy)
and $\ECSstart$ is the start of the ECS.
This result shows that the layer quality depends only on the ESD
(as in HTSR).

The link between HTSR and SETOL is summarized in Figure~\ref{fig:spectral_flow_schematic}, which should be read as a schematic spectral flow rather than as a static taxonomy. The red vertical line marks the start of the HTSR power-law tail, $\TPL$, selected by the MLE fit, while the purple vertical line marks the start of the SETOL Effective Correlation Space, $\RECS$, selected by the ERG trace-log condition.  Training (usually) begins in a Gaussian/Marchenko--Pastur regime where neither boundary is stable, then a weak heavy-tail forms, correlated learning develops, and the SETOL ECS boundary and HTSR PL-tail boundary move toward one another. At the convergence point, $\alpha \simeq 2$ and $\Tr\log\Xtilde_{\Rell}=0$; the two selected regions then align, $\Rell_{\rm ECS}^{*} \simeq \mathcal T_{\rm PL}^{*}$, equivalently $\tilde{\lambda}^{+}_{\Rell}-\tilde{\lambda}^{+}_{\rm PL}\simeq 0$. Empirically, this is also the regime in which the truncated-SVD approximation, restricted to the ECS, preserves test/generalization accuracy. The red lower panels show the two failure modes beyond this boundary: a correlation trap, where isolated rank-one perturbations in $\W$ distort the ESD and induce overfitting, and a very-heavy-tailed regime, where $\alpha < 2$ and the system  becomes strongly (over) correlated and also overfits. At the boundary is optimal learning and generalization.

SETOL suggests that the best layers in a very well trained model sit at a Wilsonian Exact RG fixed point.
The open question is how to formulate the full RG spectral flow, justifying the trace-log condition and  $\alpha=2$ as optimal.

%%%\begin{strip}\vspace{-0.35em}
%%%\centering
%%%\safeincludegraphics[width=0.98\textwidth]{figures/fig3_spectral_flow.png}
%%%\captionof{figure}{Schematic spectral flow for HTSR/SETOL training and overfitting. The Gaussian branch begins with a Marchenko--Pastur-like spectrum and no fitted power-law tail. Training then moves the layer through weakly and moderately heavy-tailed regimes. Red dashed lines mark the MLE/KS power law cutoff. Purple dashed lines mark the trace-log ECS cutoff. Strong checkpoints occur when the cutoffs nearly align near $\alpha\simeq2$. Overfitting can appear through correlation traps or by crossing into the very-heavy-tailed $\alpha<2$ regime.}
%%%\label{fig:spectral_flow_schematic}
%%%\vspace{-0.35em}\end{strip}
%%%
\clearpage
\section{The self-averaging boundary}
\label{sxn:selfavg}

It is a basic principle of statistics that reliable predictions require samples that are representative of the underlying population. The simplest counterexample is estimating the mean of a heavy-tailed power-law distribution. In such a distribution, a small number of extreme observations can dominate the average, so different finite samples may have very different sample means. A more subtle example arises in complex systems near a critical point. There, even when the ensemble mean is well defined, long-range correlations can keep sample-to-sample fluctuations large, so different large samples do not necessarily become representative of one another. In statistical mechanics, these ideas are formalized by the concept of \emph{Self-Averaging}.
\begin{tcolorbox}[
  marginalbox,
  title=\textbf{Self-averaging versus non-self-averaging}
]
Self-averaging asks if relative sample-to-sample fluctuations $\mathcal{V}_M$ disappear
as the system grows.
\begin{itemize}
\item \(\mathcal{V}_M(\mathcal{O})\to0\): a bound can be formed
\item \(\mathcal{V}_M(\mathcal{O})\not\to0\): no shrinking variance bound
\end{itemize}
\takeaway{The self-averaging boundary is identified here where relative fluctuations $\mathcal{V}_M$ cease to vanish.}
\end{tcolorbox}

Self-averaging means that a macroscopic quantity
measured in one large sample becomes close to its average over many samples as
the system size $M$ grows (i.e concentrates).. 
%
%
%In statistical  physics,, this is frequently stated in terms of the 
%data-averaged partition function $Z$ as
%\begin{equation}
%\left\langle\log\mathcal Z_M\right\rangle_{\xi_M}
%\simeq
%\log\left\langle\mathcal Z_M\right\rangle_{\xi_M} .
%\end{equation}
%This is a strong form of self-averaging; more on this below.
%
Let $\mathcal{O}_M$ be a macroscopic observable measured in a system
of size $M$, and let the mean over $M$ samples $\xi_M$ be
\begin{equation}
\mu_M:=\mathbb{E}[\mathcal{O}_M]=\langle\mathcal{O}\rangle_{\xi_M}
\end{equation}
We say that $\mathcal{O}_M$ is self-averaging when it concentrates around $\mu_M$ as
the system size grows. When the second moment is finite, its relative
sample-to-sample variance is
\begin{equation}
\mathcal V_M(\mathcal O)
:=
\frac{\operatorname{Var}(\mathcal O_M)}
{\left|\mathbb E[\mathcal O_M]\right|^2}
=
\mathbb E\!\left[
\left|
\frac{\mathcal O_M}{\mu_M}-1
\right|^2
\right].
\label{eq:relative_variance}
\end{equation}
A standard sufficient test for self-averaging is
$\mathcal{V}_M(\mathcal{O})\to0$ as $M\to\infty$. When
$\mathcal{V}_M(\mathcal{O})$ does not vanish, relative fluctuations persist and
large samples can remain different.
This is \emph{Non-Self-Averaging} in the relative-variance sense.

\begin{figure}[!b]
\centering
\safeincludegraphics[width=\columnwidth]
{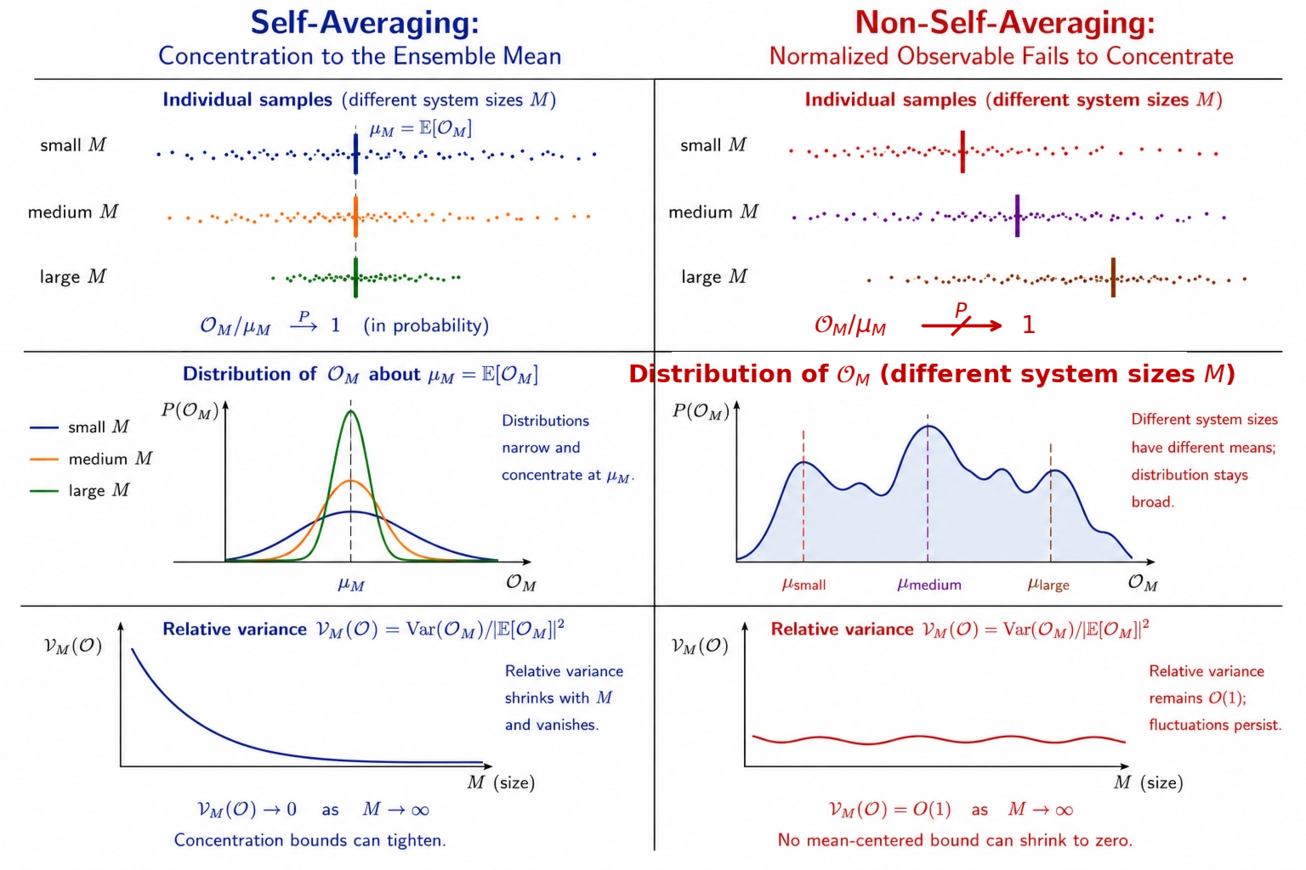}
\caption{\textbf{Self-averaging and the failure of a shrinking bound.}
Here $\mu_M=\mathbb{E}[\mathcal{O}_M]$. In the self-averaging case, the normalized
observable $\mathcal{O}_M/\mu_M$ concentrates around one and the relative variance
$\mathcal{V}_M(\mathcal{O})$ tends to zero, allowing concentration bounds to tighten.
In the illustrated non-self-averaging case, finite relative deviations retain
nonzero probability as the system grows, so no valid upper bound on that
probability can shrink to zero.}
\label{fig:self_average_schematic}
\end{figure}

In NNs, non-self-averaging arises
from two different mechanisms: localized correlation traps, and 
long-range collective non-vanishing fluctuations.
In either case, a model may describe its training sample well while
performing poorly on out-of-sample test data.

We propose that  the strongest learning occurs at the
boundary between self-averaging and non-self-averaging.
And this is not specific to NNs but a fundamental
property of all related learning algorithms.

But first, it is helpful to understand self-averaging
from different theoretical points of view:
Statistical Learning Theory (SLT), Statistical Mechanics (StatMech),
and Random Matrix Theory (RMT).

\subsection{Statistical Learning view}
The connection to Statistical Learning Theory (SLT) here is qualitive; it
is depicted in Figure~\ref{fig:self_average_schematic}.

For a fixed tolerance $\epsilon>0$, consider the probability that one sample differs
appreciably from the ensemble mean,
\begin{equation}
p_M(\epsilon)
:=
\Pr\!\left\{
\left|
\frac{\mathcal{O}_M}{\mu_M}-1
\right|>
\epsilon
\right\}.
\end{equation}
A useful concentration bound has the form
\begin{equation}
p_M(\epsilon)\leq B_M(\epsilon),
\qquad
B_M(\epsilon)\longrightarrow0
\quad\text{as}\quad M\longrightarrow\infty .
\end{equation}
Such a bound says that a sufficiently large sample is likely to be close to
the ensemble mean.

On the self-averaging side, the relative variance
$\mathcal V_M(\mathcal{O})$ decreases, so the distribution narrows and the bound can
tighten. More data or greater system size then produces a more stable and
representative answer.
If
%$\liminf_{M\to\infty} V_M(O) = 0$,
$\mathcal{V}_M(O) \to 0$
this bound shrinks to zero for every
fixed $\epsilon>0$ since
\begin{equation}
p_M(\epsilon)
\leq
\frac{\mathcal V_M(\mathcal O)}{\epsilon^2}.
\label{eq:relative_chebyshev}
\end{equation}
 A sufficiently large sample is then increasingly likely
to lie close to the ensemble mean.

On the non-self-averaging side, write $D_M:=\mathcal O_M/\mu_M-1$
and $V_M:=\mathcal V_M(\mathcal O)$. Suppose $V_M\geq v_0>0$ and
$\mathbb E[D_M^4]\leq K V_M^2$ for sufficiently large $M$, with fixed
$v_0,K$. Splitting at $|D_M|=\epsilon_0:=\sqrt{v_0/2}$ and applying
Cauchy--Schwarz gives
\[
p_M(\epsilon_0)\geq
\frac{(V_M-\epsilon_0^2)^2}{\mathbb E[D_M^4]}
\geq\frac{1}{4K}>0.
\]
The moment condition excludes variance carried solely by increasingly rare
events. Thus no valid bound $B_M(\epsilon_0)$ can vanish: increasing system
size does not restore concentration.

%The HTSR exponent $\alpha\simeq2$ is the empirical candidate for this
%boundary. Values $\alpha<2$ and correlation traps are treated as warning
%signals that the layer has crossed into the non-self-averaging regime.
\clearpage

\subsection{Statistical Mechanics view}
\label{sxn:selfavg_statmech}

In statistical physics, self-averaging is frequently stated in terms of the
data-averaged partition function $\mathcal Z_M$ as
\begin{equation}
\left\langle\log\mathcal Z_M\right\rangle_{\xi_M}
\simeq
\log\left\langle\mathcal Z_M\right\rangle_{\xi_M}.
\label{eq:selfavg_partition}
\end{equation}
Here $\xi_M$ labels one independently prepared sample of size $M$, and
$\langle\cdot\rangle_{\xi_M}$ denotes an average over many such samples.
Equation~\eqref{eq:selfavg_partition} is a strong form of self-averaging:
a single large sample gives nearly the same free-energy contribution as the
sample average.

To expose the fluctuation correction, define
\begin{equation}
\delta_M
:=
\tfrac{
\mathcal Z_M-\left\langle\mathcal Z_M\right\rangle_{\xi_M}
}{
\left\langle\mathcal Z_M\right\rangle_{\xi_M}
},
\qquad
\left\langle\delta_M\right\rangle_{\xi_M}=0,
\label{eq:relative_partition_fluctuation}
\end{equation}
so that
\begin{equation}
\mathcal V_M(\mathcal Z)
:=
\tfrac{
\operatorname{Var}_{\xi_M}(\mathcal Z_M)
}{
\left\langle\mathcal Z_M\right\rangle_{\xi_M}^{2}
}
=
\left\langle\delta_M^2\right\rangle_{\xi_M}.
\label{eq:relative_variance_partition}
\end{equation}
Whenever the expansion is controlled,
\begin{equation}
\begin{aligned}
\left\langle\log\mathcal Z_M\right\rangle_{\xi_M}
&=
\log\left\langle\mathcal Z_M\right\rangle_{\xi_M}
+
\left\langle\log(1+\delta_M)\right\rangle_{\xi_M}
\\
&=
\log\left\langle\mathcal Z_M\right\rangle_{\xi_M}
-\tfrac12\mathcal V_M(\mathcal Z)
+\tfrac13\left\langle\delta_M^3\right\rangle_{\xi_M}
-\cdots.
\end{aligned}
\label{eq:annealed_quenched_expansion}
\end{equation}
Thus the difference between the two sides of
Eq.~\eqref{eq:selfavg_partition} consists of sample-to-sample fluctuation
terms. If the relative variance and higher normalized fluctuations vanish,
the two expressions agree to leading order. If they remain finite, one large
sample need not represent the full collection of samples, which may degrade generalization.

\paragraph{SETOL and the Spectral Energy.}
Section~\ref{sxn:setol} gives the corresponding effective free energy of layer quality (squared)
generating function $\IZQ$, Eq.~\eqref{eq:setol_q_derivative}. We use this
and ,Eq.~\eqref{eq:setol_Q} to write the
ECS-averaged quality (squared)  $\tilde{Q}^{2}=\overline{Q^2}/M_{\Rell}$,
now averaged over the retained ECS $\Rell$:
\begin{equation}
\tilde{Q}^{2}
=
\frac{1}{M_{\Rell}}\sum_{\tilde{\lambda}_{i}\in\Rell}G_{\Xtilde}(\tilde{\lambda}_{i}),
\;\;
G_{\Xtilde}(\tilde{\lambda})
:=
\int_{\ECSstart}^{\tilde{\lambda}}R_{\Xtilde}(z)\,\dd z,
\label{eq:setol_spectral_functional}
\end{equation}
where $R_{\Xtilde}$ is the $R$-transform of the retained spectrum.
Its local free-cumulant expansion is
\begin{equation}
R_{\mathbf X}(z)
=
\kappa^{(f)}_1
+\kappa^{(f)}_2z
+\kappa^{(f)}_3z^2
+\cdots.
\label{eq:rtransform_cumulants}
\end{equation}
Writing
\begin{equation}
m_p
:=
\frac{1}{M_{\Rell}}\Tr{\Xtilde_{\Rell}^{p}}
=
\int\tilde{\lambda}_{\Rell}^p\widehat\rho_{\Rell}(\tilde{\lambda}_{\Rell})\,\dd\tilde{\lambda}_{\Rell},
\label{eq:spectral_moments}
\end{equation}
To simplify the analysis, let $\ECSstart\simeq 0$; evaluating $\hat{Q}^{2}$ gives
\begin{equation}
\tilde{Q}^2
=
\kappa^{(f)}_1m_1
+\tfrac{1}{2}\kappa^{(f)}_2m_2
+\tfrac{1}{3}\kappa^{(f)}_3m_3
+\cdots 
\label{eq:setol_moment_expansion}
\end{equation}
Retaining only the first-cumulant term, $\kappa^{(f)}_1=m_1$, ECS-average layer quality is
$ \tilde{Q} \simeq  m_1$.
Define
\begin{equation}
\widetilde E_{\Rell}
:=m_1
\simeq
\frac{1}{M_{\Rell}}\Tr{\Xtilde_{\Rell}}
=
\int\tilde{\lambda}_{\Rell}\widehat\rho_{\Rell}(\tilde{\lambda}_{\Rell})\,\dd\tilde{\lambda}_{\Rell}.
\label{eq:trace_law_energy}
\end{equation}
%Notice that we choose $\Tr{\X}=M$,
%but once we are on the gauge slice and have imposed the trace-log condition,
%$\Tr{\Xtilde_{\Rell}}\ne M$ (see~\ref{sxn:meanlog_gauge} below).
This is the \emph{Spectral Energy}. It is the first relevant macroscopic observable
for our RG power-counting analysis.

\paragraph{Ising-model analogy and the self-averaging test.}
A similar structure appears in the Ising model. Let $\mathcal M$
denote the total magnetization and let a field $h$ enter through
$\mathcal H(h)=\mathcal H_0-h\mathcal M$. Then $\log\mathcal Z_M(h)$ is the
cumulant-generating function for $\mathcal M$:
\begin{equation}
\log\frac{\mathcal Z_M(h)}{\mathcal Z_M(0)}
=
\sum_{n\geq1}
\frac{(\beta h)^n}{n!}\,\kappa_n(\mathcal M).
\label{eq:ising_cumulant_expansion}
\end{equation}
The first field-dependent term is
$\beta h\,\kappa_1(\mathcal M)=\beta h\langle\mathcal M\rangle$, and the
magnetization density is
\begin{equation}
m_M
:=
\frac{\langle\mathcal M\rangle_{\beta}}{M}
=
\frac{1}{\beta M}
\frac{\partial}{\partial h}\log\mathcal Z_M(h).
\label{eq:ising_magnetization_first_cumulant}
\end{equation}
where the brackets $\langle\rangle_{\beta}$ denote the ordinary thermal average within one fixed sample.
Thus the total magnetization is the first
cumulant, and $m_M$ is its leading additive density generated from the Ising
free energy. In the same sense,
$\widetilde E_{\Rell}=m_1$ is the first additive spectral quantity
generated by the SETOL expansion. It is therefore the first quantity to test
for concentration:
\begin{equation}
\mathcal V_{M_{\Rell}}(\widetilde E_{\Rell})
\underset{?}{\longrightarrow} 0.
\label{eq:energy_selfavg_condition}
\end{equation}
In the disordered Ising model and related random critical systems, the
sample-to-sample relative width of the magnetization and susceptibility does
not necessarily narrow at the critical point. Instead it can remain finite as
the system grows, so these observables are non-self-averaging
\cite{WisemanDomany1995,aharony1996,wiseman1998}.
The analogy here is that if $\widetilde E_{\Rell}$ does not self-average, then the leading SETOL
contribution already depends on the particular training sample or run.
\paragraph{Why criticality defeats averaging.}
The sample-to-sample fluctuations above compare independently prepared
systems; they are not the ordinary thermal fluctuations inside one fixed
sample. Away from criticality, the correlation length $\ell_{\rm corr}$ is
finite, so a $d$-dimensional system of linear size $L$ contains roughly
$(L/\ell_{\rm corr})^d$ nearly independent correlated regions, whose
differences average away. At a disordered critical point,
$\ell_{\rm corr}\sim L$, so the effective number of independent regions can
remain $O(1)$. Increasing the system size then need not reduce relative
fluctuations. This is one reason why magnetization and other singular observables can
remain non-self-averaging at criticality.

For the retained spectral energy,
\begin{equation}
\operatorname{Var}_{\xi_M}(\widetilde E_{\Rell})
=
\frac{1}{M_{\Rell}^{2}}
\sum_{i,j\in\Rell}
\operatorname{Cov}_{\xi_M}(\tilde{\lambda}_{\Rell,i},\tilde{\lambda}_{\Rell,j}).
\label{eq:spectral_energy_covariance}
\end{equation}
If the eigenvalue contributions fluctuate nearly independently, the
off-diagonal terms are small and the average can concentrate. If many modes
fluctuate coherently, the covariance terms add rather than cancel, and the
relative variance can remain finite even when the spectrum contains many
modes. $\widetilde E_{\Rell}$ plays a similar diagnostic role as magnetization
for critical, non-self-averaging behavior.

Correlation traps, however, provide a second, different
route to non-self-averaging, even at criticality; we discuss this in the next subsection.

\clearpage

\clearpage
\subsection{Random Matrix Theory view}
\label{sxn:selfavg_rmt_v1}

Random Matrix Theory gives a direct finite-matrix test of the self-averaging
boundary in the form of \emph{Correlation Traps}.
Starting from a trained layer, we randomize its entries elementwise,
\begin{equation}
  \mathbf W \longrightarrow
  \mathbf W^{\pi}:=\operatorname{rand}(\mathbf W),
  \label{eq:rmt_elementwise_randomization_v1}
\end{equation}
by flattening the matrix, randomly permuting all $NM$ entries, and reshaping
it.  This preserves the exact weight multiset and its one-point statistics,
but removes the learned arrangement of those weights.  For the classical
spectral test, $\mathbf W^{\pi}$ is therefore uncorrelated and effectively
i.i.d:
\begin{equation}
  W_{ij}^{\pi}
  \overset{\mathrm{eff.\ i.i.d.}}{\sim}
  \widehat{\rho}_{W}.
\end{equation}

The observable is the full ESD of
$\mathbf X^{\pi}=N^{-1}(\mathbf W^{\pi})^T\mathbf W^{\pi}$.  For a large
finite-variance matrix, this ESD approaches the deterministic
Marchenko--Pastur (MP) law~\cite{MarchenkoPastur1967}; MP is the
self-averaging baseline.  It is the large-$N$ limit ($Q=\tfrac{N}{M}$ fixed)
of the mean distribution $\rho^{\pi}(\lambda)$ for the ESD of $\mathbf W^{\pi}$.
For a typical (i.e. light-tailed) i.i.d. rectangular matrix, all the eigenvalues will
lie within the MP bulk, to within finite-size Tracy--Widom (TW)
fluctuations~\cite{TracyWidom1996,BaikBenArousPeche2005}.

\begin{figure}[!h]
\centering
\safeincludegraphics[width=0.78\columnwidth]{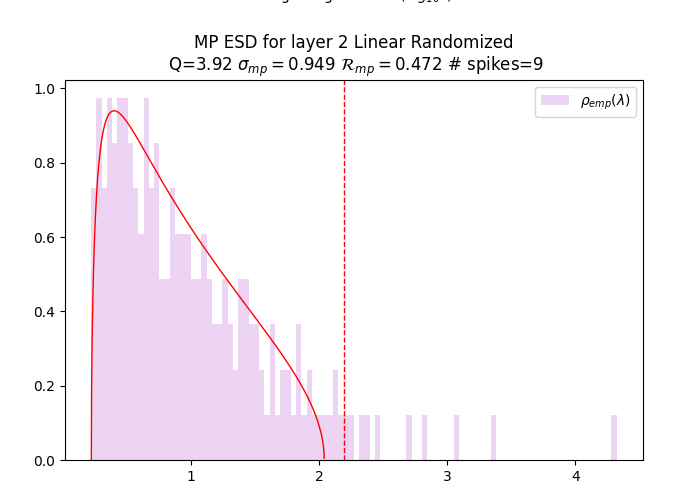}
\caption{\textbf{The randomized MP test.}
After $\mathbf W\to\operatorname{rand}(\mathbf W)$, the bulk follows the MP
fit.  The dashed line marks the far bulk plus TW edge $(\MPstart+\Delta_{\rm TW})$;
persistent eigenvalues to its right are counted as correlation traps.}
\label{fig:rmt_v1_mp}
\end{figure}

If an eigenvalue $\lambda_{\rm trap}$ appears beyond
the MP bulk edge $\MPstart$ plus the TW scale $\Delta_{\rm TW}$, we call these
outliers \emph{Correlation Traps}.
\begin{equation}
\lambda_{\rm trap} > \MPstart+\Delta_{\rm TW}
\end{equation}  
Figure~\ref{fig:rmt_v1_mp} illustrates several correlation traps.

A finite number of traps may leave the limiting MP bulk intact, since each
outlier carries mass $1/M$.  The violation is instead localized at the edge:
the largest $\lambda$ are no longer explained by ordinary finite-size
fluctuations.

\begin{figure}[!t]
\centering
\safeincludegraphics[width=\columnwidth]{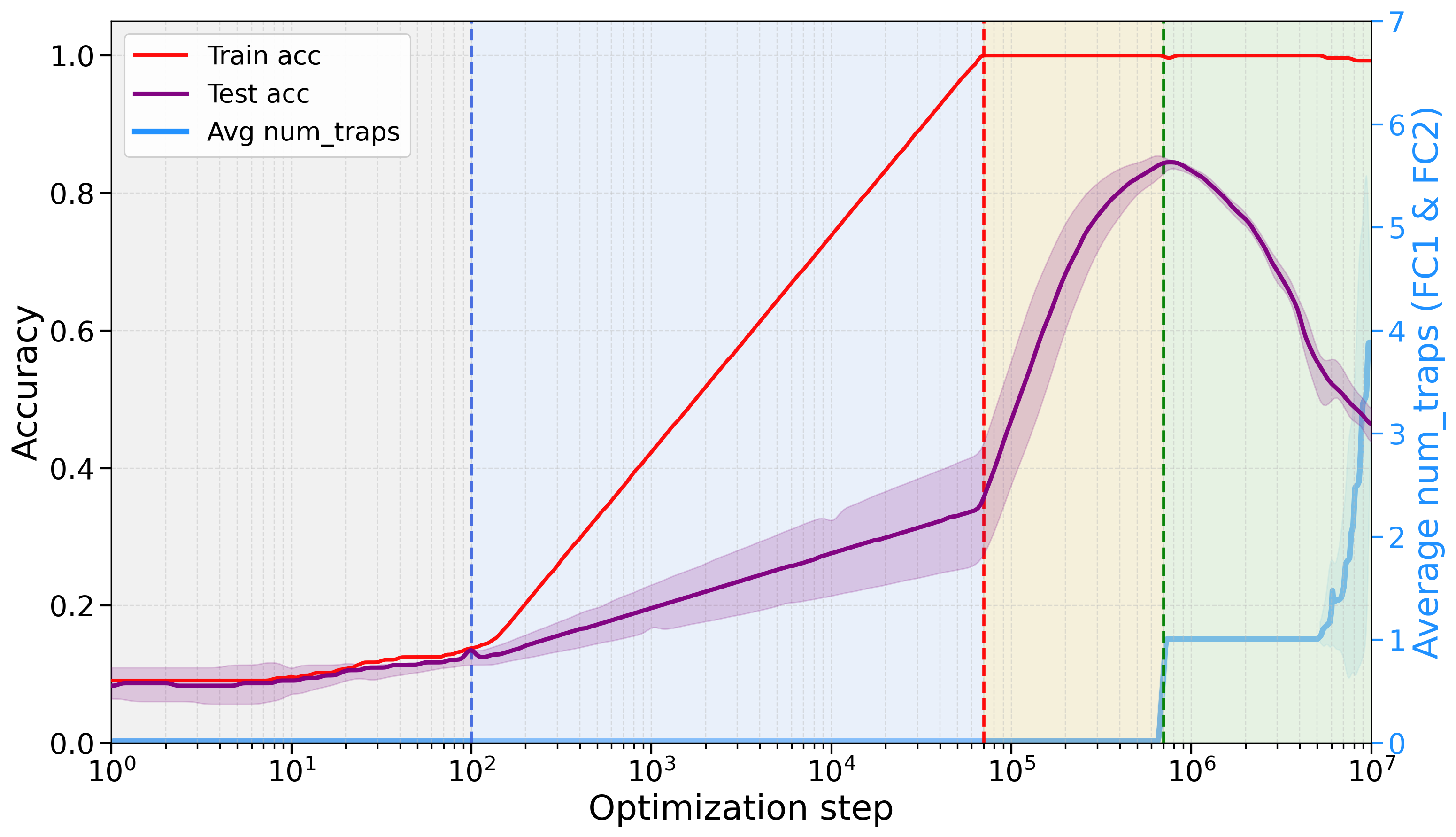}
\vspace{-0.30em}
\safeincludegraphics[width=\columnwidth]{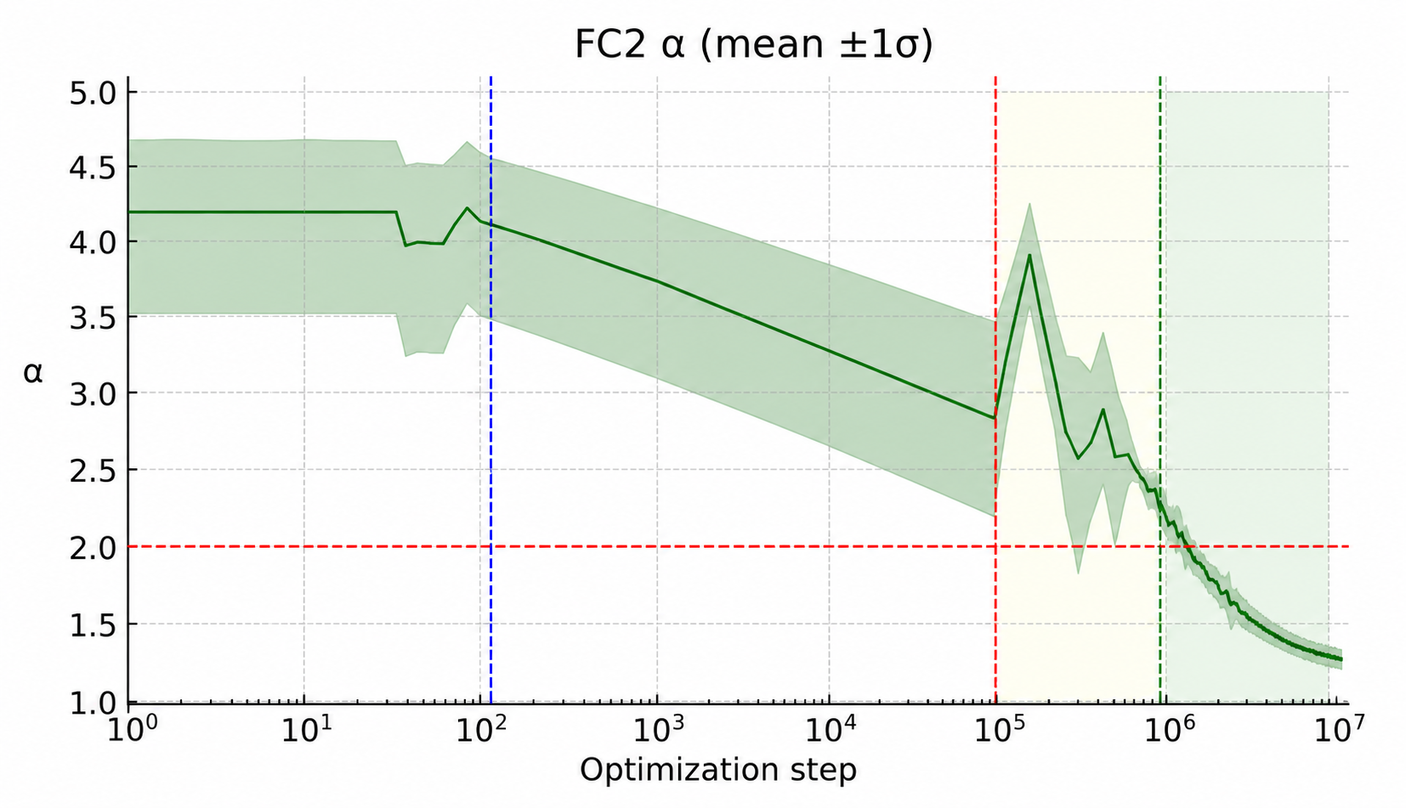}
\caption{\textbf{Long-horizon grokking.}
Top: test accuracy falls as the randomized-spectrum trap count rises.
Bottom: the fitted exponent subsequently crosses below $\alpha=2$.  The
aligned horizontal axes show the order of the warnings
\cite{PrakashMartinTraps2026,PrakashMartinLateCollapse2026}.}
\label{fig:rmt_v1_long}
\end{figure}

Figure~\ref{fig:rmt_v1_long} (top panel) depicts results from a long-horizon grokking experiment,
where a three-layer MLP is trained on a subset of MNIST until it induces
pre-grokking memorization, then good performance (grokking), and finally overfitting
(anti-grokking)\cite{PrakashMartinTraps2026,PrakashMartinLateCollapse2026}.
The red line denotes the training accuracy, which becomes perfect at about $10^5$
training steps.  The purple line denotes the test accuracy, which increases to
optimal at about $10^6$ steps, and then drops sharply.
The blue line counts the number of correlation traps in the system
(in all three layers). The traps become non-zero as soon as the system test accuracy drops
at $10^6$ steps, and increase with decreasing test accuracy.

The bottom panel displays the average layer $\alpha$ for the FC1 and FC2 layers.
Notice that $\alpha\rightarrow 2$ as the performance increases, and $\alpha<2$
after first correlation trap forms and the test error drops.
The drop in average $\alpha$ is late because the traps distort the ESD
and the PL fits.

These results indicate that conventional overfitting leaves a
spectral record in the weights.  Training accuracy remains high, but held-out
accuracy degrades as persistent MP-edge anomalies accumulate.  Traps are often
seen with aggressive learning rates or prolonged training.  Early stopping
and weight decay can help, but neither directly checks whether the randomized
ESD has returned to MP.

The trap test is different from HTSR.  Traps are measured on
$\operatorname{rand}(\mathbf W)$; HTSR is measured on the original correlated
$\mathbf W$ and treats its broad non-MP ESD as learned structure
\cite{MartinMahoneyJMLR2021}.  A trap can distort the upper ESD and damage the
power-law fit, sometimes contributing to an apparent $\alpha<2$.  Conversely,
a smooth heavy-tailed ESD can cross below two without an isolated trap.
Thus collective heavy-tailed fluctuations and localized correlation traps are
distinct mechanisms of non-self-averaging.

\clearpage
\begin{strip}\vspace{-0.55em}
\centering
\safeincludegraphics[width=0.95\textwidth]{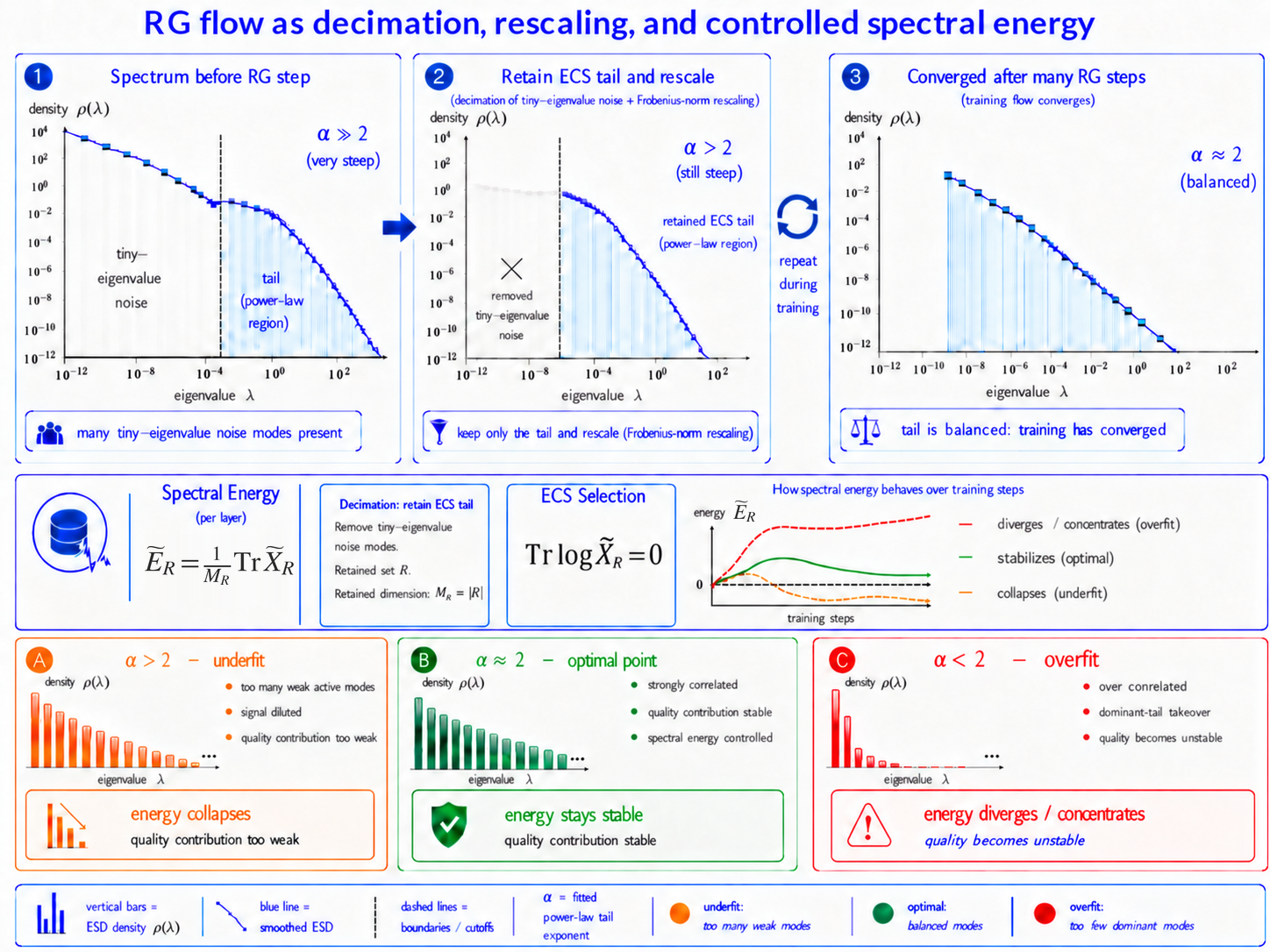}
%\captionof{figure}{Operational RG flow for one dense layer.   decimates tiny-eigenvalue noise and retains an ECS tail $\Rell$. The retained eigenvalues are $\tilde{\lambda}_{\Rell,i}$, and the retained dimension is $M_{\Rell}=|\Rell|$. Each retained ECS is selected ERG trace-log condition.}
  \captionof{figure}{Operational RG flow for one dense layer. After first-moment normalization of $\X$, the ERG trace-log condition selects the ECS tail $\Rell$ without further rescaling. The retained eigenvalues are $\tilde{\lambda}_{\Rell,i}$, and the retained dimension is $M_{\Rell}=|\Rell|$.}
\label{fig:rg_flow_energy}
\vspace{-0.75em}\end{strip}

\section{Wilsonian RG Flow}
\label{sxn:rg}
\vspace{0.35em}
\begin{tcolorbox}[marginalbox,title=\textbf{Optimizer-induced RG flow}]
Wilsonian RG describes the flow of the layer ESD(s) induced by the optimizer steps  $\Opt[\circ]$
\eqtight{ \mathbf{W}_{t+1}=\Opt[\mathbf{W}_t],
  \qquad
  \mathcal{RG}[\rho({\lambda})]\longrightarrow \rho_{\rm tail}^{*}({\lambda})
}

\takeaway{The best layers approach an RG fixed point $\Ferg$.}
\end{tcolorbox}

%\begin{tcolorbox}[marginalbox,title=\textbf{ Optimizer-induced RG flow}] 
%  RG describes the flow of the spectral density
%  \eqtight{\mathcal{R}_{\mathrm{RG}}:\rho(\lambda)\longrightarrow \rho_{\rm tail}^{*}(\lambda).}
%  \takeaway{The RG flow encodes how the spectral density $\rho(\lambda)$ approaches a candidate fixed point,
%    satisfying both the HTSR $\alpha\simeq 2$ and the SETOL ERG trace-log.}
%\end{tcolorbox}

This section describes a Wilsonian RG construction for the
spectral density $\rho({\lambda})$ as the heavy-tailed power-law density forms
and approaches the HTSR $\alpha=2$ candidate optimal state under the SETOL ERG trace-log condition.
It is the shape, not the scale, that regulates the final layer quality and therefore the final model generalization accuracy.
The retained layer should help accuracy improve, or equivalently help error decrease,
without making the layer contribution explode or collapse. 
First-moment normalization removes arbitrary scale.
The RG construction shows when the spectral shape collapses, diverges, or converges.

The goal here  is to identify an effective description of a trained
layer--the Effective Correlation Space (ECS)--by repeatedly removing
small-scale spectral components, retaining the dominant
correlation structure, and eliminating redundant overall scale.
Rather than develop a formal RG map, we describe the
construction operationally using a truncated-SVD picture
together with explicit ERG trace-log renormalization. The resulting
flow acts on the retained spectral density (in the ECS)
and motivates the HTSR $\alpha=2$ as the boundary between
learning and overfitting.

Figure~\ref{fig:rg_flow_energy} depicts the RG operational training flow which is generated by the training dynamics
of the underlying optimizer on each step: 
$\mathbf{W}_{t+1}=\Opt[\mathbf{W}_t]$.
In practice, the RG flow can be seen as a series of truncated-SVD steps, called decimations.
Each step keeps the dominant spectral directions that contribute to the generalization accuracy,
while dropping directions that do not matter at the current scale.
Each retained layer is selected according to the trace-log law.
The best performing models have layers that approach the SETOL ERG fixed point $\Ferg$,
converging to an ESD with HTSR PL exponent $\alpha\simeq 2$, neither overfitting $(\alpha<2)$
nor underfitting $(\alpha>2)$ their training data.

\subsection{HTSR phases and RG}
\label{sxn:htsr_rg}

HTSR provides the empirical starting point for the RG analysis.
Section~\ref{sxn:htsr} described trained layer spectra using HTSR heavy-tailed (HT)  universality classes.
Here we use those classes as a coarse description of training flow.
We do not start with a microscopic model of the data, the optimizer, or the architecture.
We start with the observation that trained layer ESDs often develop HT power-law spectra.

\begin{center}
\small
\setlength{\tabcolsep}{2.5pt}
\renewcommand{\arraystretch}{1.06}
\captionof{table}{HTSR classes interpreted as an RG flow.}
\label{tab:htsr_rg_flow}
\begin{tabularx}{\linewidth}{@{}lcc@{}}
\toprule
Phase & Exponent & RG meaning \\
\midrule
Random-like / Bulk+Spikes & no PL fit & Gaussian branch \\
Weakly HT & $\alpha\gg2$ & weak-energy sector \\
Moderately HT & $\alpha>2$ & approach to boundary \\
Strongly HT & $\alpha=2$ & balanced boundary \\
Very HT / traps & $\alpha<2$ & dominant-tail sector \\
\bottomrule
\end{tabularx}
\end{center}

Table~\ref{tab:htsr_rg_flow} labels the HTSR phases, and Figure~\ref{fig:rg_flow_energy} shows how a layer moves through them. Training starts with many tiny-eigenvalue modes and a weak or steep tail, then retains an ECS tail and moves the ESD toward a more structured heavy-tailed spectrum.
The spectral energy $\Ered$ from Eq.~\eqref{eq:trace_law_energy} gives this motion its learning meaning:
for $\alpha>2$ the retained energy is weak and decreases across successive spectral bands (orange panel),
near $\alpha\simeq2$ it remains balanced across scales (green panel),
and for $\alpha<2$ the tail modes can carry too much energy and drive the overfit regime (red panel)
where the energy diverges.

%%%Figure~\ref{fig:rg_flow_energy} shows this flow operationally.
%%%The top row begins with many tiny-eigenvalue modes and a weak or steep tail.
%%%Training trims the small modes, retains an ECS tail, and moves the ESD toward a more structured heavy-tailed spectrum.
%%%In the favorable case, the fitted exponent decreases toward $\alpha\simeq2$ and the retained spectrum reaches the balanced panel on the right.
%%%If the flow continues past this boundary, the layer can enter the red overfit regime, where a few dominant tail modes control the energy.
%%%
%%%The spectral energy $\Ered$ from Section~\ref{sxn:tracelog} gives these phases a simple meaning.
%%%For $\alpha>2$, the retained energy is spread over weak tail modes and can collapse, as in the orange panel of Figure~\ref{fig:rg_flow_energy}.
%%%Near $\alpha\simeq2$, the retained energy stays balanced across scales, as in the green panel.
%%%For $\alpha<2$, a few dominant modes can carry too much energy, as in the red panel.
%%%
\begin{figure}[h]
\centering
\includegraphics[width=\columnwidth]{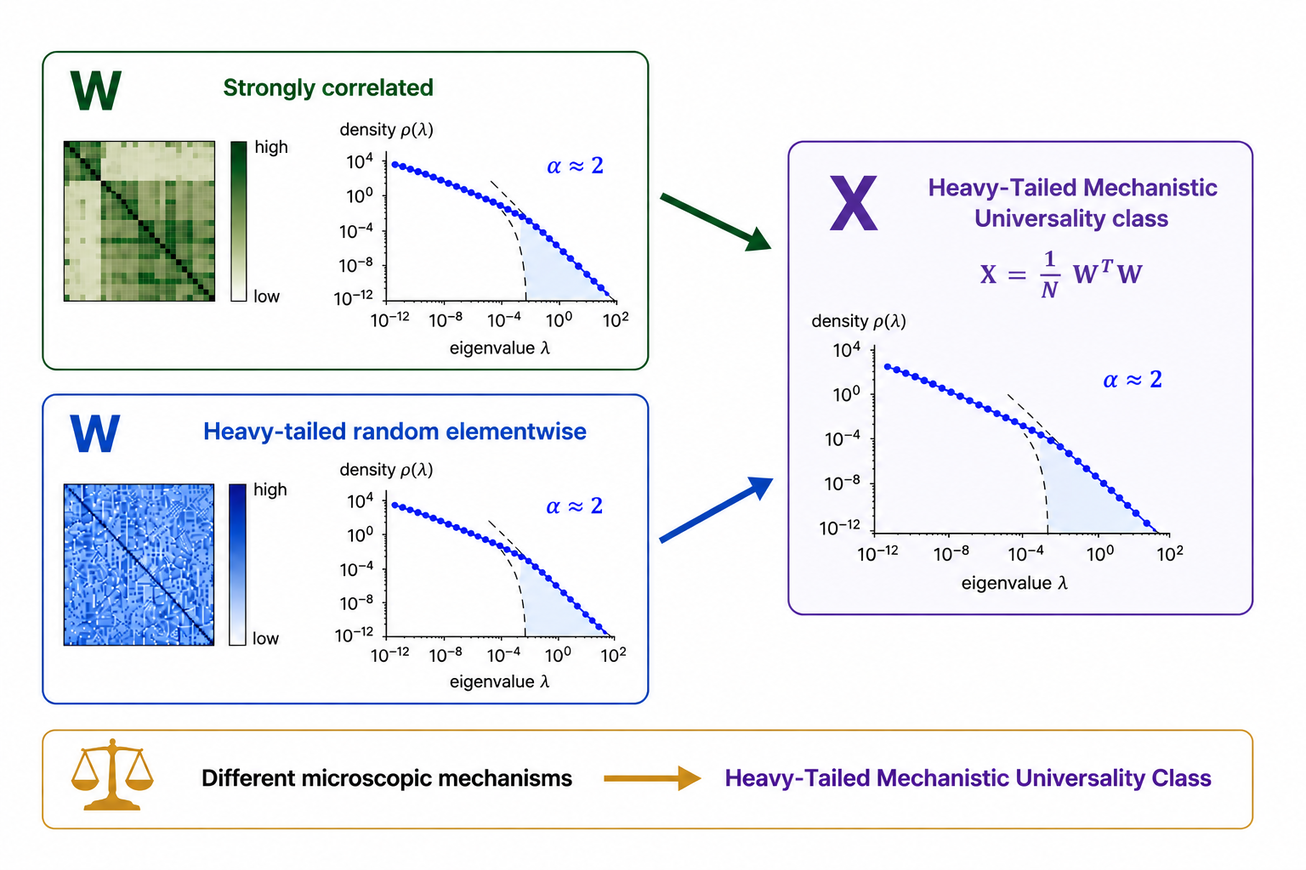}
\caption{\textbf{Heavy-Tailed Mechanistic Universality.}}
\label{fig:ht_mechanistic_universality}
\vspace{-0.6em}
\end{figure}

Figure~\ref{fig:ht_mechanistic_universality} illustrates what HTSR calls \emph{Mechanistic Universality}.
Different architectures, datasets, and optimizers can produce the same heavy-tailed spectral endpoint.
This universality is spectral, not elementwise.
It is a property of the ESD of the correlation matrix $\X=N^{-1}\W^T\W$, not a consequence of
the distribution of entries $W_{ij}$. 
Indeed, the entries are generally not HT (unless the layer is overfit); learned correlations can make the spectrum heavy-tailed.
Here we interpret this mechanistic universality as a preview of RG universality; SETOL provides the bridge.

\newpage

\subsection{SETOL and RG}
\label{sxn:setol_rg}

SETOL contributes the scale convention for the RG analysis.
Section~\ref{sxn:setol} introduced the layer quality generating function, an effective free energy  $\IZQ$
and showed that the Wilsonian ERG construction leads to the ERG trace-log condition.
The present paper does not need to evaluate the full SETOL quality $\overline{Q}$.
Instead, we use the ERG trace-log condition to define the ECS,
and we study the flow of the retained spectral energy $\Ered$.

\begin{tcolorbox}[marginalbox,title=\textbf{SETOL $\rightarrow$ RG energy flow}]
\small
\textbf{Spectral Energy:}
\[
\Ered=\frac{1}{\Mred}\Tr\Xtilde_{\Rell}
=\int \tilde{\lambda}_{\Rell} \rhohat_{\Rell}(\tilde{\lambda}_{\Rell})\,d\tilde{\lambda}_{\Rell} .
\]
\textbf{Gauge fixing:}
\[
\Tr\log\Xtilde_{\Rell}=0 .
\]
\textbf{Truncated SVD is exact}
\[
\W \rightarrow \W_{\Rell}=\U\mathbf{\Sigma}_{\Rell}\Vt
\]\[
\mathbf{\Sigma}_{\Rell}:=\diag[0,\cdots,0,\sigma_{1},\sigma_{2},\cdots\sigma_{M_{\Rell}}],\quad\sigma_{i}\sim\sqrt{\lambda_{i}}
\]
\takeaway{RG studies how the spectral components concentrate.}
\end{tcolorbox}

Figure~\ref{fig:rg_flow_energy} shows RG decimation, rescaling, and the resulting energy flow.
The  scale is set by normalizing by the First-moment.
The ERG trace-log informs the decimation step.
We then study how the retained spectral energy $\Ered$ flows
Eq.~\eqref{eq:trace_law_energy} defined this energy as the first moment of the retained ECS.
The RG question is therefore concrete: after normalizing $\X$ and selecting the ECS,
how does $\Ered$ flow across spectral scales?

We answer this question by using the trace-log to measure covariance volume. On the retained positive spectrum,
\begin{equation}
\Tr\log\X_{\Rell}
=
\log\det\X_{\Rell}
=
\sum_{i\in\Rell}\log\lambda_i .
\end{equation}
Thus the trace-log measures the logarithm of the retained covariance volume. It does not add eigenvalue gains directly;
it measures their product scale.
Once fixed, the natural comparison for the RG flow will be across fixed-ratio bands $[\Lambda,b\Lambda]$,
or equivalently across equal-width intervals in $\log\tilde{\lambda}_{\Rell}$;
this lets us define a \emph{Spectral Band Energy}, $G_E(\Lambda,b)$, below.

At the proposed Wilsonian ERG fixed point, the layer weight matrix $\W$ can be replaced
by its Truncated SVD form, $\W_{\Rell}=\W_{\Rell}=\U\mathbf{\Sigma}_{\Rell}\Vt$, where
$\mathbf{\Sigma}_{\Rell}$ is the diagonal operator of $M_{\Rell}$ singular values $(\sigma_{i})$
(properly normalized) defined on the retained ECS, and zero for the other  $M-M_{\Rell}$ ones.
Doing this, SETOL shows the reduced model faithfully reproduces the test accuracy;
the layer generalizing components fully concentrate into the ECS.

%We therefore study how much retained layer energy lies in each spectral band. Define the band energy gain
%\begin{equation}
%G_E(\Lambda,b)
%=
%\int_{\Lambda}^{b\Lambda}
%\lambda\,\widehat\rho_{\Rell}(\lambda)\,d\lambda .
%\end{equation}
%This is the portion of the retained spectral energy carried by the logarithmic band $[\Lambda,b\Lambda]$. If $G_E(\Lambda,b)$ decreases as the band moves to larger eigenvalues, the retained tail is weak. If it increases, the layer is moving toward dominant-tail concentration. If it remains comparable across bands, the spectrum is scale-balanced.

The next subsection shows that the HTSR power-law tail $\rho_{\rm tail}(\tilde{\lambda})$ makes the energy gain stable at $\alpha=2$.
That is, we illustrate that $\alpha$ is analogous to an RG universal critical exponent.

\clearpage

\subsection{Operational view of the RG flow}
\label{sxn:effdescr}

Figure~\ref{fig:rg_flow_energy} depicts the spectral RG flow for one layer.  The flow turns the ideas from Sections~\ref{sxn:htsr_rg} and~\ref{sxn:setol_rg} into a repeatable operation: decimate tiny-eigenvalue noise, retain an ECS tail, impose the SETOL ERG trace-log gauge, and compare the remaining spectral shape.  Section~\ref{sxn:htsr_rg} supplies the HTSR input: the retained ESD should approach a power-law tail, with the strongest checkpoints often near $\alpha\simeq 2$.  Section~\ref{sxn:setol_rg} supplies the SETOL input: generalization accuracy is controlled by the layer ESD through the ECS and the trace-log condition.  Section~\ref{sxn:selfavg} supplies the risk test: on the SETOL-normalized retained ECS, the retained spectral energy should self-average instead of concentrating in a few dominant tail modes.  This subsection makes that control flow operational.

%Figure~\ref{fig:rg_flow_energy} separates the RG step into four actions.  First, the step removes tiny-eigenvalue noise that does not contribute to the retained correlation structure.  Second, it keeps the ECS tail and represents the layer by the truncated matrix $\W_{\Rell}$.  Third, it evaluates the retained operator on the first-moment recaling for the whole ESD, and selects the ECS using $\Tr{\log \Xtilde_{\Rell}}=0$.
%When this trace-log condition is exact, the ECS is exact.
%Fourth, it tests the remaining spectral energy.

Figure~\ref{fig:rg_flow_energy} illustrates spectral decimation and the evolution of retained spectral energy during training. At each checkpoint, we first normalize the full ESD by its first moment, then select the ECS using $\Tr{\log \Xtilde_{\Rell}}=0$ on that fixed scale. This selection removes the tiny-eigenvalue noise shown in the figure and defines the truncated matrix $\W_{\Rell}$; when the trace-log condition holds exactly, the selected ECS is exact. We then evaluate the retained spectral energy and track its evolution during training. In this proposed RG view, optimization supplies the flow and effective rescaling, while the diagnostic selects the ECS without further rescaling.
%The point is not to reproduce SGD or any microscopic optimizer.
This RG view of the training flow  describes the learning process as a coarse-graining induced by the  optimizer trajectories
by considering the linearized spectral view at each layer.

The three regimes in Figure~\ref{fig:rg_flow_energy} give the operational meaning of underfit, balanced, and overfit.  In the orange regime, $\alpha>2$ and the tail is steep.  The retained layer may self-average, but the spectral energy can be too weak to carry enough learned correlation.  In the green regime, $\alpha\simeq2$ and the retained spectral energy stays balanced across logarithmic spectral scales.  This is the stable layer-quality regime.  In the red regime, $\alpha<2$ or correlation traps appear.  The spectral energy diverges in the RG flow, the layer quality can become strongly non-self-averaging, and
the generalization accuracy degrades.

SETOL suggests that the RG flow induces a truncated layer $\W$.  SETOL maps layer quality, and therefore generalization accuracy, to the layer ESD.  Once quality depends on the retained spectrum, not on every microscopic weight $W_{i,j}$, an effective description can replace the full layer by a retained ECS matrix.  At the fixed point, SETOL predicts that the ECS aligns with the HTSR power-law tail and that the truncated-SVD layer $\W_{\Rell}$ preserves the test/generalization contribution of the original layer.  This is the reason the operational RG step in Figure~\ref{fig:rg_flow_energy} is a truncated-SVD step.

The RG flow acts on retained spaces and retained operators together.  At step $(n)$, the effective object is
\begin{equation}
    \left(\Rell^{(n)},\,\X_{\Rell}^{(n)}\right),
    \qquad
    \Mred^{(n)}=|\Rell^{(n)}| .
    \label{eq:flowpair_operational_v5}
\end{equation}
A decimation step changes $\Rell^{(n)}$ by removing eigencomponents outside the current correlated description.
the ECS is selected by imposing the trace-log gauge condition on the (normalized) retained correlation operator:  $\Tr{\log\Xtilde^{(n)}}=0$

SETOL then selects the $\ECSstart$, the start of the ECS, based
on the trace-log condition 
\begin{equation}
  \sum_{\tilde{\lambda}_{\Rell}} \log \tilde{\lambda}_{\Rell} = 0,\quad\tilde{\lambda}_{\Rell}\ge\ECSstart
\end{equation}
At optimality the ECS starts at the PL tail, $\tilde{\lambda}^{+}_{\Rell}=\tilde{\lambda}^{+}_{\rm PL}$.
When $\alpha<2$, the ECS can be smaller than the PL, with  $\ECSstart > \PLstart$.
This indicates that a few tail eigencomponents can dominate.
Simultaneously, the PL tail can extend deep into previous bulk region,
indicating that the spectral fluctuations may not concentrate (as in Eq.~\ref{eq:spectral_energy_covariance}).
These two mechanisms are central to understanding why $\alpha<2$ is a signature of overfitting.

The ECS arises implicitly during the optimization step as the learning mechanism effectively
compresses the model at each layer, causing the generalizing components of each $\W$ to flow
into a lower-rank representation.  But it is unclear how much compression is enough,
and how much is too much.
But because the typical optimizer is not aware of the RG flow,
the ECS may be too weak and the spectral energy may collapse.
Or, the layer may be overfit to the training data, causing the spectral energy to diverge
because it concentrated in a few dominant modes, or many strongly fluctuating modes.

The retained ESD is the shape variable defined  by the RG step.  Formally, we can write
\begin{equation}
    \rhohat_{\Rell}(\tilde{\lambda}_{\Rell})
    =
    \frac{1}{\Mred}\sum_{i\in\Rell}\delta(\tilde{\lambda}_{\Rell}-\tilde{\lambda}_{\Rell,i}),
    \qquad
    \int\rhohat_{\Rell}(\tilde{\lambda}_{\Rell})\,d\tilde{\lambda}_{\Rell}=1 .
    \label{eq:prob_esd_operational_v5}
\end{equation}
In practice, however, we use the discrete retained spectral energy (on the gauge slice)
\begin{equation}
    \widetilde E_{\Rell}
    =
    \frac{1}{\Mred}\Tr\Xtilde_{\Rell}
    =
    \int \tilde{\lambda}_{\Rell}\,\rhohat_{\Rell}(\tilde{\lambda}_{\Rell})\,d\tilde{\lambda}_{\Rell} .
    \label{eq:retained_trace_energy_operational_v5}
\end{equation}
Section~\ref{sxn:selfavg} showed why this energy matters: it provides a layer-level quantity that can self-average when many weakly correlated retained eigencomponents contribute, but (again) can fail to self-average in two  ways. First, a small number of dominant modes can control the energy, as in correlation traps. Second, and more importantly here, collective correlations among the weights can keep the relative fluctuations of the spectral energy at order one across the entire ECS. This behavior is marginal at $\alpha=2$ and becomes increasingly tail-dominated for $\alpha<2$, where the largest eigencomponents can carry an $O(1)$ fraction of the total spectral energy. Although the boundary at $\alpha=2$ may therefore be marginally non-self-averaging, the spectral energy grows only logarithmically with system size.  This is shown below, however, first we would like to show that the SETOL trace-log condition naturally emerges
through  a scale-invariant RG flow defined by the unit Jacobian, the physically relevant RG transformation.

\clearpage
\subsection{Scale Invariant Flow}
An RG flow requires a series of scale-invariant decimation maps which here are characterized by the trace-log condition on the retained ECS space
during the training flow.
The trace-log comes from the Jacobian of the retained coordinate map.  Figure~\ref{fig:retained_jacobian_operational_v5} isolates the retained coordinate map used in the ERG trace-log gauge step.

In this subsection, $\W_{\Rell}$ denotes
\begin{equation}
    \W_{\Rell}:=\W\V_{\Rell},
    \label{eq:retained_coordinate_weight_map_operational_v5}
\end{equation}
where $\V_{\Rell}$ contains the retained right singular vectors, and $\W_{\Rell}$ is $N\times \Mred$ (not the zero-padded).
\begin{equation}
    \X_{\Rell}
    :=
    \V_{\Rell}^{T}\X\V_{\Rell}
    =
    \diag(\lambda_i)_{i\in\Rell} .
    \label{eq:retained_correlation_operator_operational_v5}
\end{equation}
After selecting $\Rell$, let  $a\in\mathbb R^{\Mred}$ be
\begin{equation}
    y(a)=\frac{1}{\sqrt N}\W_{\Rell}a .
    \label{eq:retained_coord_map_operational_v5}
\end{equation}
The ordinary coordinate Jacobian is therefore
\begin{equation}
    \frac{\partial y}{\partial a}
    =
    \frac{1}{\sqrt N}\W_{\Rell} .
    \label{eq:retained_coord_jacobian_operational_v5}
\end{equation}
This Jacobian is not a training-time derivative such as $d\W/dt$ or $d\X/dt$.  It is the derivative of retained output coordinates with respect to retained input coordinates.

\begin{center}
\safeincludegraphics[width=\columnwidth]{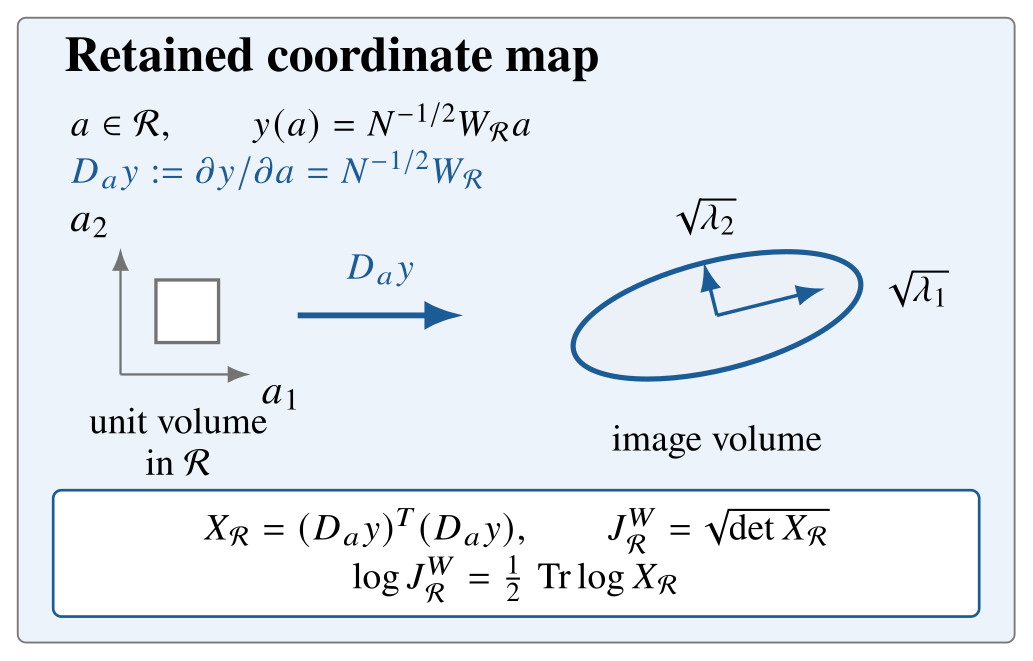}
\captionof{figure}{Retained coordinate Jacobian.  After the ECS/SVD decimation selects $\Rell$, the retained layer sends coordinates $a\in\mathbb R^{\Mred}$ to $y(a)=N^{-1/2}\W_{\Rell}a$.  The ordinary coordinate Jacobian is $\partial y/\partial a=N^{-1/2}\W_{\Rell}$.  Its raw image volume is $J_{\Rell}^{W}=\sqrt{\det\X_{\Rell}}$.}
\label{fig:retained_jacobian_operational_v5}
\end{center}

The determinant appears because the retained map changes volume.  Since $\W_{\Rell}$ may be rectangular, we do not take an ordinary determinant of $\W_{\Rell}$.  We take the induced $\Mred$-dimensional volume factor.  The squared volume factor is the Gram determinant,
\begin{equation}
    \det\!\left[
    \left(\frac{1}{\sqrt N}\W_{\Rell}\right)^T
    \left(\frac{1}{\sqrt N}\W_{\Rell}\right)
    \right]
    =
    \det\X_{\Rell} .
    \label{eq:gram_det_operational_v5}
\end{equation}
Equivalently,
\begin{equation}
    \left(\frac{1}{\sqrt N}\W_{\Rell}\right)^T
    \left(\frac{1}{\sqrt N}\W_{\Rell}\right)
    =
    \V_{\Rell}^{T}\X\V_{\Rell}
    =
    \X_{\Rell} .
    \label{eq:retained_gram_operator_operational_v5}
\end{equation}
Therefore the raw retained volume Jacobian is
\begin{equation}
    J_{\Rell}^{W}
    =
    \sqrt{\det\X_{\Rell}}
    =
    \prod_{i\in\Rell}\sqrt{\lambda_i} .
    \label{eq:retained_volume_jacobian_operational_v5}
\end{equation}
Taking logarithms gives
\begin{equation}
    \log J_{\Rell}^{W}
    =
    \frac12\log\det\X_{\Rell}
    =
    \frac12\Tr\log\X_{\Rell}
    =
    \frac12\sum_{i\in\Rell}\log\lambda_i .
    \label{eq:retained_log_jacobian_operational_v5}
\end{equation}
Thus the trace-log is the additive log-volume of the retained coordinate map.

The SETOL ERG gauge evaluates this retained volume using the normalized retained operator $\Xtilde_{\Rell}$ and retained eigenvalues $\tilde\lambda_{\Rell,i}$.  The normalized retained volume factor is
\begin{equation}
    \widetilde J_{\Rell}
    :=
    \sqrt{\det\Xtilde_{\Rell}}
    =
    \prod_{i\in\Rell}\sqrt{\tilde\lambda_{\Rell,i}} .
    \label{eq:relative_jacobian_operational_v5}
\end{equation}
Taking logarithms gives
\begin{equation}
    \log\widetilde J_{\Rell}
    =
    \frac12\Tr\log\Xtilde_{\Rell}
    =
    \frac12\sum_{i\in\Rell}\log\!\left(\tilde\lambda_{\Rell,i}\right).
    \label{eq:relative_log_jacobian_operational_v5}
\end{equation}
The scale-invariant ERG condition is
\begin{equation}
    \widetilde J_{\Rell}\simeq1
    \qquad\Longleftrightarrow\qquad
    \Tr\log\Xtilde_{\Rell}\simeq0 .
    \label{eq:trace_log_condition_operational_v5}
\end{equation}
This is the ERG trace-log gauge-fixing constraint; after the training step $\Opt[\mathbf W_t]$, it defines the retained ECS at the next step of the RG flow.
The optimizer flow thus both generates and implicitly tracks the RG flow.

We have now obtained the ECS trace-log condition in two ways.
SETOL obtains it from the HCIZ effective free energy, where the change from weights to correlations produces a log-determinant term.
The Jacobian construction  here obtains the same condition from the retained coordinate map for any selected ECS.

The fixed-point object is a normalized ESD shape, not a raw matrix.  The raw operator $\X_{\Rell}$ can change during training.  The operator $\Xtilde_{\Rell}$ places the retained spectrum on the fixed normalized scale.  A fixed point means that this retained spectral shape remains stable after further decimation and trace-log testing.  At the proposed fixed point, the SETOL/ECS window and the HTSR power-law window align,
\begin{equation}
    \Rell_{\rm ECS}^{\star}
    \simeq
    \mathcal T_{\rm PL}^{\star},
    \qquad
    \rho_{\rm tail}(\tilde{\lambda})\sim\tilde{\lambda}^{-2},
    \qquad
    \Tr\log\Xtilde_{\Rell}\simeq0 .
    \label{eq:alignment_claim_operational_v5}
\end{equation}
This alignment is not a definition.  It is the measurable agreement between the trace-log effective space and the independently fitted HTSR tail.

The truncated-SVD claim is the operational test of the fixed point.  SETOL predicts that, at the proposed RG fixed point,
one can replace $\W$ with the lower-rank (zero-padded) $\W_{\Rell}$ without degrading the layer's generalization contribution.
This is why the RG step can be described as a truncated-SVD decimation.

The next question is then to understand how $\alpha=2$ emerges as the boundary between
optimal learning and overfitting and is analogous to a critical universal exponent.

\clearpage
\subsection{RG Scale counting}
\label{sxn:rgcount}

RG scale counting asks how the layer-quality contribution changes as we move
across spectral scales. In our spectral RG description, the spectrum is not treated as a
flat list of eigenvalues. It is grouped into scale bands $[\tilde{\lambda}, b\tilde{\lambda}]$ that are
natural for a power-law distribution.  Then we ask whether each
band contributes a stable amount on the SETOL trace-log gauge, or if the gain collapses/diverges.

\begin{tcolorbox}[marginalbox,title=\textbf{RG boundary}]
RG counts gain by scale.
\eqtight{G_E(\Lambda,b)=\int_\Lambda^{b\Lambda}\tilde{\lambda}\rho({\lambda})\,d\tilde{\lambda}.}
\eqtight{\rho({\lambda})=C\tilde{\lambda}^{-\alpha},\qquad \alpha=2\Rightarrow G_E(\Lambda,b)=C\log b.}
\begin{itemize}
\item \textbf{$\alpha>2$:} larger bands contribute less and less gain.
\item \textbf{$\alpha=2$:} each log band contributes the same gain.
\item \textbf{$\alpha<2$:} spectral energy diverges
\end{itemize}
\takeaway{$\alpha\simeq2$ is the log-scale-balanced boundary.}
\end{tcolorbox}

One RG step is a spectral operation in this paper. It discards eigencomponents outside the current effective description. It evaluates the retained operator on the fixed SETOL scale, where $\Tr{\log \Xtilde_{\Rell}}=0$. It then compares the retained spectral shape with the previous retained shape. A fixed point means this decimate-and-test operation no longer changes the retained spectral shape or layer-quality gain.

The word ``exact'' is narrow here.
In SETOL, it refers to taking a single scale-invariant transformation, as in Eq.~\ref{eq:setol_RG_step}.
Here, it refers to a proposed Wilsonian map on the retained spectral distribution, not to a closed-form solution of the full network dynamics. We identify the candidate boundary and the candidate fixed-point signature of this map. It does not claim that one finite checkpoint exactly solves the full RG flow,
although it is hypothesized.

The three spectral regimes are summarized in Table~\ref{tab:regimes}.
These are direct analogues to the reduced description of the HTSR classes in Table~\ref{tab:htsr_rg_flow}.
Each regime arises by considering the behavior of the spectral band energy in the limit
of a continuous RG flow along the log-bands.  

\begin{table*}[t]
\centering
\caption{Scale counting, statistics, and empirical meaning of the three spectral regimes.}
\label{tab:regimes}
%\scriptsize
\footnotesize
\setlength{\tabcolsep}{4pt}
\renewcommand{\arraystretch}{1.15}
\begin{tabularx}{\textwidth}{@{}p{0.085\textwidth}@{\hspace{0.5em}}p{0.19\textwidth}p{0.23\textwidth}p{0.245\textwidth}X@{}}
%\begin{tabularx}{\textwidth}{@{}p{0.10\textwidth}p{0.20\textwidth}p{0.24\textwidth}p{0.22\textwidth}X@{}}
\toprule
Regime & spectral energy per scale & Scale-flow meaning & Statistical meaning & Learning meaning \\
\midrule
$\alpha>2$ & $\tilde{\lambda}^2\rho({\lambda})\sim\tilde{\lambda}^{2-\alpha}\to0$ & Spectral band gain decays; weak-tail sector or approach toward the trivial fixed point $\Fzero$. & First moment self-averages; dominant tail components stay controlled. & Stable but potentially weakly correlated or undertrained tail. \\
$\alpha=2$ & $\tilde{\lambda}^2\rho({\lambda})\sim\tilde{\lambda}^0$ & Scale-balanced logarithmic-band state; candidate $\Ferg$ when ECS aligns with the fitted tail. & Marginal self-averaging boundary. Proposed universal critical exponent. & Maximal stable structured-variation capture before tail-component domination. \\
$\alpha<2$ & $\tilde{\lambda}^2\rho({\lambda})\sim\tilde{\lambda}^{2-\alpha}\to\infty$ & Spectral band gain grows
toward the dominant tail components. & Tail components and fluctuations
can dominate the first moment. & Overfit risk; correlation traps; held-out degradation. \\
\bottomrule
\end{tabularx}
\vspace{-0.25em}
\end{table*}

The fixed point is layerwise. We do not posit one global fixed point for an entire multi-layer network. Rather, each layer has its own spectrum, its own retained ECS, and its own  Wilsonian-style spectral fixed point. A strong checkpoint is therefore a coordinated network state in which the matrix-like layers sit near their own SETOL ERG fixed points.

Because the spectral density $\rho({\lambda})$ is heavy-tailed power law for a well-trained
layer, the natural RG scale band is logarithmic in $\tilde{\lambda}$, and the natural coordinate is
\begin{equation}
x=\log\tilde{\lambda}.
\end{equation}
A fixed-width interval in $x$ is a fixed-ratio spectral shell $\tilde{\lambda}\in[\Lambda,b\Lambda]$.
This is the spectral analogue of coarse-graining by length scale in ordinary
RG. Each logarithmic band asks how much spectral energy is contributed by one
multiplicative decade of eigenvalues.

The object being counted is retained spectral energy $\Ered$~\eqref{eq:prob_esd_operational_v5}/
Since $\dd\tilde{\lambda}_{\Rell}=\tilde{\lambda}_{\Rell}\dd\log\tilde{\lambda}_{\Rell}$, we have
\begin{equation}
\dd\Ered=\tilde{\lambda}_{\Rell}\rhohat_{\Rell}(\tilde{\lambda}_{\Rell})\dd\tilde{\lambda}_{\Rell}
=\tilde{\lambda}_{\Rell}^2\rhohat_{\Rell}(\tilde{\lambda}_{\Rell})\dd\log\tilde{\lambda}_{\Rell}.
\label{eq:shell_dE}
\end{equation}
Thus the spectral energy per logarithmic spectral scale is
\begin{equation}
\frac{\dd\Ered}{\dd\log\tilde{\lambda}_{\Rell}}=\tilde{\lambda}_{\Rell}^2\rhohat_{\Rell}(\tilde{\lambda}_{\Rell}).
\label{eq:shellenergy}
\end{equation}
Substituting $\rhohat_{\Rell}(\tilde{\lambda}_{\Rell})\simeq C\tilde{\lambda}_{\Rell}^{-\alpha}$ then gives $C\tilde{\lambda}_{\Rell}^{2-\alpha}$, so $\alpha=2$ is precisely the case in which each multiplicative spectral shell contributes the same spectral energy.
In the RG sense, we say \emph{$\alpha=2$ is marginal}.

This quantity answers an interpretable question: how much of the retained layer's quality gain, and therefore how much of its possible accuracy contribution, lives near this multiplicative eigenvalue scale?

A power-law tail makes the answer transparent. Given the retained PL tail, we can evaluate Eq.~\ref{eq:shellenergy} as
\begin{equation}
\rhohat_{\Rell}(\tilde{\lambda}_{\Rell})\simeq C\tilde{\lambda}_{\Rell}^{-\alpha}
\rightarrow
\frac{\dd\Ered}{\dd\log\tilde{\lambda}_{\Rell}}
\sim C\tilde{\lambda}_{\Rell}^{2-\alpha}.
\label{eq:shellpower}
\end{equation}
Table~\ref{tab:regimes} summarizes the three RG regimes.
When $\alpha>2$, the high-eigenvalue shells carry less spectral energy with scale.
The strongest generalizing components are less and less important.  Generalization occurs but is weak.
When $\alpha<2$, high-eigenvalue shells carry more spectral energy with scale. The largest retained directions can dominate and the spectral energy diverges.
When $\alpha=2$, the change in energy is constant for all logarithmic shells; each shell carries comparable spectral energy gain. 

We now do the full RG scale-counting calculation of the spectral band energy gain for each regime.
Table~\ref{tab:rgterms} summarizes the  RG notation used below.

\begin{table*}[t!]
\centering
%\scriptsize
\footnotesize
\setlength{\tabcolsep}{5pt}
\renewcommand{\arraystretch}{1.05}
\caption{Special RG notation used in this section.}
\label{tab:rgterms}
\begin{tabularx}{\textwidth}{@{}p{0.16\textwidth}p{0.24\textwidth}X@{}}
\toprule
Term & Notation & Meaning here \\
\midrule
RG step &
$\Rell_{\rm ECS}$ &
Project to the retained ECS and evaluate the ESD on the SETOL scale. \\

Fixed point &
$\Rell_{\rm RG}[\rho^\star]=\rho^\star$ &
A SETOL-normalized ESD that is unchanged by the RG step. \\

Weak-tail branch &
$\Fzero,\quad g_E\ \text{undefined}$ &
The trivial solution  HTSR Random-like phase and nearby ESDs. \\

Correlated branch &
$\Ferg$ &
The empirical RG fixed point candidate. \\

Band Energy &
$\displaystyle G_E(\Lambda,b)$ &
Spectral energy carried by one logarithmic spectral band. \\

Band gain &
$g_E(\tilde{\lambda}_{\Rell}):=\tilde{\lambda}_{\Rell}^2\rhohat_{\Rell}(\tilde{\lambda}_{\Rell})$ &
Spectral energy per logarithmic spectral scale. \\

Spectral Beta function &
$\beta_E(g_E)=d g_E/d\log\tilde{\lambda}_{\Rell}$ &
The scale derivative of the spectral energy. For a PL tail, $\beta_E(g_E)=(2-\alpha)g_E$. \\

Scaling dimension &
$y_E=2-\alpha$ &
The RG exponent controlling energy or spectral energy.\\

Marginal condition &
$y_E=2-\alpha=0$ &
The logarithmic spectral bands are balanced and carry comparable energy/gain. \\
\bottomrule
\end{tabularx}
\vspace{-0.6em}
\end{table*}

The finite-shell calculation says the same thing without derivatives. For a multiplicative shell $[\Lambda,b\Lambda]$,
\begin{equation}
G_E(\Lambda,b)=\int_\Lambda^{b\Lambda}\tilde{\lambda}_{\Rell}\rhohat_{\Rell}(\tilde{\lambda}_{\Rell})\dd\tilde{\lambda}_{\Rell}.
\label{eq:finite_shell}
\end{equation}
If $\rhohat_{\Rell}(\tilde{\lambda}_{\Rell})=C\tilde{\lambda}_{\Rell}^{-\alpha}$, then for $\alpha\ne2$,
\begin{equation}
G_E(\Lambda,b)=
\frac{C\left[(b\Lambda)^{2-\alpha}-\Lambda^{2-\alpha}\right]}{2-\alpha},
\label{eq:finite_shell_ne2}
\end{equation}
while for $\alpha=2$,
\begin{equation}
G_E(\Lambda,b)=C\log b.
\label{eq:finite_shell_eq2}
\end{equation}
The last expression is independent of $\Lambda$. That is the cleanest statement of spectral scale balance: a fixed-ratio band carries the same spectral energy no matter where the band is placed in the tail.

We therefore use the spectral shell gain itself as the running coupling,
\begin{equation}
g_E(\tilde{\lambda}_{\Rell})
:=
\frac{\dd\Ered}{\dd\log\tilde{\lambda}_{\Rell}}
=
\tilde{\lambda}_{\Rell}^2\rhohat_{\Rell}(\tilde{\lambda}_{\Rell}).
\label{eq:gE_correct}
\end{equation}
This derivative is taken within one fixed checkpoint while neighboring spectral bands are compared.

The local slope is a scale derivative. It is not a training-time derivative. It compares neighboring eigenvalue bands inside one checkpoint. In RG language, this scale derivative is called a \emph{beta function}. Define
\begin{equation}
\alpha_{\rm loc}(\tilde{\lambda}_{\Rell})=
-\frac{\dd\log\rhohat_{\Rell}(\tilde{\lambda}_{\Rell})}{\dd\log\tilde{\lambda}_{\Rell}}.
\label{eq:alpha_loc}
\end{equation}
Then
\begin{equation}
s(\tilde{\lambda}_{\Rell})=\frac{\dd}{\dd\log\tilde{\lambda}_{\Rell}}
\log\left[\tilde{\lambda}_{\Rell}^2\rhohat_{\Rell}(\tilde{\lambda}_{\Rell})\right]
=2-\alpha_{\rm loc}(\tilde{\lambda}_{\Rell}),
\label{eq:slope_correct}
\end{equation}
and equivalently
\begin{equation}
\frac{\dd g_E}{\dd\log\tilde{\lambda}_{\Rell}}=
\left[2-\alpha_{\rm loc}(\tilde{\lambda}_{\Rell})\right]g_E(\tilde{\lambda}_{\Rell}).
\label{eq:beta_local}
\end{equation}
For an exact power law on the fitting window, we can then define
\emph{the leading, or tree-level, RG beta function} 
\begin{equation}
\beta_E(g_E)
\equiv
\frac{dg_E}{d\log\tilde{\lambda}_{\Rell}}
=
(2-\alpha)g_E 
\end{equation}
where the associated scaling dimension is
\begin{equation}
y_E = 2-\alpha .
\end{equation}
The sign of $y_E$ determines whether spectral energy is
relevant, irrelevant, or marginal under rescaling.
This scale-counting argument identifies the marginal
boundary, but it does not by itself prove a complete
Wilsonian fixed point.

The fixed-point language should therefore be read operationally. A fixed point is not a single training checkpoint.
It is a SETOL-normalized spectral shape that is stable  after the final decimation and trace-log testing.

The weak-tail branch is the trivial solution in which no extensive
retained ECS survives the coarse graining:
\begin{equation}
  \Fzero:
    \qquad
    w_{\Rell}
    :=
    \frac{M_{\Rell}}{M}
    \longrightarrow 0 .
        \qquad
g_E\;\text{undefined}
    \label{eq:trivial_rg_branch}
\end{equation}
The correlated solution is the ERG fixed-point candidate,
\begin{equation}
\Ferg:\qquad
0<g_E^\star<\infty,
\quad
\alpha\simeq2,
\quad
\Rell_{\rm ECS}^\star\simeq\mathcal T_{\rm PL}^\star.
\label{eq:FHTSR}
\end{equation}
Power-law scale invariance alone does not uniquely select $\alpha=2$;
what distinguishes $\alpha^\star\simeq2$ is that the spectral coupling
is marginal.

This is the RG critical boundary.
The scale exponent $y_E=2-\alpha$ changes sign there, so $\alpha=2$ plays the role of the critical exponent for this HTSR spectral universality class.
Thus $\alpha=2$ is the tree-level scale-balance condition for logarithmic bands, while $\Ferg$ is the empirical correlated fixed-point candidate.

Returning to the earlier discussion, notice that the
HTSR boundary between self-averaging and non-self-averaging phases occurs at the same exponent as the RG scale-balance boundary.
That is why $\alpha=2$ is not merely a fitted number.
It is the proposed operational critical point where a layer has learned as much correlation structure as possible before overfitting.

By leveraging this RG spectral view, one can now design optimizers that approximate the RG flow;
the next section explores this.

%\input{section_5_51}
%\input{section_5_52}
% Two-page revision: fixed-rank motivation, scale matching, and empirical ERG gaps.
% One compact SETOL citation; no separate bibliography in the standalone insert.
\subsection{The SETOL and MLE normalizations}
\label{sxn:meanlog_gauge}

SETOL and the power-law fit choose different spectral reference scales,
so explaining their agreement near marginality requires both a reason
for scaling eigenvalues by $M$ and a test of whether the two rules select
the same ECS.
SETOL uses $\lambda/M$; the fit uses
$\lambda/\PLstart$.

\begin{tcolorbox}[marginalbox,title=\textbf{Matching the normalizations}]
  We have two different normalizations
\eqtight{
\text{SETOL: }\tilde{\lambda}_{\mathrm ECS}:=\frac{\lambda}{M},
\qquad
\text{MLE: }\tilde{\lambda}_{\mathrm PL}:=\frac{\lambda}{\PLstart}.
}
At $\alpha=2$, matching requires $M\simeq e\PLstart$.
\takeaway{RG scale invariance motivates $1/M$.
Why these normalizations align remains an open question.}
\end{tcolorbox}

\paragraph{Why marginality motivates $1/M$.}
A stable marginal tail motivates division by matrix size because a fixed
upper rank moves to proportionally larger eigenvalues as the spectrum grows.
Let the probability-normalized spectrum have survival tail
$\overline F(\lambda)\simeq C\lambda^{1-\alpha}$, with unchanged amplitude
$C$ and common units across sizes.
The characteristic location of the $r$th largest eigenvalue obeys
\begin{equation}
\begin{aligned}
r&\simeq M\overline F(\lambda_{(r)}),
&\lambda_{(r)}&\simeq(CM/r)^{1/(\alpha-1)},\\
\alpha&=2:
&\frac{\lambda_{(r)}}{M}&\simeq\frac Cr.
\end{aligned}
\label{eq:fixed_rank_marginal_scaling}
\end{equation}
Doubling $M$ therefore doubles this fixed-rank scale while leaving
its normalized position unchanged (Figure~\ref{fig:fixed_rank_scaling}).

\begin{center}
\includegraphics[width=\columnwidth]{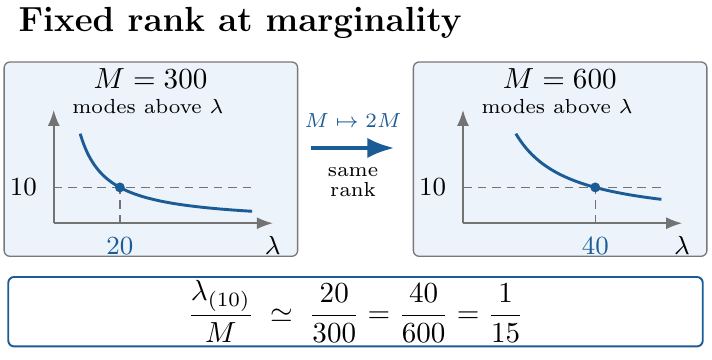}
\vspace{-0.65em}
\captionof{figure}{\textbf{Division by $M$ preserves a fixed upper-rank scale.}
For the illustrative tail $\overline F(\lambda)=2/(3\lambda)$,
doubling $M$ moves the threshold with ten expected exceedances
from $20$ to $40$.  The normalized scale is unchanged.
This is an analytic illustration, not a measured width sweep.}
\label{fig:fixed_rank_scaling}
\end{center}

This scaling applies to fixed upper ranks away from an endpoint;
a fixed fraction $r/M$ instead gives a size-independent threshold.
The retained count $M_{\mathcal R}$ normalizes ECS averages.
Using it as an eigenvalue denominator requires a separate RG argument,
because truncation changes the cutoff and retained fraction.

% Standalone layout only; these hooks are inert in the full draft.
\ifdefined\RGTheoryColumnBreak\RGTheoryColumnBreak\fi
\paragraph{How the MLE estimator sets the scale.}
The~\WW~tool fits the power law tail using a Maximum Likelihood Estimator (MLE),
which itself is defined by estimating the mean log-eigenvalue of the tail
\begin{equation}
\left\langle\log\tilde{\lambda}_{\mathrm{PL}}\right\rangle_{\rm MLE}=\frac{1}{\hat\alpha-1},
\qquad
G_{\rm PL}=\PLstart e^{1/(\hat\alpha-1)}.
\label{eq:mle_logmean_bridge}
\end{equation}
where $G_{\rm PL}$ is the \emph{retained geometric mean}.
At $\hat\alpha=2$, the expected or mean log-eigenvalue is
$\langle \ln \tilde{\lambda}_{\mathrm{PL}}\rangle_{\rm MLE}=1.0$ when
the tail eigenvalues $\tilde{\lambda}_{\mathrm{PL}}$ are normalized by $\PLstart$.
In $G_{\rm PL}$, the factor $e$ moves the mean log-ratio from $1\rightarrow 0$.
The SETOL condition places $G_{\mathcal R}$ near $M$.
Agreement therefore requires $G_{\rm PL}\simeq G_{\mathcal R}\simeq M$;
the scale-invariance argument here alone does not impose this equality,
but it is a start.

The physical matching hypothesis identifies the marginal PL tail with
the persistent correlations that survive the RG construction.
If the two selection rules approach the same effective modes and mean
log in common units, their geometric means approach one another.
This provides a physical reason to expect a small gap near marginality.
We have no mathematical proof that the two selections converge,
and the fixed-rank argument does not fix their relative amplitude.
Instead, we rely on empirical evidence.

\begin{figure*}[!t]
\centering
\includegraphics[width=\textwidth]{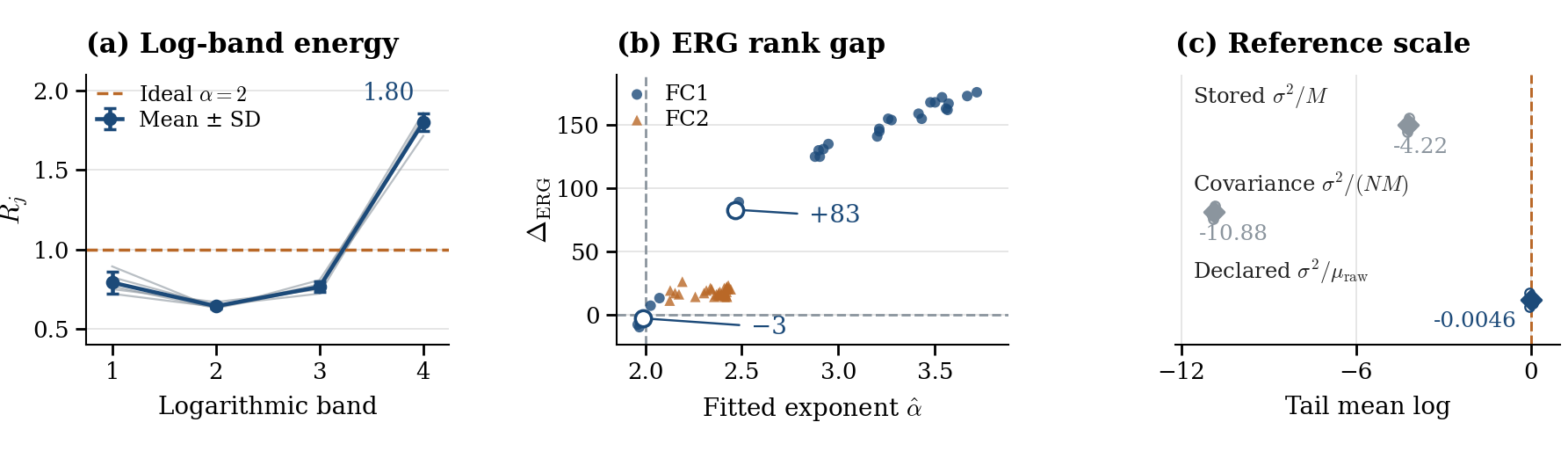}
\vspace{-0.8em}
\caption{\textbf{Trained checkpoints show approximate matching and substantial exceptions.}
(a) Five FC1 tails at learning rate $0.32$ have
$\overline{\hat\alpha}=2.0036$ but unequal log-band energies.
$R_j$ divides each band's energy by that checkpoint's mean band energy;
error bars are between-seed standard deviations.
(b) ERG rank gaps for both hidden layers of all 30 checkpoints; labels
mark the paired seed-4 FC1 examples.
(c) The same five FC1 tails have different mean-log offsets under the
stored, covariance, and declared full-spectrum conventions.
The archive is an independent MNIST protocol replication.}
\label{fig:normalization_checkpoint_audit}
\end{figure*}

\paragraph{What marginality balances.}
The same marginal law gives equal logarithmic bands equal expected
energy because declining mode counts compensate for increasing
individual energies.
For probability density $\rho(\lambda)=C\lambda^{-2}$, the mean
energy $\lambda/M$ per log interval is
\begin{equation}
\frac{dE_M}{d\log(\lambda/M)}
=\frac{\lambda^2\rho(\lambda)}{M}=\frac CM.
\label{eq:energy_per_log_band}
\end{equation}
Figure~\ref{fig:normalization_band_theory} depicts this balance,
while Figure~\ref{fig:normalization_checkpoint_audit}(a) shows
this for a typical well trained layer (described below).

\begin{center}
\includegraphics[width=\columnwidth]{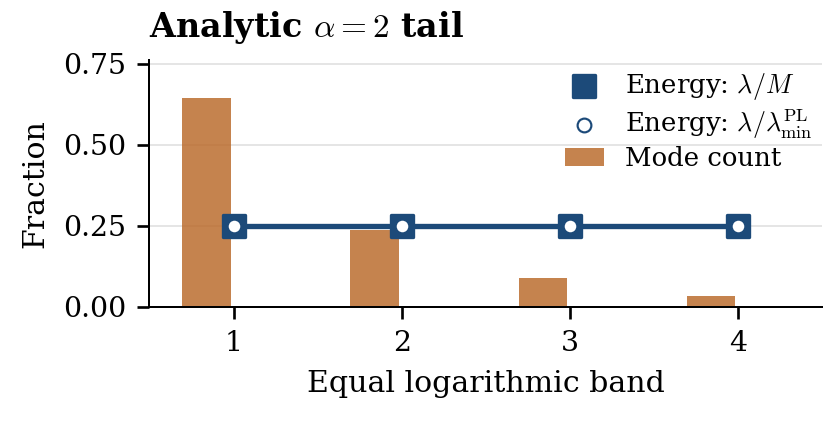}
\vspace{-0.65em}
\captionof{figure}{\textbf{Marginality balances energy under either reference.}
An analytic $\alpha=2$ density on four equal logarithmic bands
has decreasing count fractions and equal energy fractions.
Both energy curves use the same modes and divide by the energy in
the plotted window; their reference scales cancel.}
\label{fig:normalization_band_theory}
\end{center}

\ifdefined\RGDataPageBreak\RGDataPageBreak\fi
\paragraph{Which normalization the checkpoints test.}
Checkpoint comparisons require a fixed spectral convention because the
saved weight scale determines the trace-log reference.
Similar to the experiments in~\cite{MartinHinrichsSETOL2025}.
we examine $30$ checkpoints of a
$784\to300\to100\to10$ MNIST MLP: plain SGD, batch size $32$,
six learning rates $0.01$--$0.32$, and five seeds.
The archived Frobenius convention divides
$\lambda_i^{\rm raw}=\sigma_i(\W)^2$ by its full-spectrum mean
$\mu_{\rm raw}=M^{-1}\sum_i\lambda_i^{\rm raw}$.
In SETOL coordinates this becomes
\begin{equation}
\lambda_i:=\frac{M}{\mu_{\rm raw}}\lambda_i^{\rm raw},
\qquad
\tilde{\lambda}_{\mathrm{ECS},i}:=\frac{\lambda_i}{M}.
\label{eq:declared_checkpoint_conversion}
\end{equation}
The cutoff undergoes the same conversion.

The fitted tail's mean-log offset tests whether its center agrees
with the declared reference:
\begin{equation}
\Delta_{\rm mean-log}:=\left\langle\log\tilde{\lambda}_{\mathrm{ECS}}\right\rangle_{\rm MLE}
=\frac{1}{\hat\alpha-1}
-\log\frac{M}{\PLstart}.
\label{eq:reconcile_normalization_mismatch}
\end{equation}
In Figure~\ref{fig:normalization_checkpoint_audit}(a), the first $3$ bands
are fairly flat, where as the last band diverges up (presumably
due to the finite cut-off at $\lambda_{max}$).

More directly, Figure~\ref{fig:normalization_checkpoint_audit}(b) shows that
the ERG rank gap shrinks to $\Delta_{ERG}\rightarrow 0$ as $\alpha \rightarrow 2$.
Additional, similar results are reported in~\cite{MartinHinrichsSETOL2025}.

Finally, Figure~\ref{fig:normalization_checkpoint_audit}c) shows that trace-log agreement depends on the spectral normalization.
Dividing $\lambda_{i}^{\rm raw}$ by the full-ESD mean $\mu_{\rm raw}$
gives mean $\Delta_{\rm mean-log}=-0.0046$ and
mean $|\Delta_{\rm mean-log}|=0.051$ for the five FC1 fits at learning rate $0.32$.
These small offsets place the tail geometric means near the full-ESD arithmetic means.
Dividing instead by $M=300$ gives a mean offset of $-4.219$; the covariance convention $\sigma_i^2/(NM)$, with $N=784$, gives $-10.883$.

\paragraph{Why the ERG rank gap can remain nonzero.}
We define the \emph{ERG rank gap} as
\begin{equation}
\Delta_{\rm ERG}:=M_{\mathcal R}-M_{\rm PL}.
\label{eq:erg_rank_gap}
\end{equation}
In other models, the ERG rank gap persists and can be as large as $50$.
For nested upper tails, its sign agrees with the normalized cutoff difference 
$\Delta\tilde{\lambda}_{\min}:=\PLstart/M-\ECSstart$ used in SETOL,
although the units differ.
When a zero crossing exists, a positive fitted-tail log sum calls for
adding lower modes below the reference; a negative sum calls for
removing low modes.  The normalization and cutoff region therefore
help determine the gap, alongside the exponent.

\ifdefined\RGDataColumnBreak\RGDataColumnBreak\fi
\paragraph{Small and large gaps in the same training setup.}
The recovered checkpoints show a useful near-marginal correspondence
without exact agreement of the two retained spaces.
For FC1 with seed $4$, the original selections give
\begin{center}
\small
\begin{tabular}{c@{\quad}c@{\quad}r@{\quad}r@{\quad}r}
Learning rate & $\hat\alpha$ & $M_{\rm PL}$ & $M_{\mathcal R}$ & $\Delta_{\rm ERG}$\\
\hline
$0.16$ & $2.468$ & $119$ & $202$ & $+83$\\
$0.32$ & $1.987$ & $162$ & $159$ & $-3$
\end{tabular}
\end{center}
A three-mode difference between roughly $160$-mode tails is close
physical agreement; an $83$-mode difference is substantial.
Across the five near-marginal FC1 fits, gaps span $[-10,+13]$
despite their signed mean of $-0.2$.

The observed ERG rank gap usually approaches zero from above as $\alpha \rightarrow 2$.
All $57$ hidden-layer fits with $\hat\alpha\ge2$ have positive rank gaps;
the three with $\alpha < 2$ have negative gaps (Figure~\ref{fig:normalization_checkpoint_audit}b).
SETOL reports a similar approach toward zero from above, crossings below
$\alpha=2$, with a few exceptions: some layers with $\alpha\approx 2$ have large cutoff gaps,
and some DenseNet layers with $\alpha < 2$ retain positive gaps.

The same near-marginal checkpoints retain uneven band energies, so
center matching does not establish exact spectral scale invariance.
Their four mean relative energies,
$(0.792,0.643,0.764,1.801)$, survive scalar rescaling
(Figure~\ref{fig:normalization_checkpoint_audit}a).
Why do these differ ? Possible causes include sampling effects,
conditioning on $\lambda_{max}$, and departures from an ideal power law.

%More broadly, we still need to identify how and why the scale convention and RG
%transformation that align the independently selected  ECS and power law tail.
%This is left to future work.

\clearpage
% Section 5.7. Layer dilation identities and the associated coupled scale flow.
\subsection{Layer RG Beta Functions}
\label{sxn:sgd_layer_flow}

Our spectral RG separates a layer's overall amplitude from its
normalized spectrum. The same amplitude appears directly in
backpropagation: rescaling every weight in a layer turns its loss
response into a logarithmic scale derivative. These derivatives
define coupled dilation flows through one common loss.

Let $s_\ell=\ln c_\ell$ measure a layer dilation,
$\mathbf z_\ell$ its weighted input, and
$\mathcal L_{\mathcal B}=\langle L\rangle_{\mathcal B}$ the mean
prediction loss. Holding the minibatch, layer shapes, and biases
fixed, use this loss as a potential to define a continuous flow
in the log scales $\mathbf s=(s_1,\ldots,s_K)$, with parameter $\tau$:
\begin{tcolorbox}[marginalbox,title=The coupled dilation beta functions]
\eqtight{\begin{aligned}
 \beta_\ell(\mathbf s)
   :=\frac{ds_\ell}{d\tau}
   &=-\frac{\partial\mathcal L_{\mathcal B}}{\partial\ln c_\ell}\\
   &=-\left\langle\sum_i z_{\ell i}
               \frac{\partial L}{\partial z_{\ell i}}\right\rangle_{\mathcal B}\\
   &=-\langle\W_\ell,\nabla_{\W_\ell}\mathcal L_{\mathcal B}\rangle_F .
\end{aligned}}
\takeaway{Backpropagation supplies the logarithmic scale derivative
that generates each layer's loss-decreasing dilation.}
\end{tcolorbox}

\paragraph{Loss and layer scale.}
For a feedforward network, write $L^{(n)}$ for the prediction loss of
example $n$. Its minibatch mean is
\[
  \mathcal L_{\mathcal B}
   :=\frac{1}{|\mathcal B|}\sum_{n\in\mathcal B}L^{(n)}
   =\langle L\rangle_{\mathcal B},
\]
and its weight derivative is
$\nabla_{\W_\ell}\mathcal L_{\mathcal B}$.
Parameterize each layer by
\[
  \W_\ell=e^{s_\ell}\overline{\W}_\ell,\qquad
  c_\ell=e^{s_\ell}>0,\qquad
  \mathbf z_\ell=\W_\ell\mathbf h_{\ell-1} .
\]
Each fixed nonzero $\overline{\W}_\ell$ specifies the layer shape;
varying $s_\ell$ changes only its common amplitude.

\paragraph{The logarithmic derivative in backpropagation.}
For one example, suppress its index. The activation is
$\mathbf h_\ell=f_\ell(\mathbf z_\ell+\mathbf b_\ell)$.
Varying only $s_\ell$ gives
$\partial_{s_\ell}\mathbf z_\ell=\mathbf z_\ell$.
Where the network is differentiable, the chain rule gives
\[
\begin{aligned}
 \nabla_{\W_\ell}L
   &=(\nabla_{\mathbf z_\ell}L)\mathbf h_{\ell-1}^{\top},\\
 \frac{\partial L}{\partial s_\ell}
  &=\mathbf z_\ell^{\top}\nabla_{\mathbf z_\ell}L
   =\langle\W_\ell,\nabla_{\W_\ell}L\rangle_F .
\end{aligned}
\]
For each nonzero component, hold its sign and all other components
fixed. Its logarithmic derivative is
\[
 z_{\ell i}\frac{\partial L}{\partial z_{\ell i}}
   =\frac{\partial L}
          {\partial\ln|(\W_\ell\mathbf h_{\ell-1})_i|} .
\]
The response to a common scale of the weighted-input vector sums
these component responses. Averaging over examples gives
$\partial\mathcal L_{\mathcal B}/\partial\ln c_\ell$.
The operator $\partial_{\ln c_\ell}$ generates an infinitesimal
dilation, the rescaling part of RG. The boxed beta functions turn
this loss response into a flow along each layer's dilation coordinate.

\newpage
\paragraph{One loss couples the layer flows.}
Each $\beta_\ell(\mathbf s)$ responds to the other layer scales
through the common loss. The input $\mathbf h_{\ell-1}$ depends on
earlier layers, while
$\nabla_{\mathbf z_\ell}L$ carries the response through later layers.
Changing one layer can therefore change the responses of others.
A stationary scale for layer $\ell$ satisfies
$\beta_\ell(\mathbf s)=0$; its location can depend on the other
layers. A layer can be stationary while its neighbors still move,
and their motion can shift its stationary scale.

\begin{center}
\resizebox{\columnwidth}{!}{%
\begin{tikzpicture}[x=1cm,y=1cm,font=\sffamily\small,>=Stealth]
 \path[use as bounding box] (-.15,-.12) rectangle (8.05,3.75);
 \node[font=\sffamily\bfseries\small,text=boxblue]
   at (3.95,3.54) {One common minibatch loss};
 \node[font=\normalsize] (loss) at (3.95,3.08)
   {$\mathcal L_{\mathcal B}(s_1,s_2,s_3)$};
 \foreach \x/\l in {1.1/1,3.95/2,6.8/3}{
   \node[draw=boxblue,fill=boxlightblue,rounded corners=2pt,
     minimum width=2.2cm,minimum height=1.00cm,
     align=center,line width=.7pt] (layer\l) at (\x,2.03)
     {layer $\l$\\[3pt]$\dot s_{\l}=\beta_{\l}(\mathbf s)$};
   \draw[->,boxblue!65,line width=.65pt]
     (loss.south) -- (layer\l.north);
 }
 \draw[<->,boxblue,line width=1pt] (layer1.east) -- (layer2.west);
 \draw[<->,boxblue,line width=1pt] (layer2.east) -- (layer3.west);
 \draw[boxblue!55] (layer1.south) -- (1.1,.92) -- (6.8,.92) -- (layer3.south);
 \draw[boxblue!55] (layer2.south) -- (3.95,.92);
 \draw[->,boxblue!70] (3.95,.92) -- (3.95,.60);
 \node[font=\small] at (3.95,.30)
   {$\beta_1(\mathbf s^\star)=\beta_2(\mathbf s^\star)
     =\beta_3(\mathbf s^\star)=0$};
\end{tikzpicture}}
\captionof*{figure}{\textbf{Coupled layer dilations.}
Each layer contributes one scale response to the shared loss.
A joint stationary point satisfies every layer's dilation condition.}
\end{center}

The coupled flow decreases the common minibatch loss,
\[
 \frac{d\mathcal L_{\mathcal B}}{d\tau}
   =-\sum_{\ell=1}^{K}\beta_\ell(\mathbf s)^2\leq0 .
\]
At a joint fixed point, $\boldsymbol\beta(\mathbf s^\star)=0$, so
no layer dilation changes the loss to first order. The loss itself
need not vanish.

\paragraph{The same dilation rescales the ESD.}
For $\W_\ell\in\mathbb R^{N_\ell\times M_\ell}$, the correlation matrix is
$\X_\ell=\W_\ell^\top\W_\ell/N_\ell$.
A dilation acts on this operator, its eigenvalues, and its
probability-normalized ESD as
\[
\begin{aligned}
 \W_\ell&\mapsto c_\ell\W_\ell,&
 \X_\ell&\mapsto c_\ell^2\X_\ell,\\
 \lambda_{\ell i}&\mapsto c_\ell^2\lambda_{\ell i},&
 \rho_{\ell,c}(\lambda)&=c_\ell^{-2}
                 \rho_\ell(\lambda/c_\ell^2).
\end{aligned}
\]
Along the flow, $\partial_{s_\ell}\X_\ell=2\X_\ell$ gives
\[
 \frac{d\X_\ell}{d\tau}=2\beta_\ell(\mathbf s)\X_\ell,\qquad
 \frac{d\ln\lambda_{\ell i}}{d\tau}=2\beta_\ell(\mathbf s)
\]
for $\lambda_{\ell i}>0$ in the second identity. Thus every
positive eigenvalue shifts at the same rate on a logarithmic axis.
The full-spectrum first-moment normalization of
Section~\ref{sxn:meanlog_gauge} removes this common amplitude,
\[
 \widetilde{\X}_\ell=\frac{\X_\ell}{m_\ell},\qquad
 m_\ell=\frac{1}{M_\ell}\Tr\X_\ell,\qquad
 \partial_{\ln c_\ell}\widetilde{\X}_\ell=0 .
\]
Both $\X_\ell$ and its positive mean $m_\ell$ acquire the factor
$c_\ell^2$, so their ratio is unchanged. The normalized ESD retains
the spectral shape used for ECS selection. Thus backpropagation
exposes the same layer dilation whose amplitude our spectral RG
normalization removes.

\clearpage
% Include after subsection 5.8, immediately before Section 6:
% \input{implications_of_marginality}
% Uses the parent paper's marginalbox, boxblue, boxlightblue, and TikZ setup.
% All equations and the analytic figure are unnumbered.
% Starts a new page and uses one column for each subsubsection.
\clearpage
\subsection{Implications of marginality}
\label{sxn:implications_marginality}

\subsubsection{The model-average exponent}
\label{sxn:model_average_marginality}
SETOL combines positive layer qualities into the model-quality approximation
$\overline Q_{\rm NN}\simeq\prod_{\ell=1}^K\overline Q_\ell$
[Eq.~\eqref{eq:setol_layer_product}]. Its logarithm converts the
layer product into a sum, giving the average logarithmic quality,
\[
 \frac{1}{K}\log\overline Q_{\rm NN}
 \simeq\frac{1}{K}\sum_{\ell=1}^K\log\overline Q_\ell .
\]
The layer product motivates averaging the logarithmic gains of
the spectral bands. At a fixed checkpoint, Eq.~\eqref{eq:shellpower} gives
$g_{E,\ell}(\Lambda_\ell)\propto\Lambda_\ell^{2-\alpha_\ell}$.
Comparing $\Lambda_\ell$ with $b\Lambda_\ell$ inside each fitted window
multiplies the gain by $b^{2-\alpha_\ell}$. A common factor
$b>1$ gives every layer the same step in logarithmic scale.
The powers of their product then add.

\begin{tcolorbox}[marginalbox,title=Marginality on average]
\[
\begin{gathered}
 \bar\alpha:=\frac{1}{K}\sum_{\ell=1}^K\alpha_\ell,\qquad
 \prod_{\ell=1}^K b^{2-\alpha_\ell}=b^{K(2-\bar\alpha)},\\[3pt]
 \bar y_E:=\frac{1}{K}\sum_{\ell=1}^K(2-\alpha_\ell)
          =2-\bar\alpha .
\end{gathered}
\]
\textcolor{boxblue}{The product of the band gains is marginal
when $\bar\alpha=2$. The mean logarithmic band slope then vanishes.}
\end{tcolorbox}

This average expresses model marginality as a balance across layers.
Layers with $\alpha_\ell=2$ give $\bar\alpha=2$. Equal and opposite
deviations from two also yield this average through compensation.

Two layers with $\alpha_1=1.5$ and $\alpha_2=2.5$ achieve
$\bar\alpha=2$ through compensation. Doubling the spectral scale
multiplies their band gains by $\sqrt{2}$ and $1/\sqrt{2}$.
These reciprocal factors have product one and balance the layer slopes.

\begin{center}
\resizebox{.96\columnwidth}{!}{%
\begin{tikzpicture}[x=1cm,y=1cm,font=\sffamily\small,>=Stealth]
 \path[use as bounding box] (-.85,-.62) rectangle (7.0,3.28);
 \node[text=boxblue,font=\sffamily\bfseries\small] at (3.05,3.0)
   {Layers balance their spectral slopes};
 \draw[->,gray] (0,0) -- (6.7,0);
 \draw[->,gray] (0,0) -- (0,2.65);
 \foreach \x/\lab in {0/1,2.4/2,4.8/4}{
   \draw[gray] (\x,0) -- (\x,-.08);
   \node[below,font=\small] at (\x,-.08) {$\lab$};
 }
 \foreach \y/\lab in {.3/{1/2},1.3/{1},2.3/{2}}{
   \draw[gray] (0,\y) -- (-.08,\y);
   \node[left,font=\small] at (-.08,\y) {$\lab$};
 }
 \draw[boxblue,line width=1.2pt] (0,1.3) -- (4.8,2.3);
 \draw[orange!80!black,line width=1.2pt] (0,1.3) -- (4.8,.3);
 \draw[black,dashed,line width=.9pt] (0,1.3) -- (4.8,1.3);
 \node[right,text=boxblue] at (4.86,2.3) {$\alpha_1=1.5$};
 \node[right,text=orange!80!black] at (4.86,.3) {$\alpha_2=2.5$};
 \node[right,font=\sffamily\scriptsize] at (4.86,1.3) {geometric mean};
 \node[below,font=\sffamily\scriptsize] at (3.0,-.35)
   {spectral scale factor $b$ (log scale)};
\end{tikzpicture}}
\captionsetup{justification=raggedright,singlelinecheck=false}
\captionof*{figure}{\textbf{Marginal on average.}
The band-gain ratios $g_{E,\ell}(b\Lambda_\ell)/g_{E,\ell}(\Lambda_\ell)$
have a geometric mean of one. Both axes use logarithmic scales.}
\end{center}

The mean exponent diagnoses average marginality, and its spread
describes how the layers achieve that balance. Exponents clustered
near two indicate coordinated layerwise marginality. Exponents
on opposite sides of two indicate compensation.

\newpage
\subsubsection{Mode seeking versus geometry seeking}
\label{sxn:mean_variance_marginality}
The spectral moments place the mean and variance at different
marginal boundaries. For a tail
$\rho(\lambda)\simeq C\lambda^{-\alpha}$ with $\alpha>1$,
we extend the upper cutoff $\Lambda$ while holding the lower cutoff
$\lambda_0>0$ and amplitude $C$ fixed. The tail contribution to the
$k$th raw moment obeys
\[
\begin{aligned}
 \mu_k(\Lambda)&:=\int_{\lambda_0}^{\Lambda}
                  \lambda^k\rho(\lambda)\,d\lambda,\\
 \frac{d\mu_k}{d\log\Lambda}
   &=\Lambda^{k+1}\rho(\Lambda)
     \simeq C\Lambda^{k+1-\alpha}.
\end{aligned}
\]
Equal contributions from successive logarithmic bands define
moment marginality. At $\alpha=k+1$, these contributions give
$\mu_k(\Lambda)\sim C\log(\Lambda/\lambda_0)$.
For $\alpha>k+1$, the moment converges. Thus $\alpha=2$ is marginal for the first moment and
$\alpha=3$ is marginal for the second moment. Since the mean is
finite at $\alpha=3$, the variance inherits the second moment's
logarithmic growth.
Both ideal moments are finite for $\alpha>3$.

\paragraph{Mode seeking.}
We hypothesize that layers with $1<\alpha<2$ seek particular
modes of the training data and become more prone to overfitting.
Their upper spectral bands contribute progressively more
first-moment energy, favoring concentration of squared gain in a
few leading singular directions. These
directions amplify training-specific patterns and concentrate the
layer's response on them. At $\alpha=2$, equal logarithmic bands
carry equal first-moment energy, balancing that gain across
spectral scales.

\paragraph{Geometry seeking.}
We hypothesize that layers with $2<\alpha<3$ seek the geometry of
the data by amplifying structured variation across directions.
Their ideal mean is finite while their variance grows with the
upper cutoff. Higher bands contribute less to
the mean but more to the second moment. At $\alpha=2.5$, for
example, moving up one octave multiplies the first-moment band
weight by $1/\sqrt{2}$ and its second-moment weight by $\sqrt{2}$.

This spread measures how unevenly a layer amplifies its singular
directions. For the full
mean-normalized correlation operator $\widetilde{\mathbf X}$
with $M$ eigenvalues and $\langle\tilde\lambda\rangle=1$,
\[
 \operatorname{Var}_{\rm ESD}(\tilde\lambda)
   =\frac{1}{M}\sum_{i=1}^M(\tilde\lambda_i-1)^2
   =\frac{1}{M}\|\widetilde{\mathbf X}-I_M\|_F^2 .
\]
The normalized spectrum expresses these hypotheses through
concentration and directional spread. The fraction of the trace
carried by the leading modes measures concentration. The variance
of the eigenvalues measures the spread of squared directional
gains. In the proposed learning picture, mode seeking emphasizes
the first quantity and geometry seeking emphasizes the second.
\clearpage

\clearpage
\subsection{Removing Redundant RG Directions}
\label{sxn:redundant_rg_directions}

A Wilsonian RG flow is a flow on equivalence classes, not on every microscopic coordinate. Field redefinitions, coordinate changes, and normalization conventions can move the representative of a theory without changing its long-distance content. These motions are the \emph{redundant directions} of RG theory. They are distinct from irrelevant directions: an irrelevant perturbation changes the theory but decays under coarse graining, whereas a redundant perturbation changes only its description. In the retained spectral description, the physical state is therefore the trace-log-gauge-fixed ECS modulo transformations that leave its quotient spectral shape unchanged.

For a small displacement $\delta\W$, two redundancies are relevant:
\begin{itemize}
\item \textbf{Trace-log (volume) drift.} The update changes the retained product scale and leaves the chosen gauge slice, while leaving the normalized spectral shape essentially unchanged.
\item \textbf{Isospectral coordinate motion.} The update changes retained coordinates by an orthogonal similarity transformation without changing the retained eigenvalue multiset.
\end{itemize}
The remaining component is the candidate physical RG motion: it can redistribute spectral mass, change the tail burden, or change the fitted exponent $\alpha$.

\begin{center}
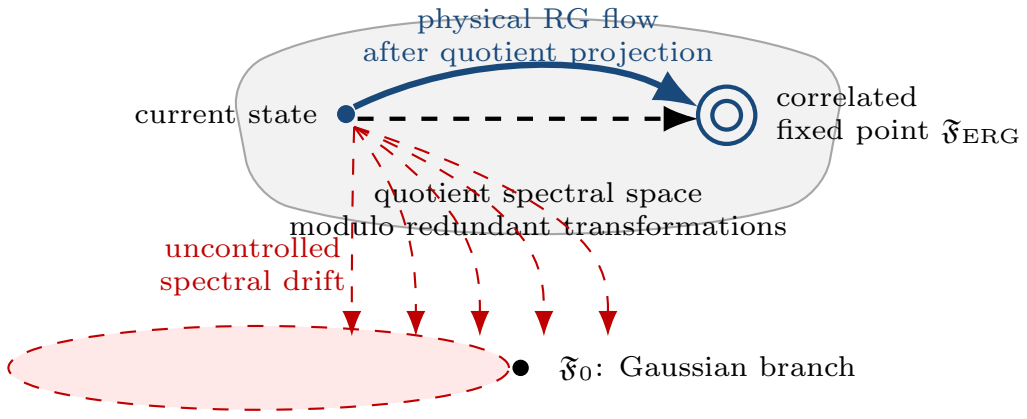

\begin{tikzpicture}[x=1cm,y=1cm,>=Latex,scale=.77,transform shape]
  \path[draw=gray!70,fill=gray!10,rounded corners=7pt]
    (-2.65,1.0) .. controls (-1.3,1.65) and (1.25,1.55) .. (2.65,1.05)
    -- (2.45,-.05) .. controls (1.0,-.45) and (-1.0,-.45) .. (-2.45,-.05) -- cycle;
  \node[font=\scriptsize,align=center] at (0,-.15) {quotient spectral space\\modulo redundant transformations};
  \fill[boxblue] (-1.65,.66) circle (2.2pt);
  \node[font=\scriptsize,anchor=east] at (-1.77,.66) {current state};
  \draw[very thick,boxblue,->] (-1.58,.72) .. controls (-.6,1.2) and (.6,1.15) .. (1.38,.72);
  \node[font=\scriptsize,boxblue,align=center] at (0,1.32) {physical RG flow\\after quotient projection};
  \draw[thick,dashed,->] (-1.55,.62) -- (1.38,.62);
  \draw[boxblue,thick] (1.62,.65) circle (7pt);
  \draw[boxblue,thick] (1.62,.65) circle (3.5pt);
  \node[font=\scriptsize,anchor=west,align=left] at (1.92,.65) {correlated\\fixed point $\Ferg$};
  \path[draw=red!75!black,dashed,fill=red!9] (-2.4,-1.52) ellipse (2.15 and .36);
  \fill (-.15,-1.52) circle (2pt);
  \node[font=\scriptsize,anchor=west] at (0.05,-1.52) {$\Fzero$: Gaussian branch};
  \foreach \x in {-1.6,-1.05,-.5,.05,.6} {
    \draw[red!75!black,dashed,->] (-1.58,.55) .. controls (\x,-.15) and (\x,-.7) .. (\x,-1.25);
  }
  \node[font=\scriptsize,red!75!black,align=center] at (-2.45,-.65) {uncontrolled\\spectral drift};
\end{tikzpicture}
\vspace{-0.6em}
\captionof{figure}{\textbf{Redundant directions in the spectral RG flow.} A raw update contains physical and redundant components. Projection by $\mathbf I-\Pred$ removes motion inside the redundant fiber and leaves the quotient-space flow toward $\Ferg$.}
\label{fig:rg_redundant_flow_abstract}
\end{center}
\vspace{-0.25em}

Let the abstract coarse-graining map be
\begin{equation}
  \Pi_{\rm RG}:\W\longmapsto[\Xtilde_{\Rell}],
\end{equation}
where brackets denote the equivalence class after gauge fixing. A redundant direction $\mathbf Z$ lies in the kernel of the linearized map,
\begin{equation}
  D\Pi_{\rm RG}(\W)[\mathbf Z]=0.
\end{equation}
Thus every infinitesimal update decomposes as
\begin{equation}
  \delta\W=\delta\W_{\rm phys}+\delta\W_{\rm red},
  \qquad \delta\W_{\rm red}\in\ker D\Pi_{\rm RG}(\W),
\end{equation}
and the quotient representative is
\begin{equation}
  \delta\W_{\rm RG}=(\mathbf I-\Pred)\delta\W.
\end{equation}
By construction, this removal must preserve the physical infinitesimal motion,
\begin{equation}
  D\Pi_{\rm RG}(\W)[\delta\W_{\rm RG}]
  =D\Pi_{\rm RG}(\W)[\delta\W].
\end{equation}

The trace-log condition provides a convenient gauge slice,
\begin{equation}
  T_{\Rell}(\W):=\Tr\log\Xtilde_{\Rell}=0.
\end{equation}
A projected vector must remain tangent to this slice to first order,
\begin{equation}
  DT_{\Rell}(\W)[\delta\W_{\rm RG}]=0,
\end{equation}
or equivalently $T_{\Rell}(\W+\epsilon\delta\W_{\rm RG})=O(\epsilon^2)$. This removes scale drift normal to the slice, but it is not the full quotient condition: retained-coordinate transformations can satisfy $DT_{\Rell}=0$ and still be redundant. Abstractly, if $\{\mathbf Z_\mu\}$ spans the chosen redundant subspace and
\begin{equation}
  H_{\mu\nu}=\langle\mathbf Z_\mu,\mathbf Z_\nu\rangle_F,
\end{equation}
then the local orthogonal projector is
\begin{equation}
  \Pred\delta\W
  =\sum_{\mu,\nu}\mathbf Z_\mu(H^{\dagger})_{\mu\nu}
    \langle\mathbf Z_\nu,\delta\W\rangle_F.
\end{equation}
Writing $\mathbf G_T:=\nabla_{\W}T_{\Rell}$, the normal trace-log component is
\begin{equation}
  \delta\W_{\log}
  =\frac{\langle\delta\W,\mathbf G_T\rangle_F}
  {\langle\mathbf G_T,\mathbf G_T\rangle_F}\,\mathbf G_T.
\end{equation}
Subtracting it produces a vector tangent to the gauge slice. This is necessary but not sufficient. If retained coordinates are rotated by
\begin{equation}
  \mathbf O(\epsilon)=e^{\epsilon\mathbf A},
  \qquad \mathbf A^{\top}=-\mathbf A,
\end{equation}
then
\begin{equation}
  \Xtilde_{\Rell}\longmapsto
  \mathbf O(\epsilon)^{\top}\Xtilde_{\Rell}\mathbf O(\epsilon),
  \qquad
  \delta\Xtilde_{\Rell}=[\Xtilde_{\Rell},\mathbf A].
\end{equation}
%This motion preserves both $\operatorname{spec}(\Xtilde_{\Rell})$ and $T_{\Rell}$, so it lies inside the gauge slice while remaining redundant.
%The quotient projector must therefore remove both the normal scale direction and the selected tangent generators of these isospectral orbits.
Subtracting the normal trace-log component makes the update tangent to the gauge slice:
\begin{equation}
  DT_{\Rell}(\W)[\delta\W-\delta\W_{\log}]=0.
\end{equation}
This correction can change eigenvalue ratios. The quotient projector separately removes the selected generators of isospectral motions,
\begin{equation}
  \delta\Xtilde_{\Rell}=[\Xtilde_{\Rell},\mathbf A],
  \qquad \mathbf A^{\top}=-\mathbf A,
\end{equation}
which preserve both the retained spectrum and $T_{\Rell}$.

The interpretation is geometric. The trace-log condition chooses a local section through a bundle of equivalent spectral descriptions; $\Pred$ removes the vertical component of the raw vector field; and $\delta\W_{\rm RG}$ is its horizontal representative. Fixed points should consequently be defined in the quotient: a representative may continue to move by a redundant transformation even when its physical RG state has stopped evolving. Conversely, satisfying $T_{\Rell}=0$ does not by itself identify a fixed point, because substantial physical shape flow can remain tangent to that slice.

This subsection states only that RG geometry. A finite optimizer step generally leaves the slice at second order and can change the retained support. It therefore requires a retraction, damping, acceptance safeguards, and recomputation of the ECS. Those implementation choices, including how the redundant basis is approximated without overwhelming the supervised optimizer trajectory, are deferred to Section~6.

\clearpage
\section{Optimizer Extensions}
\label{sxn:better_optimizers}

The base optimizer supplies the supervised training trajectory. The RG construction does not replace that dynamics; it acts on the completed optimizer step and nudges selected retained spectral coordinates associated with redundant RG motion. Any correction should therefore remain local, damped, and subordinate to the loss-driven update.

\begin{tcolorbox}[
  marginalbox,
  title=RG-aware correction,
  left=1.6mm,right=1.6mm,top=1.1mm,bottom=1.1mm,
  before skip=4pt,after skip=5pt
]
For each proposed optimizer update $\Delta\W_t$, select the working ECS
(i.e by the midpoint rule)
\eqtight{
  M_{\Rell_t}
  =
  \left\lfloor
  \frac{M_{{\rm PL},t}+M_{{\rm TL},t}}{2}
  \right\rfloor,
  \qquad
  \Rell_t=\{(1),\ldots,(M_{\Rell_t})\}.
}
Use the off-gauge residual
\eqtight{
  T_t(\W_t)
  =
  \Tr\log\overline{\X}_{\Rell_t}(\W_t)
}
to nudge the proposed update toward
\eqtight{
  \Tr\log\Xtilde_{\Rell_t}=0,
  \qquad
  \alpha_{\Rell_t}=2,
}
and remove redundent isospectral motion
\eqtight{
  \overline{\X}_{\Rell_t}
  \longmapsto
  \mathbf O^{\top}\overline{\X}_{\Rell_t}\mathbf O,
  \qquad
  \mathbf O(\epsilon)=e^{\epsilon\mathbf A},
  \qquad
  \mathbf A^{\top}=-\mathbf A.
}
Nudge the optimizer trajectory toward the RG flow.
\end{tcolorbox}

The construction summarized above has three parts. First, we select a working retained space between the trace-log ECS boundary and the fitted power-law boundary. The midpoint rule is a conservative finite-size choice: it avoids relying on either noisy boundary alone while retaining modes that are plausibly both correlated and described by the fitted tail. Second, the off-gauge trace-log residual measures motion in the retained log-volume direction. Contraction along this direction drives the correlated sector back toward the weak-tail trivial fixed point $\Fzero$; suppressing that component prevents this redundant return flow, while the spectral shape is nudged toward the marginal state $\alpha=2$. Finally, one would also like to remove isospectral rotations within the retained space. This step is harder because a rotation that leaves one layer's ESD unchanged need not leave the full network function unchanged unless neighboring layers transform compatibly.

The first operation is a one-dimensional projection fixed by the trace-log.  The second is a larger quotient-space problem: an isospectral rotation of one layer is not automatically redundant for the complete network.

\subsection{RG-guided optimizer flow}
\label{sxn:optimizer_general}

Let the base optimizer propose
\begin{equation}
  \widehat\W_{t+1}
  =
  \operatorname{Proposal}_t(\W_t,\mathcal H_t),
  \qquad
  \Delta\W_t=\widehat\W_{t+1}-\W_t,
  \label{eq:base_optimizer_proposal}
\end{equation}
where $\mathcal H_t$ is its internal state.  The RG extension acts on this finite displacement after momentum, preconditioning, clipping, weight decay, and the other microscopic optimizer rules have already been applied.  Its purpose is not to replace the supervised optimizer, but to nudge the proposed displacement and the resulting optimizer trajectory toward the candidate spectral RG flow.

\paragraph{1. Select a working ECS and measure its off-gauge trace-log.}
At step $t$, select a working ECS candidate $\Rell_t$ from the spectrum of the current layer (or, in a finite-step implementation, from the base proposal), and let $\mathbf V_{\Rell_t}$ contain its retained right singular vectors.  This selection identifies the retained space on which the RG diagnostic is evaluated.  It does \emph{not} assert that the retained operator already satisfies the trace-log condition.  During the local differential calculation, $\Rell_t$ and $\mathbf V_{\Rell_t}$ are held fixed.  The equations below linearize at $\W_t$; a post-proposal implementation may instead evaluate the same residual and gradient at $\widehat\W_{t+1}$.  Neither choice assumes that the evaluated state lies on the trace-log target.

Define the pre-gauge retained operator
\begin{equation}
  \overline{\X}_{\Rell_t}(\W)
  :=
  \mathbf V_{\Rell_t}^{\top}
  \left(\frac{1}{N}\W^{\top}\W\right)
  \mathbf V_{\Rell_t},
  \qquad
  M_{\Rell_t}=|\Rell_t|,
  \label{eq:optimizer_raw_retained_operator}
\end{equation}
and its trace-log residual
\begin{equation}
  T_t(\W)
  :=
  \Tr\log\overline{\X}_{\Rell_t}(\W),
  \qquad
  \mathcal G_t
  :=
  \{\W:T_t(\W)=0\}.
  \label{eq:optimizer_trace_log_coordinate}
\end{equation}
The set $\mathcal G_t$ is the candidate trace-log target for the fixed working support.  The actual retained state generally lies off this target:
$T_t(\W_t)$ and $T_t(\widehat\W_{t+1})$ need not vanish.  Their nonzero values are the control signal used to measure how the external optimizer trajectory departs from the proposed RG flow.

Writing
\begin{equation}
  \mathbf G_{T,t}
  :=
  \nabla_{\W}T_t(\W_t),
  \label{eq:optimizer_trace_log_normal}
\end{equation}
the fixed-support gradient is
\begin{equation}
  \mathbf G_{T,t}
  =
  \frac{2}{N}\,
  \W_t\mathbf V_{\Rell_t}
  \overline{\X}_{\Rell_t}(\W_t)^{-1}
  \mathbf V_{\Rell_t}^{\top},
  \label{eq:optimizer_trace_log_gradient_explicit}
\end{equation}
provided the retained operator is positive definite.  Because $\overline{\X}_{\Rell_t}$ is the pre-gauge operator rather than a pointwise normalized one, $T_t$ is not identically zero and $\mathbf G_{T,t}$ is generally nonzero.  If $\langle\mathbf G_{T,t},\mathbf G_{T,t}\rangle_F$ falls below a numerical tolerance, the correction is skipped or regularized.

\paragraph{2. Nudge the proposed update toward the trace-log target.}
Let
\begin{equation}
  d_t
  :=
  DT_t(\W_t)[\Delta\W_t]
  =
  \langle\mathbf G_{T,t},\Delta\W_t\rangle_F.
  \label{eq:optimizer_trace_log_directional_derivative}
\end{equation}
A tangent projection would impose $DT_t[\Delta\W]=0$ and would therefore preserve the current residual to first order.  That is not the desired operation when $T_t(\W_t)\neq0$.  Instead, choose a tracking rate $0\leq\gamma_{T,t}\leq1$ and replace the normal component of the base update by a component that reduces the current residual:
\begin{equation}
\begin{aligned}
  \Delta\W_t^{\rm track}
  &:=
  \Delta\W_t
  -
  \frac{
    d_t+\gamma_{T,t}T_t(\W_t)
  }{
    \langle\mathbf G_{T,t},\mathbf G_{T,t}\rangle_F
  }
  \mathbf G_{T,t},\\
  DT_t(\W_t)[\Delta\W_t^{\rm track}]
  &=-\gamma_{T,t}T_t(\W_t).
\end{aligned}
  \label{eq:minimal_trace_log_projection}
\end{equation}
Thus the correction does not assume that the current retained ECS is already on the trace-log surface.  It uses the off-gauge residual to steer the optimizer trajectory toward that surface.  When $T_t(\W_t)=0$, Eq.~\eqref{eq:minimal_trace_log_projection} reduces to the ordinary tangent projection as a limiting case.

Early in training it may be preferable to use only a one-sided branch-protection rule rather than active tracking.  With
\begin{equation}
  a_t=
  \frac{d_t}
       {\langle\mathbf G_{T,t},\mathbf G_{T,t}\rangle_F},
  \qquad
  a_t^-:=\min(a_t,0),
  \label{eq:one_sided_volume_coefficient}
\end{equation}
use
\begin{equation}
  \Delta\W_t^{\not\rightarrow\Fzero}
  =\Delta\W_t-a_t^-\mathbf G_{T,t}.
  \label{eq:one_sided_volume_projection}
\end{equation}
This cancels only first-order contraction of the retained log-volume; updates that expand it are unchanged.  It is a safeguard against motion toward $\Fzero$, not an assertion that $T_t=0$.

\paragraph{3. Remove only validated isospectral coordinate motion.}
The trace-log correction above is a steering term.  It should be distinguished from quotienting genuinely redundant coordinate motion.
Let $\mathbf A^{\top}=-\mathbf A$ and $\mathbf O(\epsilon)=e^{\epsilon\mathbf A}$.  Then
\begin{equation}
  \overline{\X}_{\Rell_t}
  \longmapsto
  \mathbf O(\epsilon)^{\top}
  \overline{\X}_{\Rell_t}
  \mathbf O(\epsilon),
  \qquad
  \delta\overline{\X}_{\Rell_t}
  =[\overline{\X}_{\Rell_t},\mathbf A].
  \label{eq:optimizer_isospectral_orbit}
\end{equation}
This motion preserves the retained eigenvalues, the ESD, and $T_t$, whether or not $T_t$ is zero.  It is therefore tangent to a fixed trace-log level set.  Removing it requires tangent generators in addition to the scalar trace-log tracking correction.

For a selected weight-space basis $\{\mathbf Z_{A_j,t}\}_{j=1}^{m_t}$ of candidate isospectral generators, define
\begin{align}
  H_{ij,t}
  &=\langle\mathbf Z_{A_i,t},\mathbf Z_{A_j,t}\rangle_F,
  \notag\\[-0.2em]
  \mathbf P_{{\rm iso},t}\Delta\W
  &=
  \sum_{i,j=1}^{m_t}
  \mathbf Z_{A_i,t}(H_t^+)_{ij}
  \langle\mathbf Z_{A_j,t},\Delta\W\rangle_F.
  \label{eq:optimizer_full_projector}
\end{align}
The RG-nudged update is then
\begin{equation}
  \Delta\W_t^{\rm RG}
  =
  (\mathbf I-\mathbf P_{{\rm iso},t})
  \Delta\W_t^{\rm track}.
  \label{eq:operational_projected_update}
\end{equation}
Because an exact isospectral generator satisfies
$DT_t(\W_t)[\mathbf Z_{A_j,t}]=0$, this quotient step preserves the first-order trace-log tracking condition in Eq.~\eqref{eq:minimal_trace_log_projection}.

This construction is potentially implementable, but harder for three reasons.  The full isospectral tangent space has dimension $M_{\Rell_t}(M_{\Rell_t}-1)/2$; its generators can be nearly dependent; and rotating the retained coordinates of one layer generally changes the inputs presented to the next layer.  Such a direction is redundant for the network only when neighboring layers transform compatibly or a direct functional test shows that the network output is unchanged.  A practical approximation therefore samples only a small, well-conditioned family with
\begin{equation}
  m_t\ll M_{\Rell_t}^{2},
  \label{eq:sampled_redundant_basis}
\end{equation}
and retains only generators whose effect is stable under resampling.

The tangent part needs an additional network-level test.  An ESD regards retained rotations as equivalent, but the full network generally does not: a right-coordinate rotation of one layer changes the coordinates passed to the next.  A sampled generator should therefore be paired with a compensating neighboring-layer transformation or rejected unless, on a small calibration batch $\{x\}$,
\begin{equation}
  \frac{
    \|f_{\W+\epsilon\mathbf Z_{A,t}}(x)-f_{\W}(x)\|
  }{
    \epsilon\|\mathbf Z_{A,t}\|_F
  }
  \leq\varepsilon_{\rm func}.
  \label{eq:functional_redundancy_test}
\end{equation}
Only generators that pass this functional check should enter the sampled projector.

\paragraph{4. Take a finite step, measure the residual, and optionally retract partway.}
A damped finite candidate is
\begin{equation}
  \W_{t+1}^{\rm cand}
  =
  \W_t+\rho_t\Delta\W_t^{\rm RG},
  \qquad 0<\rho_t\leq1.
  \label{eq:damped_projected_candidate}
\end{equation}
For the fixed working support, define the base and corrected residuals
\begin{equation}
  \tau_t^{\rm base}
  :=T_t(\widehat\W_{t+1}),
  \qquad
  \tau_t^{\rm cand}
  :=T_t(\W_{t+1}^{\rm cand}).
  \label{eq:finite_trace_log_residual}
\end{equation}
The corrected residual is generally nonzero.  Indeed, that is expected away from the endpoint: the purpose of the extension is to nudge the optimizer trajectory toward the RG target, not to assume that every intermediate retained ECS already lies on it.  At fixed support, the first-order relation is
\begin{equation}
  T_t(\W_{t+1}^{\rm cand})
  =
  \bigl(1-\rho_t\gamma_{T,t}\bigr)T_t(\W_t)
  +\mathcal O(\rho_t^2).
  \label{eq:trace_log_tracking_rate}
\end{equation}

If a finite spectral restoration is desired, let $\overline\lambda_i$ denote the retained eigenvalues of $\overline{\X}_{\Rell_t}(\W_{t+1}^{\rm cand})$ and define
\begin{equation}
\begin{aligned}
  \X_{\Rell_t}^{(\kappa_t)}
  &:={}
  \exp\!\left(
    -\kappa_t\frac{\tau_t^{\rm cand}}{M_{\Rell_t}}
  \right)
  \overline{\X}_{\Rell_t}(\W_{t+1}^{\rm cand}),\\
  \lambda_i^{(\kappa_t)}
  &:={}
  \overline\lambda_i
  \exp\!\left(
    -\kappa_t\frac{\tau_t^{\rm cand}}{M_{\Rell_t}}
  \right),
  \qquad 0\leq\kappa_t\leq1.
\end{aligned}
  \label{eq:partial_trace_log_restore}
\end{equation}
Then
\begin{equation}
  \Tr\log\X_{\Rell_t}^{(\kappa_t)}
  =
  (1-\kappa_t)\tau_t^{\rm cand}.
  \label{eq:partial_trace_log_residual}
\end{equation}
Thus $\kappa_t<1$ is a partial nudge and does not place the actual retained operator on the trace-log surface.  Only the exact retraction $\kappa_t=1$ produces the gauge-fixed representative, for which one may write
\begin{equation}
  \Xtilde_{\Rell_t}
  :=
  \X_{\Rell_t}^{(1)},
  \qquad
  \Tr\log\Xtilde_{\Rell_t}=0.
  \label{eq:exact_trace_log_retraction}
\end{equation}
The tilde is therefore reserved for the result of an exact retraction (or the limiting fixed-point representative), not for the pre-gauge operator that is differentiated during training.

After the finite correction, the ECS is recomputed.  Define the post-correction raw residual on the recomputed support by
\begin{equation}
  \tau_t^+
  :=
  \Tr\log\overline{\X}_{\Rell_t^+}(\W_{t+1}^{\rm cand}).
  \label{eq:recomputed_trace_log_residual}
\end{equation}
The correction should be accepted only when it remains local.  For minibatch loss $L_t$ and recomputed support $\Rell_t^+$, useful checks are
\begin{equation}
\begin{split}
  L_t(\W_{t+1}^{\rm cand})
  &\leq L_t(\widehat\W_{t+1})+\delta_{L,t},\\
  \frac{|\Rell_t^+\cap\Rell_t|}
       {|\Rell_t^+\cup\Rell_t|}
  &\geq s_{\min},
  \qquad
  |\tau_t^+|\leq|\tau_t^{\rm base}|+\delta_{T,t}.
\end{split}
\label{eq:optimizer_acceptance_tests}
\end{equation}
Failure of any test reduces $\rho_t$, $\gamma_{T,t}$, or $\kappa_t$, or skips the RG correction.  These safeguards prevent the spectral nudge from overwhelming supervised descent or jumping across a moving ECS boundary.

Operationally, the proposed extension is
\begin{equation}
\begin{aligned}
  \text{Propose}
  &\rightarrow \text{select }\Rell_t
  \rightarrow \text{measure }T_t,\\[-0.15em]
  &\rightarrow \text{nudge}
  \rightarrow \text{validated quotient}
  \rightarrow \text{recompute}.
\end{aligned}
  \label{eq:optimizer_pipeline}
\end{equation}
The target surface must not be confused with the actual retained ECS encountered along training.  At a generic optimizer step,
\begin{equation}
  \Tr\log\overline{\X}_{\Rell_t}(\W_t)\neq0
  \qquad\text{in general}.
  \label{eq:actual_ecs_off_gauge}
\end{equation}
That nonzero residual is the point of the construction: it tells the RG extension how to alter $\Delta\W_t$ so that the external optimizer trajectory moves toward, rather than being assumed to lie on, the candidate RG flow.  The zero trace-log condition is imposed only by an exact retraction or reached asymptotically at the proposed fixed point.

A staged implementation would monitor the raw trace-log residual, the overlap of successive ECS supports, the supervised loss, and the functional residual of each sampled generator.  Early in training, only the one-sided contraction safeguard may be used.  After the retained support stabilizes, $\rho_t$, $\gamma_{T,t}$, and $\kappa_t$ may be increased gradually and a small isospectral basis may be sampled.  If any diagnostic deteriorates, the method falls back to the raw optimizer proposal.  Muon and WW-PGD below are therefore partial approximations, not implementations of the complete trace-log-tracking plus validated-isospectral quotient construction.

\newpage
\subsection{Muon: a crude approximation}
\label{sxn:muon_method}

Muon is a matrix-aware optimizer for hidden-layer weights~\cite{JordanMuon2024,JordanBlog,LiuMuonScalable2025}.  It maintains a momentum-like matrix update and approximately orthogonalizes it with a short Newton--Schulz iteration before applying the learning rate; non-matrix parameters are normally handled by AdamW.  The motivation in Keller Jordan's blog is spectral: ordinary adaptive updates can drive hidden-layer matrices toward undesirable low-rank structure, so Muon replaces the anisotropic singular spectrum of the raw update by an approximately isotropic one.  This is effective empirically, but it is not derived from the RG quotient geometry and does not identify which directions are actually redundant.

Let the momentum-like update have thin SVD
\begin{equation}
  \mathbf G_t
  =
  \mathbf U_t\boldsymbol\Sigma_t\mathbf V_t^{\top}.
  \label{eq:muon_update_svd}
\end{equation}
The polar idealization of Muon replaces it by
\begin{equation}
  \Delta\W_t
  =
  \operatorname{Ortho}(\mathbf G_t)
  \simeq
  \mathbf U_t\mathbf V_t^{\top}.
  \label{eq:muon_polar_update}
\end{equation}
Consequently,
\begin{equation}
\begin{aligned}
  \Delta\W_t^{\top}\Delta\W_t
  &\simeq
  \mathbf V_t\mathbf V_t^{\top},\\
  \operatorname{spec}_{+}
  (\Delta\W_t^{\top}\Delta\W_t)
  &=\{1,\ldots,1\}.
\end{aligned}
  \label{eq:muon_update_covariance}
\end{equation}
so the active update singular values share one scale and
\begin{equation}
  \operatorname{Tr}_{+}\log
  (\Delta\W_t^{\top}\Delta\W_t)
  \simeq 0.
  \label{eq:muon_update_trace_log}
\end{equation}
The learning rate remains outside this normalization,
\begin{equation}
  \W_{t+1}=\W_t-\eta_t\Delta\W_t.
  \label{eq:muon_layer_step}
\end{equation}

Muon therefore removes anisotropic scale and conditioning from the \emph{update}, which may suppress excessive motion along a few singular directions.  But it does not select an ECS, compute the trace-log normal $\mathbf G_T$, identify motion toward $\Fzero$, remove tangent isospectral generators, or restore the trace-log of the layer state.  It can remove some redundant motion, but it also changes physical components because it does not know the quotient decomposition.

Muon should thus be interpreted as a crude operational test: broad update isotropization can change the spectral route followed during training, but it is not an implementation of the sampled isospectral quotient in Eq.~\eqref{eq:optimizer_full_projector} and does not establish exact RG projection or critical slowing down.

\clearpage
\subsection{WW-PGD: Projected Gradient Descent}
\label{sxn:wwpgd_method}

WeightWatcher-guided Projected Gradient Dynamics (WW-PGD) is the first-pass implementation designed to test the operational RG idea more directly.  It does not compute the sampled isospectral quotient of Eq.~\eqref{eq:optimizer_full_projector}, and it does not attempt to remove the complicated isospectral directions tangent to fixed trace-log level sets.  Instead, the base optimizer proposes a finite layer matrix, after which WW-PGD modifies a selected spectral window.  The method removes or restores one scalar retained-volume coordinate and simultaneously nudges the remaining shape coordinates toward the marginal $\alpha\simeq2$ rank-order form.

In the branch picture, this scalar direction is the trace-log drift toward the chosen trace-log target.  One sign of that drift reduces retained correlated volume and points back toward the weak-tail or trivial branch $\Fzero$; the opposite sign can overconcentrate the retained spectrum.  WW-PGD controls this one-dimensional volume motion while leaving the harder isospectral quotient problem for future work.  It is therefore a partial test of the proposal, not a realization of the complete RG optimizer.

\begin{center}
\safeincludegraphics[width=\columnwidth]{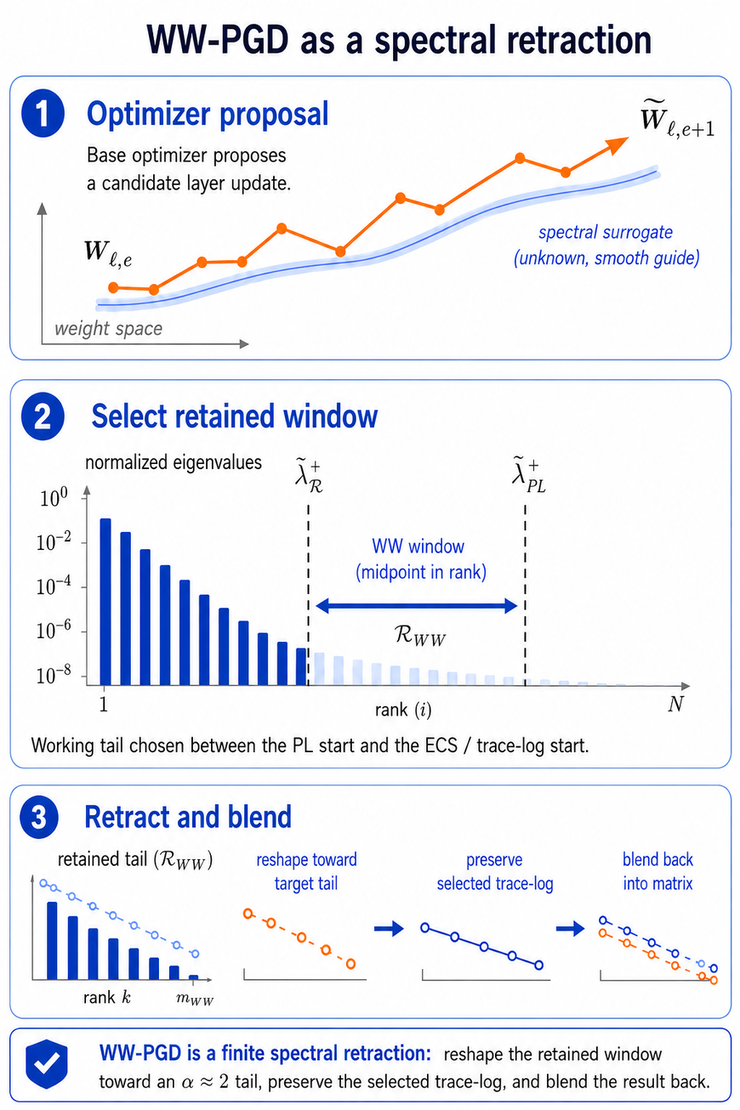}
\captionof{figure}{\textbf{WW-PGD as an approximate finite spectral correction.} The base optimizer proposes a candidate layer.  WW-PGD selects a window between the fitted PL start and the ECS/trace-log start, reshapes it toward an $\alpha\simeq2$ target, restores the selected trace-log before blending, and mixes the result back into the optimizer trajectory.}
\label{fig:wwpgd_optimizer}
\end{center}

For epoch or outer step $e$, write
\begin{equation}
  \widehat\W_{\ell,e+1}
  =
  \mathcal O_{\rm base}^{(e)}(\W_{\ell,e}),
  \qquad
  \W_{\ell,e+1}
  =
  \mathcal R_{\rm WW}(\widehat\W_{\ell,e+1}).
  \label{eq:wwpgd_outer_map}
\end{equation}
The SVD and spectral diagnostics of the proposed layer are then evaluated using the conventions already established earlier in the paper.  Let $m_{\rm PL}$ and $m_{\Rell}$ denote the numbers of eigenvalues retained by the fitted PL tail and by the trace-log ECS.  WW-PGD chooses the midpoint in rank space,
\begin{equation}
\begin{aligned}
m_{\mathrm{WW}}
&= \min\!\left\{
m_{\mathrm{PL}},
\left\lfloor \frac{m_{\mathrm{PL}}+m_R}{2} \right\rfloor
\right\}, \\
\Rell_{\rm WW}
&= \{(1),\ldots,(m_{\rm WW})\}.
\end{aligned}
\label{eq:wwpgd_midpoint_window}
\end{equation}
The correction is skipped when this window is too small or unstable.

Inside $\Rell_{\rm WW}$, rank $k$ is assigned the target
\begin{equation}
  \mu_k=k^{-q},
  \qquad q=1,
  \qquad \alpha_{\rm target}=1+q^{-1}=2.
  \label{eq:wwpgd_rank_target}
\end{equation}
The target $\tilde\lambda_k^{\star}=A\mu_k$ uses an amplitude that preserves the current selected trace-log,
\begin{equation}
  A
  =
  \exp\!\left[
  \frac{
  \sum_{i\in\Rell_{\rm WW}}\log\tilde\lambda_i
  -
  \sum_{k=1}^{m_{\rm WW}}\log\mu_k
  }{m_{\rm WW}}
  \right].
  \label{eq:wwpgd_target_amplitude}
\end{equation}
Thus the target changes the rank-order shape without independently imposing a new product scale.

The move is deliberately soft.  With homotopy $h_e\in[0,1]$, define
\begin{align}
  g_k
  &=
  \log\tilde\lambda_k-
  \log\tilde\lambda_k^{\star},
  \notag\\[-0.3em]
  r_k
  &=
  \operatorname{clip}\!\left(
  \frac{1-\eta_C h_e g_k}
       {1+\eta_C h_e g_k},
  r_{\min},r_{\max}
  \right),
  \label{eq:wwpgd_cayley_move}
\end{align}
with provisional values
\begin{equation}
  \widehat{\tilde\lambda}_k
  =
  \tilde\lambda_k r_k.
  \label{eq:wwpgd_provisional_eigenvalues}
\end{equation}
The selected trace-log is then restored before reconstruction,
\begin{align}
  \tilde\lambda_k^{\rm ret}
  &=
  \widehat{\tilde\lambda}_k
  \exp\!\left[
  \frac{
  T_{\rm WW}
  -
  \sum_{j=1}^{m_{\rm WW}}
  \log\widehat{\tilde\lambda}_j
  }{m_{\rm WW}}
  \right],
  \label{eq:wwpgd_trace_log_restore}\\
  T_{\rm WW}
  &=
  \sum_{i\in\Rell_{\rm WW}}
  \log\tilde\lambda_i.
  \label{eq:wwpgd_current_trace_log}
\end{align}
In log-eigenvalue coordinates, Eq.~\eqref{eq:wwpgd_trace_log_restore} removes exactly one scalar component: the uniform retained log-volume direction.  The remaining $m_{\rm WW}-1$ coordinates describe spectral shape.  This is the direct connection to the redundant-direction discussion: WW-PGD controls one normal gauge coordinate but does not project the tangent isospectral orbit.

The retained singular values are reconstructed through their scale-free ratios, producing $\W^{\rm shaped}$.  The final state is blended with the base proposal,
\begin{equation}
  \W^{+}
  =
  (1-\eta_B h_e)\widehat\W
  +
  \eta_B h_e\W^{\rm shaped}.
  \label{eq:wwpgd_final_blend}
\end{equation}
The restoration in Eq.~\eqref{eq:wwpgd_trace_log_restore} is exact on the selected shaped spectrum before this final blend.  The complete WW-PGD update is intentionally approximate: after blending, the selected trace-log generally has a residual.  This is a feature of the first-pass design.  The method nudges the optimizer toward the proposed RG trajectory without forcing every intermediate checkpoint exactly onto the final trace-log slice.

A useful diagnostic is the post-blend residual
\begin{equation}
  \varepsilon_T^{+}
  =
  \left|
  \sum_{i\in\Rell_{\rm WW}}
  \log\tilde\lambda_i(\W^{+})
  -T_{\rm WW}
  \right|.
  \label{eq:wwpgd_postblend_residual}
\end{equation}
Together with the supervised loss, the stability of the selected window, and the movement of the PL and ECS boundaries, this residual can be used to adapt $\eta_B$, $\eta_C$, and the homotopy schedule.  If the correction damages the optimizer trajectory, it should be weakened or skipped.  Near a stable endpoint, it may be tightened.

WW-PGD is therefore the first operational experiment on the proposed RG extension.  It tests whether removing one retained-volume degree of freedom and steering the remaining spectral shape toward the marginal target can improve training without imposing the full quotient geometry.  A more complete optimizer would add the local update-space quotient of Eq.~\eqref{eq:optimizer_full_projector}, including carefully selected isospectral generators, while retaining the same safeguards against trajectory distortion.

\clearpage
\section{Experimental Results}
\label{sxn:experiments}

\subsection{Muon test}
\label{sxn:muon_empirical_test}

\begin{figure*}[!b]
\centering
\includegraphics[width=0.8\textwidth]{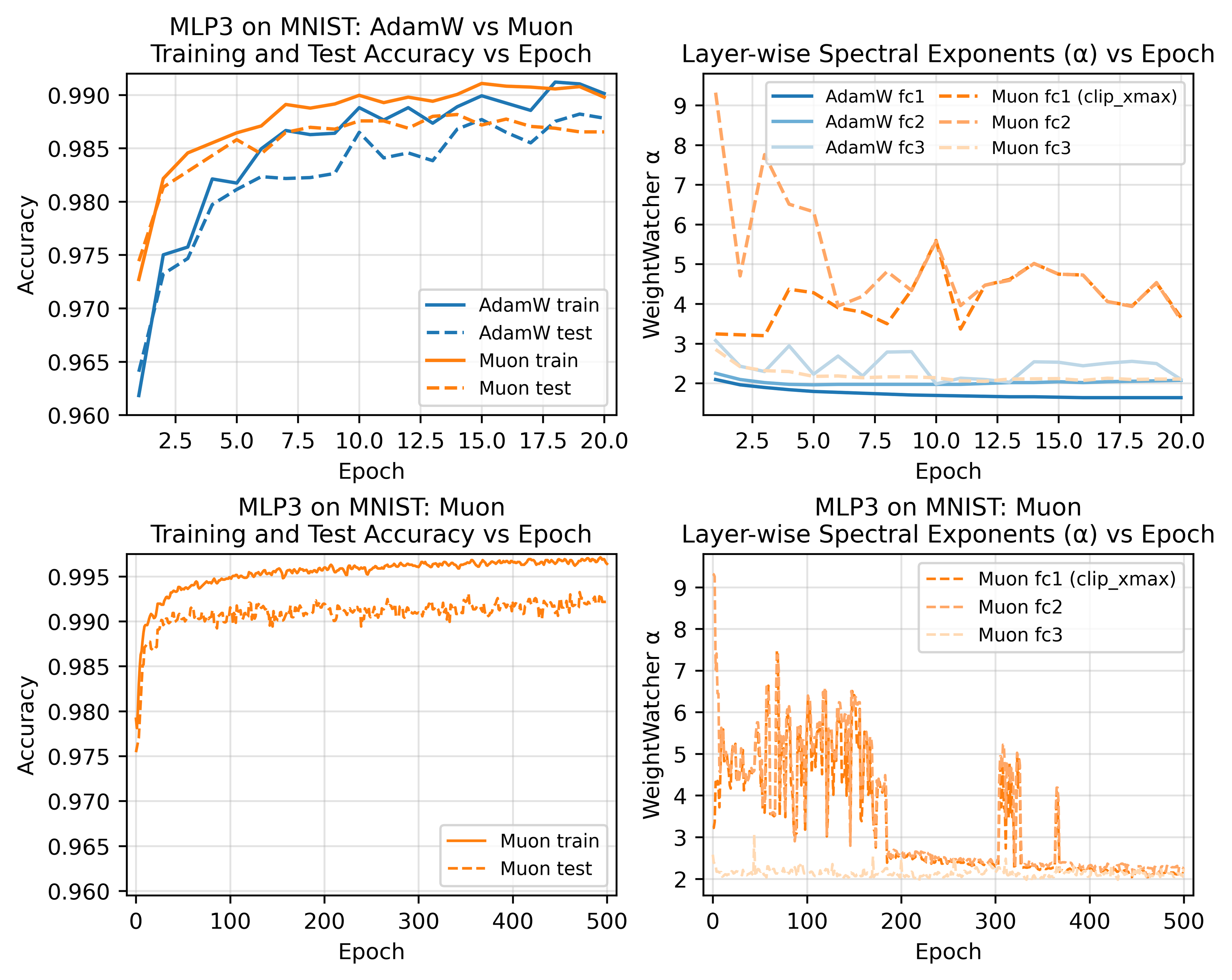}
\caption{\textbf{MNIST optimizer comparison: AdamW versus Muon.}
Same three-layer MNIST MLP trained with AdamW and Muon.
Short-horizon panel compares early epochs; long-horizon panel tracks the Muon $\alpha_{\ell}(t)\rightarrow 2$.}
\label{fig:mnist_optimizer_comparison}
\end{figure*}

%\begin{tcolorbox}[marginalbox,title=\textbf{14. Muon test}]
%\begin{itemize}
%\item \textbf{Compare:} AdamW and Muon.
%\item \textbf{Model:} three-layer MNIST MLP.
%\item \textbf{Measure:} layerwise \(\alpha(t)\).
%\end{itemize}
%\takeaway{Muon changes the spectral route.}
%\end{tcolorbox}

The Muon test asks whether update geometry changes the spectral route.  The controlled model is a three-layer MNIST MLP, described
in the Appendix~\ref{app:optimizer_experiments}.
%The comparison holds the data, model, and diagnostics fixed.
%Only the matrix-valued update rule changes: AdamW~\cite{LoshchilovHutter2019} is compared to Muon.  At checkpoints, \WW{} fits the HTSR exponent of each fully connected layer
The comparison uses the same data and model for AdamW~\cite{LoshchilovHutter2019} and Muon.
For FC1, the reported exponents use the standard ~\WW fit for AdamW and \texttt{clip\_xmax} for Muon.
\begin{equation}
        \alpha_\ell(t),
        \qquad
        \ell\in\{\mathrm{FC1},\mathrm{FC2},\mathrm{FC3}\}.
\end{equation}

Figure~\ref{fig:mnist_optimizer_comparison} compares results for Muon against AdamW~\cite{LoshchilovHutter2019};
it depicts the test and training accuracies and the layerwise $\alpha_{\ell}(t)$ per epoch.
The top panels compare AdamW and Muon over the first twenty $(20)$ epochs, while the bottom panels
track Muon for five-hundred $(500)$ epochs.

Consider the top panels first.
AdamW reaches high train and test accuracy more slowly than Muon, but reaches a comparable range after twenty $(20)$ epochs.
The AdamW and Muon layerwise $\alpha_\ell(t)$ curves behave very differently over these early epochs.
For AdamW, all layer $\alpha_\ell(t)$ values drop quickly. FC2 and FC3 approach the optimal value $\alpha=2$,
but FC1 immediately collapses below $2.0$ and remains trapped in the proposed overfitting regime, even as accuracy increases.
In contrast, only Muon's FC3 layer approaches $\alpha=2$ quickly,
while FC1 and FC2 fluctuate at first, and then remain near $\alpha=4$.

The bottom panels show Muon over very long-horizon runs.
Neither Muon training accuracy nor test accuracy drops; both increase slowly and asymptotically.
Muon FC1 and FC2 track each other, first fluctuating strongly with $\alpha_\ell(t)\in[3,6]$.
Eventually, however, all layer values satisfy $\alpha_\ell(t)\rightarrow 2$, again very slowly.

This comparison is intentionally untuned. We made no attempt to optimize learning rates or make Muon win the MNIST benchmark.

AdamW immediately drives the FC1 $\alpha<2$, while Muon approaches the proposed RG fixed  point
asymptotically slowly.  This is consistent with Muon removing the
 RG redundant scale directions, and could be associated with RG \emph{critical slowing down}.
\clearpage
\newpage
\subsection{WW-PGD test}
\label{sxn:wwpgd_empirical_test}

%\begin{tcolorbox}[marginalbox,title=\textbf{15. WW-PGD repair tests}]
%\begin{itemize}
%\item \textbf{Question:} Can WW-PGD repair overfit models?
%\item \textbf{Test 1:} Fix the AdamW FC1 overfit layer so $\alpha\simeq 2$.
%\item \textbf{Test 2:} Fix an anti-grokking checkpoint, recover the test accuracy.
%\end{itemize}
%\takeaway{WW-PGD can both drive $\alpha\simeq 2$ and recover lost generalization accuracy.}
%\end{tcolorbox}

\paragraph{Test 1: Fixing AdamW runs.} 
The WW-PGD test asks whether the optimizer can reduce AdamW overfitting by preventing the layer exponents
$\alpha_\ell(t)$ from dropping below $2$, while simultaneously helping all layers approach the RG fixed point
$\alpha\simeq 2$. The experiment is similar to the Muon test (Section~\ref{sxn:muon_empirical_test}), and is described in
Appendix~\ref{app:optimizer_experiments}.

The WW-PGD condition uses the same AdamW minibatch optimizer,
but applies the epoch-boundary retraction in Eq.~\eqref{eq:wwpgd_outer_map}.
The experiment runs for 35 epochs and five independent runs. The WW-PGD hyperparameters are
$(q=1;\;\eta_B=0.5;\;\eta_C=0.25)$ with zero warmup and a five-epoch homotopy ramp.

\begin{figure}[t]
  \centering
\includegraphics[width=\linewidth]{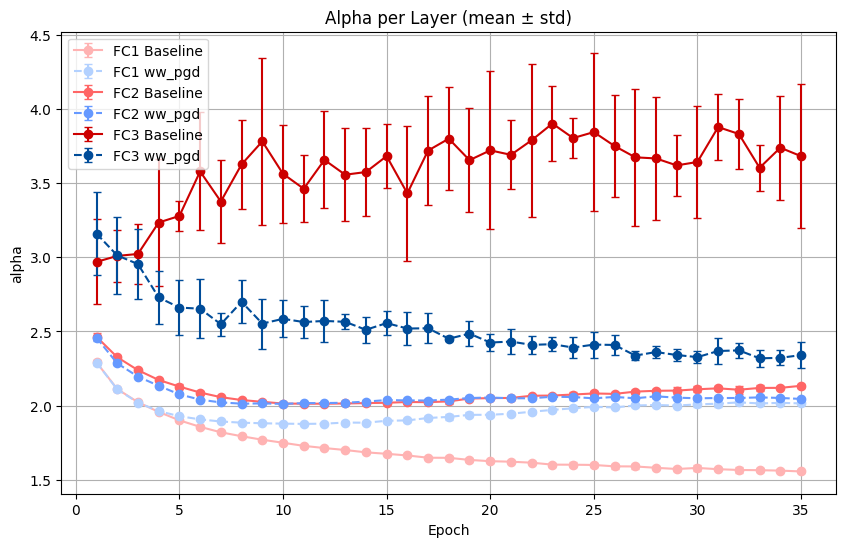}\\[-0.35em]
\includegraphics[width=\linewidth]{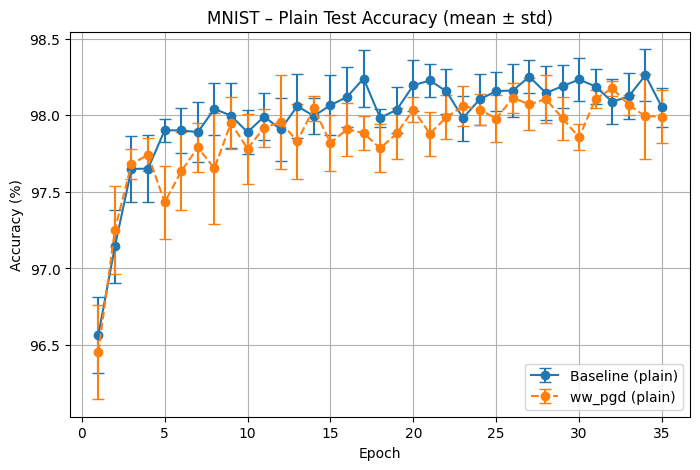}
\caption{\textbf{WW-PGD stabilizes layerwise HTSR exponents while preserving test accuracy.}
A three-layer MNIST MLP is trained with AdamW alone or AdamW plus WW-PGD. Top: layerwise fitted HTSR exponents for FC1--FC3. Bottom: ordinary MNIST test accuracy. Curves show mean \(\pm\) standard deviation across five runs.}
\label{fig:wwpgd_alpha_per_layer}
\end{figure}

Figure~\ref{fig:wwpgd_alpha_per_layer} shows the main effect.
At each epoch, \WW{} logs the HTSR exponent $\alpha_\ell(t)$.
The plotted curves report the mean and standard deviation for each layer across five runs
(FC1, FC2, FC3).

Figure~\ref{fig:wwpgd_alpha_per_layer} shows the layerwise results.
AdamW alone lets the $\alpha_\ell(t)$ exponents separate by layer.
As in the Muon comparison, the FC1 exponent satisfies $\alpha_\ell(t)<2$ throughout most of training, while the FC2 exponent satisfies $\alpha_\ell(t)\geq2$.
FC3  satisfies $\alpha_\ell(t)\gg2$; however, this does not materially change the results.

WW-PGD improves the layer $\alpha_\ell(t)$ for all layers, while keeping the training and test accuracies close to, or only slightly below, the baseline AdamW results.

The convergence to the RG optimal boundary is much faster than with Muon. This is the key takeaway.
This makes WW-PGD a direct finite-step test of the spectral-RG optimizer hypothesis.

\begin{figure}[!h]
  \centering
\includegraphics[width=\linewidth]{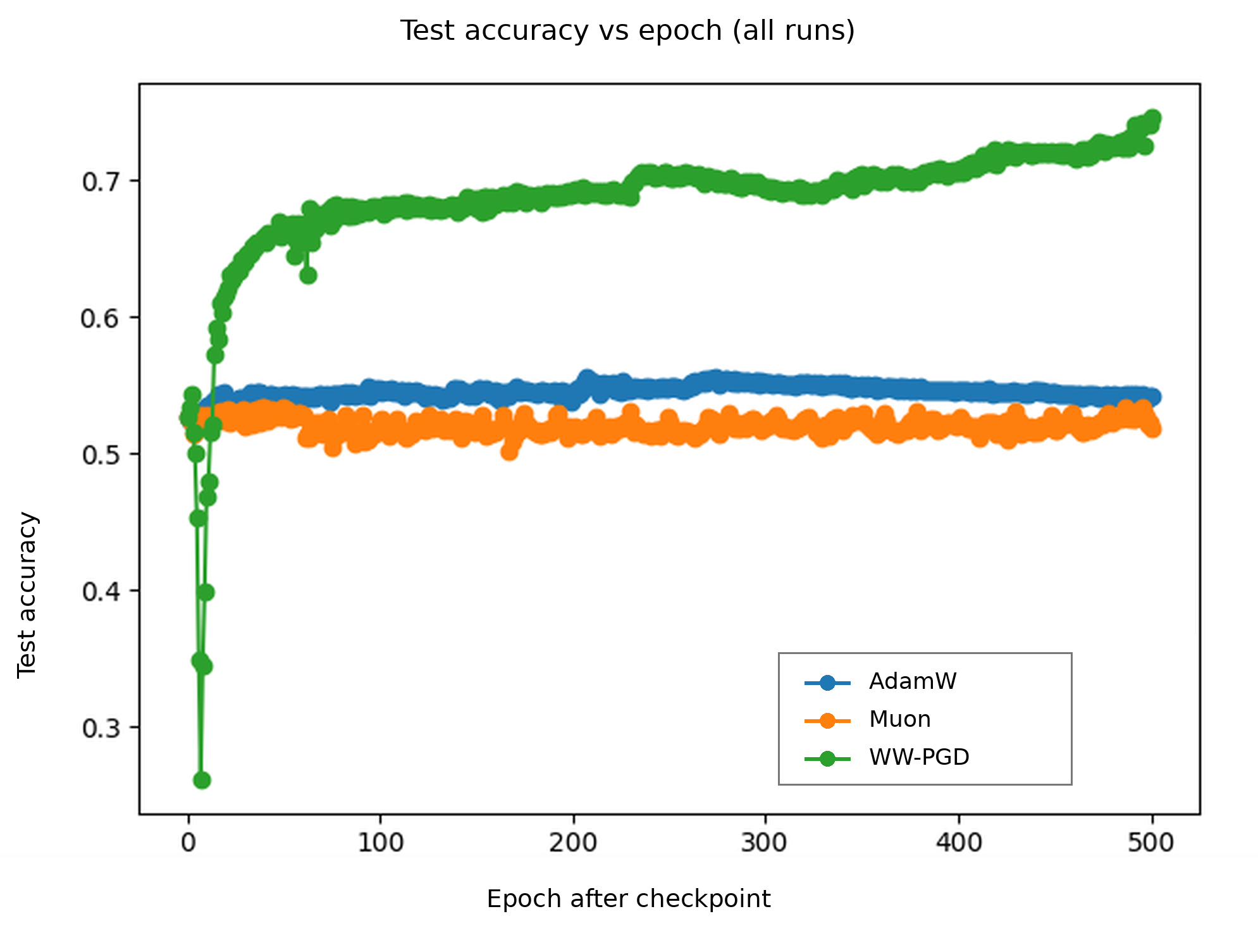}
\caption{\textbf{WW-PGD corrects overfit layers.} Extending training on an anti-grokking checkpoint from Figure~\ref{fig:rmt_v1_long}.~\cite{PrakashMartin2026Ongoing}}
\label{fig:ww_fixed_grok}
\end{figure}

\paragraph{Test 2: Fixing an anti-grokking checkpoint.}
As a further stress test, we test whether WW-PGD can repair a deeply overfit checkpoint.
Starting from an anti-grokking checkpoint shown in
Figure~\ref{fig:rmt_v1_long}~\cite{PrakashMartinLateCollapse2026}, we resume training using AdamW, Muon, and WW-PGD.
Figure~\ref{fig:ww_fixed_grok} depicts preliminary experiments~\cite{PrakashMartin2026Ongoing}, showing
that the WW-PGD optimizer can recover the test accuracy even after long-horizon overfitting.

Together, these two experiments provide preliminary evidence that WW-PGD repairs
the overfit checkpoints tested here while driving all layer $\alpha_\ell(t)\simeq 2$.

\clearpage
\section{Implications of Universality}
\label{sec:implications}

\begin{figure}[!t]
\centering
\includegraphics[width=\columnwidth]{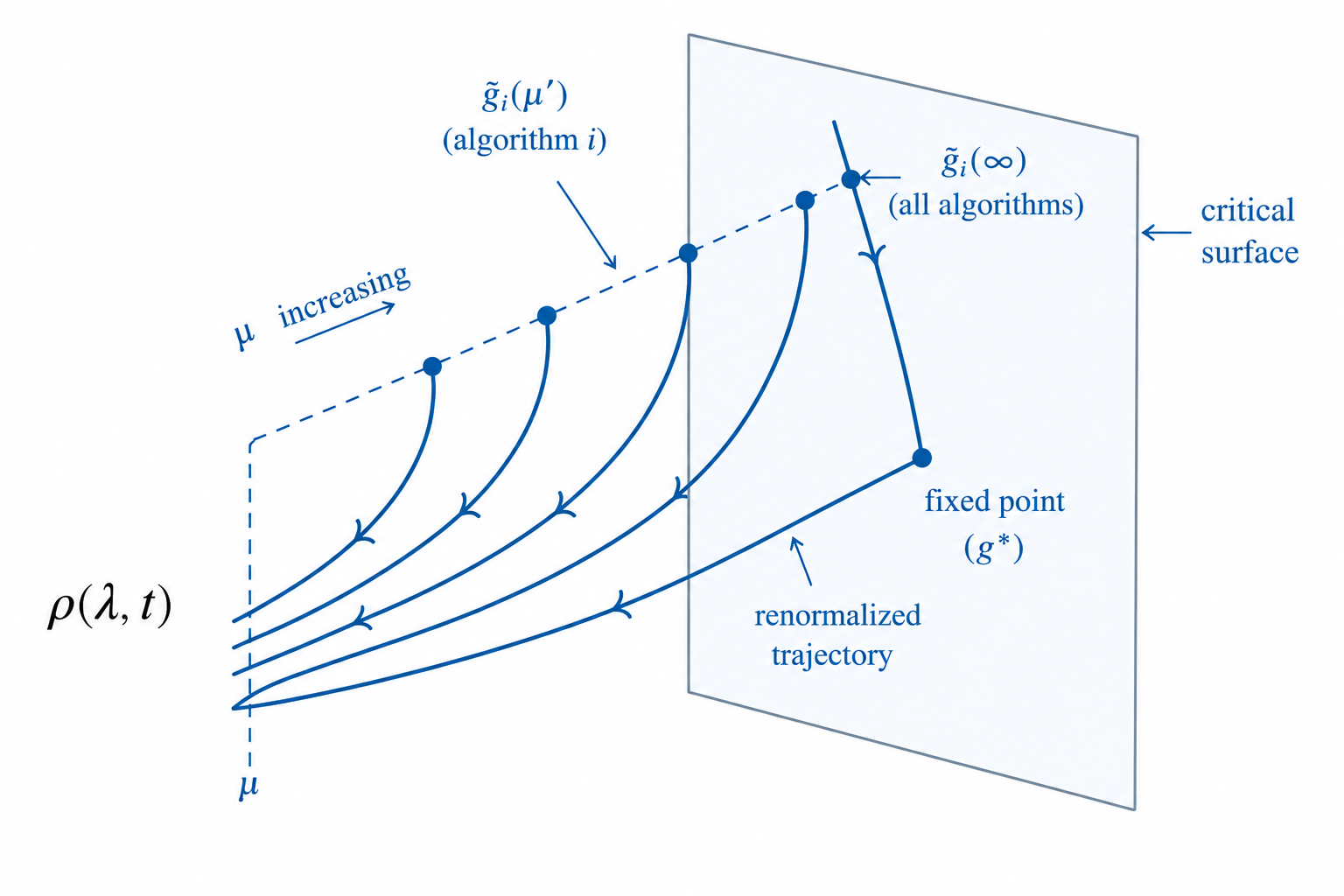}
\caption{RG flow in the space of effective correlations. Different nonlinear learning algorithms follow different microscopic trajectories, but coarse-graining can suppress irrelevant directions and carry retained spectral structure toward a candidate fixed point. Candidate universal quantities are read off near the fixed point, not from the microscopic update rule.}
\label{fig:rg-universality-flow}
\end{figure}

It is conjectured that neural networks, boosted trees, and related methods can be grouped under one universality story. They use different update rules, but each can induce an effective correlation structure on the data~\cite{GalluccioBouchaudPotters1998}. Figure~\ref{fig:rg-universality-flow} represents that structure by effective couplings $\widetilde g_i(\mu)$ that flow with RG scale $\mu$. The RG question is whether these spectra approach a common critical surface and candidate fixed point $g^\star$.

\begin{tcolorbox}[marginalbox,title=\textbf{Universality of learning algorithms}]
Universality asks whether different nonlinear learners can flow to the same coarse-grained candidate spectral fixed point.
\begin{itemize}
\item Nonlinear learning algorithms can be linearized near their fixed points.
\item Spectra, not parameters, may characterize the optimal conditions.
\item Self-averaging bulk modes support transfer.
\item Non-self-averaging modes reveal dominant atypical structure that causes overfitting.
\item XGBoost is a conjectural test case.
\end{itemize}
\takeaway{RG universality means different algorithms may share a candidate fixed-point spectrum.}
\end{tcolorbox}

The RG claim is local. A learner can be nonlinear away from convergence. Near a stable endpoint, its coarse-grained flow can be linearized in a few spectral coordinates $\widetilde g_i$. In Figure~\ref{fig:rg-universality-flow}, the renormalized trajectories bend toward the same critical surface. The stable directions define the universality class, and the fixed point $g^\star$ defines the universal spectral observables.

Near convergence, the invariant object is the spectrum of the retained correlation operator.
Parameters change with architecture, but the proposed ERG point may be universal. A non-neural learner needs its own effective correlation operator. Universality becomes testable only after that operator and its $\widetilde g_i$ coordinates are defined.

Self-averaging is another proposed natural universal property. The bulk spectral averages should stabilize across samples, initializations, and checkpoints.  Non-self-averaging modes should concentrate into a few dominant elements in the spectral tail.
Tail modes are not automatically bad. They become risky when a few directions carry too much spectral energy gain,
causing correlation traps to form and/or $\alpha < 2$.

Neural networks give the direct case. Different depths, widths, initializations, and optimizers can produce similar endpoint spectra. The shared object is the normalized effective correlation spectrum. This view also reframes optimizer design. A good optimizer should reduce loss while steering the spectrum toward a self-averaging marginal boundary between
optimality and overfitting, with $\alpha\simeq 2$ as a candidate universal spectral exponent.

XGBoost gives a plausible extension~\cite{ChenGuestrin2016}. Boosting has no dense layer weight matrix, so the current layer-specific theory does not apply directly. Still, each tree changes the residuals and the prediction function. Those updates induce a data-data correlation structure in prediction space. A practical test would form a centered Gram operator from tree contributions or prediction increments, normalize its scale, and track its spectrum across boosting rounds.
Early work (not shown) suggests this is possible. Future work will test this construction on XGBoost models.

Universality does not say that all algorithms compute the same function. It says that successful learners may share the same coarse-grained spectral organization. Generalization would then depend less on update details and more on self-averaging, marginality, and controlling the dominant modes to prevent overfitting or collapse.
More broadly, universality suggests transferability. In semi-empirical electronic-structure theory, transferable parameters work because different molecules can share the same effective valence-space structure. Martin and Freed tested this idea by deriving semiempirical $\pi$-electron parameters from ab initio effective valence-shell Hamiltonians and studying when those parameters transfer across related molecules \cite{MartinFreed1994I,Martin1996SchiffBases,Martin1996LargeSchiffBases}.
The analogy is illustrative. The SETOL effective free energy $\IZQ$ is evaluated by taking a single step of the
Wilsonian Exact RG (ERG) and is expressed in terms of the R-Transform of the layer, which is akin to a renormalized self-energy
and analogous to the more general effective Hamiltonian approach.
Moreover, if learned correlation spectra have candidate universal fixed-point structure, then spectral diagnostics may transfer across related models and tasks. Transferability would then come from the effective correlation space, not from the microscopic implementation.

\clearpage
\section{Conclusions}
\label{sxn:conclusions}

This paper proposes an RG theory of NN layer convergence, integrating the older phenomenological HTSR approach and the more recently developed
SETOL theory. HTSR supplies the empirical fact: strong layers often develop heavy-tailed spectra near $\alpha\simeq2$. SETOL supplies the trace-log condition: remove arbitrary layer scale and study the retained effective correlation space. The main claim is that these two pictures meet at the same spectral endpoint.

The value $\alpha=2$ has two meanings.
%Statistically, it marks the boundary between stable bulk-supported learning and domination by a few extreme eigencomponents. RG scale counting gives the same boundary.
We propose it as the boundary between stable bulk-supported learning and domination by a few extreme eigencomponents.
At $\alpha=2$, each logarithmic spectral shell contributes comparable spectral energy gain.
This agreement makes $\alpha\simeq2$ a critical boundary, not just a fitted number.

The fixed point is layerwise. Each dense layer has its own retained spectrum, trace-log scale, and dominant-tail burden. A trained network is therefore not one monolithic fixed point. It is a coordinated state in which many layers approach compatible spectral endpoints.

The practical diagnostic is spectral. A good layer should keep useful correlation gain spread across many retained eigencomponents. A bad layer can concentrate too much gain in a few dominant tail modes. The trace-log condition selects the ECS. The power-law tail measures shape.
The dominant-tail eigencomponents test whether the layer remains self-averaging.

The optimizer message is nuanced.
Raw optimizers can spend update norm on scale-only motion. A better optimizer should reduce loss while steering the retained spectrum toward the correlated, self-averaging branch. This view suggests why matrix-aware optimizers such as Muon are relevant. It also suggests new optimizer designs based on trace-log geometry.

The \WW{} message becomes sharper. The HTSR exponent $\alpha$, the trace-log coordinate, ECS--tail alignment, trap counts,
and the tail burden are not isolated heuristics.
They can be read as candidate RG observables of the same layerwise spectral flow.
They require the weights and the weights alone, without labels, gradients, or test or even training data.

The empirical program is direct.
Track these observables across checkpoints. Compare them across optimizers.
Ask whether held-out performance improves when the spectrum approaches the
marginal boundary and degrades when $\alpha<2$ and/or traps form.
This turns the RG picture into a falsifiable diagnostic program.

The limitations are also direct.
This paper does not prove convergence of SGD, AdamW, or Muon to the spectral RG fixed point.
It proposes an operational RG interpretation of retained layer spectra and connects that interpretation
to HTSR and SETOL theory and validated empirical diagnostics.
The evidence supports this picture on the models studied here, but it does not establish universality
across all architectures, datasets, optimizers, or nonlinear learners.  This remains to be explored.

The broader implication is transferability. Semi-empirical electronic-structure
theory uses transferable parameters because related molecules share effective valence-space structure.
The analogy here is direct. If learned correlation spectra have candidate universal fixed-point structure,
then useful spectral parameters may transfer across related models, tasks, optimizers, and algorithm families.

The final message is simple. Learning is not only loss minimization. Learning is spectral organization.
In this picture, strong layers move toward a RG correlated fixed point where scale is removed, the tail is balanced, and the
weights remain self-averaging.

\begin{tcolorbox}[marginalbox,title=\textbf{The theory in one pass}]
Dense layers learn by moving their spectra.
\begin{itemize}
\item HTSR observes heavy-tailed spectra near $\alpha\simeq2$.
\item SETOL connects heavy-tailed layer spectra to model accuracy (called layer quality), and yields the ERG trace-log laws.
\item Self-averaging helps explain why $\alpha=2$ is a useful boundary.
\item RG scale counting explains why $\alpha=2$ is marginal, suggesting a candidate universal spectral exponent.
\item Optimizers should reduce scale-only motion and move toward the ERG fixed point.
\item The WW-PGD optimizer extension can repair overfit models.
\item The theory may be extensible to XGBoost and related algorithms:
  $\alpha=2$ and $\Tr{\log\Xtilde_{\Rell}}=0$.
\end{itemize}
\takeaway{Normalize $\X$, Keep the learned tail and test whether the retained layer stays balanced.}
\end{tcolorbox}

Table~\ref{tab:notation-conventions} summarizes the notation and conventions used throughout.

\section*{Acknowledgements}

The author gratefully acknowledges Hari K. Prakash for valuable discussions, constructive feedback, guidance throughout this work, and for his help in developing the WW-PGD optimizer.  The author used ChatGPT to assist with selected illustrations, manuscript editing, and the exploration and refinement of arguments, and takes full responsibility for the final content.

\clearpage
\onecolumn
\thispagestyle{plain}
\begin{center}
\vspace*{-1.0em}
\setlength{\tabcolsep}{3pt}
\renewcommand{\arraystretch}{1.04}
\captionof{table}{Notation and conventions used in this report.}
\label{tab:notation-conventions}
\resizebox{\textwidth}{!}{%
\begin{tabular}{@{}lll@{}}
\toprule
Group & Symbol & Meaning \\
\midrule
Layer & $\W\in\mathbb R^{N\times M}$, $N\ge M$ & Dense-layer weights. \\
Layer & $\X=N^{-1}\W^T\W$ & Layer correlation operator. \\
Layer & $\lambda_i>0$, $\Splus=\{i:\lambda_i>0\}$ & Positive spectral support. \\
Layer & $\sigma(\W) = \sqrt{N\lambda(\X)}$, & Singular values of $\W$ (pre-rescaling). \\
Layer & $\Tr{\X}=M$, & First-moment rescaling. \\
PL tail & $\rho_{\rm tail}(\lambda)\sim\lambda^{-\alpha}$ & HTSR tail; fitted exponent $\alpha$. \\
PL tail & $\TPL$, $\TPL^\star$ & CSN-selected PL window. \\
ECS & $\Rell\subseteq\Splus$, $\Mred=|\Rell|$ & Retained ECS and size. \\
ECS & $\RECS^\star$ & Stabilized SETOL ECS. \\
Alignment & $\RECS^\star\simeq\TPL^\star$ & SETOL--HTSR window match. \\
Trace-log & $\XR=\V_{\Rell}^T\X\V_{\Rell}$ & ECS-restricted operator. \\
Trace-log & $\Tr\log\Xtilde_{\Rell}=\sum_{i\in\Rell}\log\tilde{\lambda}_{\Rell,i}=0$ & Scale-invariant condition. \\
%Energy & $E_+=M^{-1}\Tr\Xtilde$ & Retained spectral energy   \\
Energy & $\Ered=\Mred^{-1}\sum_{i\in\Rell}\tilde{\lambda}_{\Rell,i}=\int\tilde{\lambda}_{\Rell}\rhohat_{\Rell}(\tilde{\lambda}_{\Rell})d\tilde{\lambda}_{\Rell}$ & Retained spectral energy. \\
Tail & $\phi_k^R=\sum_{j=1}^k\tilde{\lambda}_{\Rell,R(j)}/\sum_{i\in\Rell}\tilde{\lambda}_{\Rell,i}$ & Top-$k$ gain fraction. \\
Tail & $M_{\rm tr}^R=(\sum_{i\in\Rell}\tilde{\lambda}_{\Rell,i})^2/\sum_{i\in\Rell}\tilde{\lambda}_{\Rell,i}^2$ & Effective contributor count. \\
Scale & $x=\log\tilde{\lambda}_{\Rell}$, $[\Lambda,b\Lambda]$ & Log coordinate and shell. \\
Scale & $G_E(\Lambda,b)=\int_{\Lambda}^{b\Lambda}\tilde{\lambda}_{\Rell}\rhohat_{\Rell}(\tilde{\lambda}_{\Rell})d\tilde{\lambda}_{\Rell}$ & Band energy gain. \\
Scale & $\displaystyle g_E(\tilde{\lambda}_{\Rell})=\frac{\dd\Ered}{\dd\log\tilde{\lambda}_{\Rell}}=\tilde{\lambda}_{\Rell}^2\rhohat_{\Rell}(\tilde{\lambda}_{\Rell})$;  & Spectral energy  gain per logarithmic scale. \\
Beta & $\beta_E(g_E)=d g_E/d\log\tilde{\lambda}_{\Rell}=(2-\alpha)g_E$ & Leading PL scale derivative. \\
Marginal & $y_E=2-\alpha=0$ & Log-band balance. \\
Fixed point & $\Rell_{\rm RG}[\rho^\star]=\rho^\star$ & Normalized spectral fixed point. \\
Branch & $\Fzero$, $\Ferg$ & RG spectral fixed points or branches. \\
Branch & $\Fzero:g_E\ \text{undefined}$ & Trivial/random-like branch. \\
Branch & $\Ferg:0<g_E^\star<\infty,\ \alpha\simeq2,\ \RECS^\star\simeq\TPL^\star$ & Empirical SETOL ERG fixed point or branch. \\
Quality & $\overline Q_{\rm NN}$, $\overline Q_\ell$ & Model and layer quality. \\
Optimizer & $A_{\Rell}=\V_{\Rell}\XR^{-1}\V_{\Rell}^T$, $G_T=\nabla_W T_{\Rell}$ & Trace-log gradient objects. \\
Optimizer & $\Delta\W_\perp$ & Trace-log projected update. \\
WW-PGD & $\Tell$, $m=|\Tell|$ & Selected tail and size. \\
WW-PGD & $\tilde{\lambda}_k^\star=A k^{-q}$, $q=1$ & Rank-order tail target. \\
WW-PGD & $\eta_B,\eta_C,h_e$ & Blend, Cayley, homotopy factors. \\
\bottomrule
\end{tabular}}
\end{center}
\clearpage

\appendix
\section{Optimizer experiment details}
\label{app:optimizer_experiments}
\small

This appendix records the optimizer protocols.
%Both tests use MNIST and the same three-layer MLP, $784\rightarrow512\rightarrow512\rightarrow10$, with ReLU nonlinearities. The linear layers are logged as FC1--FC3. MNIST inputs use mean $0.1307$ and standard deviation $0.3081$.
Both tests use MNIST and three-layer MLPs, with the configurations specified in the accompanying notebooks.
At checkpoints, \WW{} records the fitted HTSR exponent $\alpha$ and the tail diagnostics for each fully connected layer.

\paragraph{Reproduction notebooks.}
The Muon panel uses the public notebooks \href{https://github.com/CalculatedContent/WeightWatcher-Examples/blob/main/AdamWvsMuon.ipynb}{AdamWvsMuon.ipynb}, \href{https://github.com/CalculatedContent/WeightWatcher-Examples/blob/main/MLP3-MNIST-AdamW.ipynb}{MLP3-MNIST-AdamW.ipynb}, and \href{https://github.com/CalculatedContent/WeightWatcher-Examples/blob/main/MLP3-MNIST-Muon.ipynb}{MLP3-MNIST-Muon.ipynb}. The WW-PGD wrapper is reproduced from \href{https://github.com/CalculatedContent/WW_PGD/blob/main/WW_PGD_QuickStart.ipynb}{WW\_PGD\_QuickStart.ipynb}. The plotted curves use the settings below.

\paragraph{Muon test.}
The Muon test compares AdamW and Muon on the same model and data. AdamW gives the baseline trajectory. Muon is applied to matrix-valued hidden-layer updates. Scalar, bias, and other non-matrix parameters are handled by the associated base optimizer. The short-horizon panel tracks the early optimizer separation. The long-horizon panel tracks the slower Muon approach over hundreds of epochs. Training and test accuracy are logged with the layerwise curves $\alpha(t)$, $\ell\in\{\mathrm{FC1},\mathrm{FC2},\mathrm{FC3}\}$.

\paragraph{WW-PGD test.}
The WW-PGD test compares AdamW with AdamW+WW-PGD. The minibatch optimizer is fixed. The batch size is $128$. AdamW uses learning rate $10^{-3}$. Gradients are clipped to norm $1.0$. The main run uses $35$ epochs and five independent runs. At every epoch, \WW{} analyzes the model with \texttt{detX=True}, \texttt{randomize=False}, and \texttt{plot=False}. WW-PGD applies one epoch-boundary spectral retraction with \texttt{min\_tail}=5, $q=1$, $\eta_B=0.5$, $\eta_C=0.25$, \texttt{use\_detx=True}, \texttt{warmup\_epochs}=0, \texttt{ramp\_epochs}=5, and \texttt{enable\_trap\_pgd=False}.

\paragraph{WW-PGD correction.}
For each eligible layer, WW-PGD forms $\lambda_i=s_i^2$ and selects the retained tail from \WW{} diagnostics for $\tilde{\lambda}^{+}_{\rm PL}$ and $\tilde{\lambda}^{+}_{\Rell}$. It sorts the tail eigenvalues in decreasing order and uses the target $\tilde{\lambda}_k^\star=A k^{-1}$. The amplitude $A$ preserves the retained tail trace-log. The moved tail is recentered, reconstructed with the original singular vectors, and blended back into the post-AdamW layer. The standard test loader uses ordinary MNIST. The perturbed loader uses random rotation by $7^\circ$, random translation by $3\%$, small Gaussian blur, and no horizontal flip.

\clearpage
%\enlargethispage{3\baselineskip}

\end{document}